\documentclass[journal]{IEEEtran}
\usepackage{amsmath,amsfonts}
\usepackage{algorithmic}
\usepackage{array}
\usepackage{textcomp}
\usepackage{stfloats}
\usepackage{url} 
\usepackage{verbatim}
\usepackage{graphicx}
\usepackage{cite, amsmath,amssymb, bm}
\usepackage{multirow}
\usepackage[linesnumbered,ruled]{algorithm2e}
\def\BibTeX{{\rm B\kern-.05em{\sc i\kern-.025em b}\kern-.08em
    T\kern-.1667em\lower.7ex\hbox{E}\kern-.125emX}}
\usepackage{balance}

\usepackage[labelformat=simple]{subcaption}

\usepackage{subfiles}
\usepackage{bbding}
\usepackage{color, xcolor}
\usepackage{amsthm}
\usepackage[resetlabels]{multibib}
\usepackage{multicol} 
\usepackage{lipsum}

\newcites{appendix}{Appendix References}

\usepackage{hyperref}
\hypersetup{ colorlinks=false, linkcolor=blue, anchorcolor=blue, citecolor=blue}

\allowdisplaybreaks[4]

\newtheorem{theorem}{Theorem}

\newtheorem{lemma}{Lemma}

\newtheorem{proposition}{Proposition}

\newtheorem{corollary}{Corollary}

\newtheorem{property}{Property}

\newtheorem{remark}{Remark}

\newtheorem{claim}{Claim}

\begin{document}
\title{NeuPAN: Direct Point Robot Navigation with End-to-End Model-based Learning
}

\author{Ruihua Han$^{1,2}$, Shuai Wang$^{3, \dagger}$, Shuaijun Wang$^{2}$, Zeqing Zhang$^{1}$, Jianjun Chen$^{2}$, Shijie Lin$^{1}$, \\ Chengyang Li$^{3, 4}$, Chengzhong Xu$^{5}$, Yonina C. Eldar$^{6}$, Qi Hao$^{2, \dagger}$, Jia Pan$^{1, \dagger}$

\thanks{$^{1}$ Department of Computer Science, University of Hong Kong, Hong Kong, {\tt\footnotesize hanrh@connect.hku.hk}}
\thanks{$^{2}$ Department of Computer Science and Engineering, Southern University
of Science and Technology, Shenzhen, China}
\thanks{$^{3}$ Shenzhen Institute of Advanced Technology, Chinese Academy of Sciences,
Shenzhen, China}
\thanks{$^{4}$ Robotics and Autonomous Systems Thrust, Hong Kong University of
Science and Technology (Guangzhou), Guangzhou, China}
\thanks{$^{5}$ IOTSC, University of Macau, Macau, China}
\thanks{$^{6}$ Weizmann Institute of Science, Rehovot, Israel}

\thanks{$^\dagger$ denotes the corresponding authors: Jia Pan ({\tt\footnotesize jpan@cs.hku.hk}), Qi Hao ({\tt\footnotesize hao.q@sustech.edu.cn}), Shuai Wang ({\tt\footnotesize s.wang@siat.ac.cn}).}
  }

\maketitle

\begin{abstract}
Navigating a nonholonomic robot in a cluttered, unknown environment requires accurate perception and precise motion control for real-time collision avoidance. This paper presents NeuPAN: a real-time, highly accurate, map-free, easy-to-deploy, and environment-invariant robot motion planner. Leveraging a tightly coupled perception-to-control framework, NeuPAN has two key innovations compared to existing approaches: 1) it directly maps raw point cloud data to a latent distance feature space for collision-free motion generation, avoiding error propagation from the perception to control pipeline; 2) it is interpretable from an end-to-end model-based learning perspective. The crux of NeuPAN is solving an end-to-end mathematical model with numerous point-level constraints using a plug-and-play (PnP) proximal alternating-minimization network (PAN), incorporating neurons in the loop. This allows NeuPAN to generate real-time, physically interpretable motions. It seamlessly integrates data and knowledge engines, and its network parameters can be fine-tuned via backpropagation. We evaluate NeuPAN on a ground mobile robot, a wheel-legged robot, and an autonomous vehicle, in extensive simulated and real-world environments. Results demonstrate that NeuPAN outperforms existing baselines in terms of accuracy, efficiency, robustness, and generalization capabilities across various environments, including the cluttered sandbox, office, corridor, and parking lot. We show that NeuPAN works well in unknown and unstructured environments with arbitrarily shaped objects, transforming impassable paths into passable ones.
\end{abstract}


\section{Introduction}\label{section 1}

Real-time navigation of robots in densely confined environments is crucial for a broad range of applications, including household robotics, logistics, and autonomous driving. In contrast to wide-open environments, the aforementioned cluttered scenarios require robot perceptions (i.e., providing necessary information about the environment) and motions (i.e., computing a sequence of feasible actions to connect the current and target states) to be very accurate under kinematic constraints; otherwise, efficiency or safety could be greatly compromised. The situation becomes even more nontrivial if the navigation system needs to work properly in previously unseen environments. 

Existing approaches fail to address the problem because of the following reasons: 
1) Compressing a high-dimensional sensor space (e.g., massive points per second) into a low-dimensional action space (e.g., throttle and steer) while guaranteeing interpretability and preventing error propagation is nontrivial~\cite{zhang2020optimization,kousik2020safe}; 2) The navigation problem is $\mathsf{PSPACE}$ hard, and existing solutions~\cite{rosmann2017kinodynamic, zhou2021raptor, han2022reinforcement} have to balance the trade-off between accuracy and complexity, resulting in either inexact motions, leading to conservative policies, or delayed motions, increasing the risk of collisions; 3) The solution should be stable, complete, and user-friendly, involving little hand-engineering and retraining for new robots and environments~\cite{chen2019crowd, fan2020distributed}. The above system, algorithm, and engineering issues make dense-scenario navigation with only onboard computation resources a long-standing challenge. 

\begin{figure*}[t]
    \centering
        \includegraphics[width=1\textwidth]{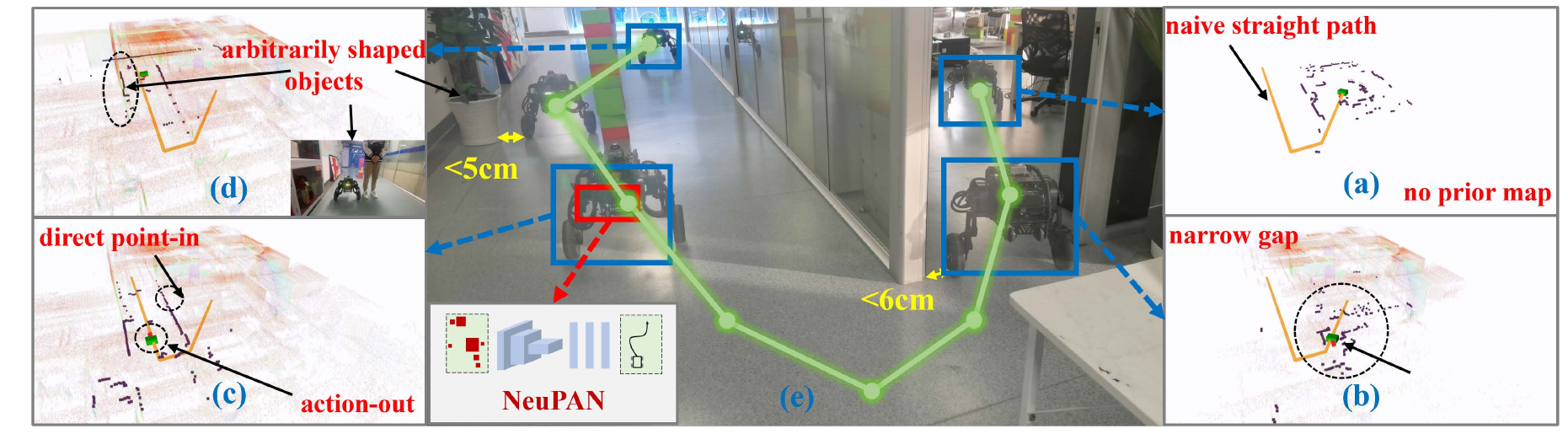}
        \caption{ Wheel-legged robot mapless navigation in the office empowered by NeuPAN: (a) the robot navigates along a naive path without a prior map; (b) the robot navigates through narrow gaps; ($<6\,$cm); (c) NeuPAN utilizes a ``direct point-in action-out'' manner at a high frequency; (d) NeuPAN can handle arbitrarily shaped (moving) objects. (e) the robot trajectory.} 
        \label{fig1}
\end{figure*}

To address these issues, this paper proposes NeuPAN, a direct point, end-to-end, model-based learning approach designed to achieve a real-time, highly accurate (e.g., over 2x improvement compared to state-of-the-art approaches), map-free, easy-to-deploy, and environment-invariant collision avoidance. Particularly, our solution employs lidar sensors to provide obstacle points because of their ability to deliver direct, active, and accurate depth measurements of environments, as well as their insensitivity to illumination variations and motion blur. Building upon these inherent features of lidar scans and empowered by NeuPAN, a wheel-legged robot can navigate through narrow gaps without a prior map while avoiding collisions with arbitrarily shaped dynamic objects, as demonstrated in Fig.~\ref{fig1}.

The key contributions that lead to the superior performance of our system are as follows: 1) In contrast to existing approaches that convert point clouds to convex sets or occupancy grids, and adopt inexact minimum distances, NeuPAN directly processes massive raw points and computes the point flow based on predictive ego motion. Then a neural encoder is leveraged to map the point flow to a corresponding latent distance feature space, which represents the exact distance from the ego robot to the obstacle points within the receding horizon. 
2) The latent distance features are seamlessly incorporated into the motion planner as a norm regularizer added to the loss function, accounting for the reward for departing from obstacles. The predictive states and actions generated by the motion planner are fed back to the front-end for re-computation of the point flow. This forms a tightly-coupled loop between perception and control. 
3) To gain deeper insights into NeuPAN, we formulate an end-to-end mathematical program with point-level constraints. We demonstrate that NeuPAN is equivalent to solving this problem using the plug-and-play (PnP) proximal alternating-minimization (PAN) algorithm. To the best of our knowledge, this marks the first approach where a navigation algorithm is interpreted from a model-based learning perspective, thereby seamlessly integrating data-driven systems with rigorous mathematical models.
4) We conduct various experiments to evaluate the effectiveness of the developed NeuPAN system. Exhaustive benchmark comparisons in various simulated scenarios show that NeuPAN achieves consistently higher success rates with shorter navigation time than those of state-of-the-art robot navigation systems. We finally show the effectiveness of NeuPAN in real-world dynamic, cluttered, and unstructured environments including the sandbox, office, corridor, and parking lot on ground mobile robots, wheel-legged robots, and passenger autonomous vehicles. Experiment videos and more details can be found on our project page: \url{https://hanruihua.github.io/neupan_project/.}    

The remainder of this paper is organized as follows.
Section \ref{section 2} reviews related work. 
Section \ref{section 3} states the problem. 
Section \ref{section 4} describes the system architecture.
Subsequently, Section~\ref{section 5} presents the neural encoder network and the neural regularized motion planner. 
Section \ref{section 6} presents simulation and real-world experiments. 
Finally, Section~\ref{section 7} concludes this work.

\section{Related Work}\label{section 2}

\subsection{Modular vs. End-to-End}

Modular approaches, which divide navigation into distinct modules (e.g., an object detector and a motion planner in the simplest case), are the most widely adopted framework due to its reliability and ease of debugging~\cite{han2020cooperative, tordesillas2021faster, zhang2022generalized}. However, these approaches suffer from error propagation from perception to control modules: 1) At the front-end perception layer, the detected object representation (e.g. bounding box) generated by even the best detector can deviate from the ground truth, necessitating the incorporation of a larger safety distance in the motion planner to guarantee worst-case collision avoidance; 2) At the intermediate representation layer, the box or convex sets cannot exactly match real-world objects that may possess arbitrary nonconvex shapes; These errors in individual layers will add up together to the navigation pipeline, making the modular approach prone to getting stuck in cluttered environments.

To mitigate the error propagation inherent in the modular approach, recent robot navigation is experiencing a paradigm shift towards an end-to-end approach, which directly maps sensor inputs to motion outputs~\cite{devo2020towards, fan2020distributed, xiao2023barriernet}. Earlier end-to-end solutions focused on using a single black-box deep neural network (DNN) to learn the mapping but were later found difficult to train and not generalizable to unseen situations. The emerging end-to-end approach adopts multiple blocks but differs from the modular approach in three aspects: 1) it exchanges features (e.g., encoder outputs) instead of explicit representations (e.g., bounding boxes) among modules; 2) the interaction among modules is bidirectional instead of unidirectional; 3) the entire system can be trained in an end-to-end manner. Such generalized end-to-end approaches have shown effectiveness in vision-based autonomous driving and encoder-based robot manipulation. For instance, the unified autonomous driving (UniAD) framework consists of backbone, perception, prediction, and planning modules, with task queries as interfaces connecting each node~\cite{hu2023planning}.

The majority of the industry practice has exploited end-to-end approaches for autonomous driving tasks, which leverage vision transformers to generate bird's eye view (BEV) ~\cite{li2022bevformer} for occupancy mapping~\cite{mescheder2019occupancy} that is directly used for subsequent planning. Similar insights have been obtained for robot manipulation, which can be accomplished by a motion planning network (MPNet) consisting of a neural encoder and a motion planner~\cite{qureshi2020motion}. The limitation of these purely data-driven techniques lies in their lack of interpretability. This makes it difficult to adjust the network parameters for new robot platforms. When generalizing to a new environment, they often require extensive hand-engineering and prolonged training and usually need to be accompanied by other conservative strategies to guarantee safety. These methods also demand significant data collection and annotation efforts. 

\subsection{Model-based vs. Learning-based}

Algorithms determine the inference mapping of each module in modular or end-to-end frameworks. Existing algorithms can be categorized into model-based and learning-based.

Model-based algorithms utilize mathematical or statistical formulations that represent the underlying physics, prior information, and domain knowledge. Classical model-based algorithms include graph search-based, sampling-based, and optimization-based.
Graph search-based techniques, e.g., A$^*$, approximate the configuration space as a discrete grid space to search the path with minimum cost~\cite{montemerlo2008junior}.
Sampling-based techniques, e.g., rapidly-exploring random tree (RRT) or RRT$^*$, probe the configuration space with a sampling strategy~\cite{salzman2016asymptotically}. The fast-likelihood-based collision avoidance method (Falco)~\cite{zhang2020falco}, maximizes the likelihood to reach the goal in determining the next navigation actions.   
Optimization-based techniques generate trajectories through minimizing cost functions under dynamics and kinematic constraints~\cite{zhang2020optimization, tordesillas2021faster}.

In the context of dense-scenario navigation, optimization-based techniques, such as model predictive control (MPC), are more attractive due to their ability to compute the current optimal input to produce the best possible behavior in the future, resulting in high-performance trajectories. One main drawback of optimization-based techniques is their high computational cost, which limits their real-time applications. Particularly, the number of collision avoidance constraints is proportional to the number of obstacles (spatial), and the length of the receding horizon (temporal). In cluttered environments, where the shape of obstacles is also taken into account, the computation time is further multiplied by the number of surfaces from each obstacle.
To overcome the nonconvex collision avoidance constraints, the optimization-based collision avoidance (OBCA) algorithm is developed in~\cite{zhang2020optimization} for full-shape control objects, which adopts duality to reformulate the exact distance based collision avoidance constraints. Moreover, full-shape robot navigation is further accelerated by a space-time decomposition algorithm in~\cite{han2023rda}. However, the frequency of these algorithms is not satisfactory when dealing with a dozen of objects.
On the other hand, most existing works forgo exact distances and adopt inexact ones, such as center-point distances, approximate signed distances, or spatial-temporal corridors~\cite{ding2021epsilon, han2023efficient}. Additionally, it is possible to convert hard constraints into soft regularizers for speed-up, as exemplified by the EGO planner, which removes the collision avoidance constraints by adding penalty terms~\cite{zhou2020ego}. While inexact algorithms achieve a high frequency (e.g., up to $100\,$Hz), they are not suitable for dense-scenario navigation, as discussed earlier in Section~\ref{section 1}.

\begin{table*}[t]
  \centering
  \caption{Characteristics comparison of NeuPAN and existing navigation approaches}
  \label{related_work}
  \resizebox{\textwidth}{!}{%
    \begin{tabular}{c|c|c|c|c|c|c|c}
      \hline
      Scheme                                        & System Input & Full Shape     & Exact distance & Map Free       & Generalization & Latency$^\dag$   & Constraints Guarantee \\ \hline
      TEB~\cite{rosmann2017kinodynamic}             & Grids        & \CheckmarkBold & \XSolidBrush   & \XSolidBrush   & \CheckmarkBold & Low              & \CheckmarkBold        \\ \hline
      OBCA~\cite{zhang2020optimization}             & Object Sets  & \CheckmarkBold & \CheckmarkBold & \XSolidBrush   & \CheckmarkBold & High             & \CheckmarkBold        \\ \hline
      RDA~\cite{han2023rda}                         & Object Sets  & \CheckmarkBold & \CheckmarkBold & \CheckmarkBold & \CheckmarkBold & Medium$\sim$high & \CheckmarkBold        \\ \hline
      STT~\cite{han2023efficient}                   & Object Sets  & \CheckmarkBold & \XSolidBrush   & \XSolidBrush   & \CheckmarkBold & Low              & \XSolidBrush          \\ \hline
      AEMCARL~\cite{wang2022adaptive}               & Object Poses & \XSolidBrush   & N/A            & \CheckmarkBold & \XSolidBrush   & Low              & \XSolidBrush          \\ \hline
      Hybrid-RL~\cite{fan2020distributed}           & lidar scan & \XSolidBrush   & N/A            & \CheckmarkBold & \XSolidBrush   & Low              & \XSolidBrush          \\ \hline
      MPC-MPNet~\cite{qureshi2020motion, li2021mpc} & Voxel & \CheckmarkBold & \XSolidBrush   & \CheckmarkBold & \XSolidBrush   & Low              & \CheckmarkBold        \\ \hline
      Falco~\cite{zhang2020falco} & Voxel & \CheckmarkBold & \XSolidBrush   & \CheckmarkBold & \CheckmarkBold   & Low              & \CheckmarkBold        \\ \hline
      NeuPAN (Ours)                                 & Points & \CheckmarkBold & \CheckmarkBold & \CheckmarkBold & \CheckmarkBold & Low$\sim$Medium  & \CheckmarkBold        \\ \hline
    \end{tabular}%
  }
  \caption*{$\dag$ low latency: $<50\,$ms; medium latency: $50\sim500\,$ms; high latency: $>500\,$ms.}
\end{table*}

In contrast to model-based approaches, learning algorithms dispense with analytical models by extracting features from data. This is particularly useful in complex systems where analytical models are not known. Consequently, learning algorithms are promising for end-to-end navigation tasks. In this direction, reinforcement learning (RL) is a major paradigm~\cite{tampuu2020survey}. The idea of RL is to learn a neural network policy by interacting with the environment and maximizing cumulative action rewards~\cite{francis2020long}. In particular, RL has been widely used for the dynamic collision avoidance (so-called CARL based approaches), where the motion information of obstacles is mapped to the robot actions directly, such as CADRL~\cite{chen2017decentralized}, LSTM-RL~\cite{everett2018motion}, SARL~\cite{chen2019crowd},  RGL~\cite{chen2020relational}, and AEMCARL~\cite{wang2022adaptive}.
However, the performance of RL-based methods is influenced by the training dataset's distribution. Therefore, these methods are promising in simulation but often challenging to implement in real-world settings. Additionally, finding a suitable reward function for dense-scenario navigation is often difficult.

As mentioned in Section~\ref{section 2}-A, learning-based algorithms lack interpretability. Hence, an emerging paradigm involves the cross-fertilization of model-based and learning-based algorithms. This has led to a variety of model-based learning approaches in robot navigation. In particular, a direct way is to use learning algorithms to imitate complex dynamics or numerical procedures by transforming iterative computations into a feed-forward procedure. This is achieved by learning from solvers' demonstrations using DNNs. Examples include the neural PID in~\cite{lin2019flying} and the neural MPC in~\cite{salzmann2023real}.
On the other hand, learning-based algorithms are adopted to generate candidate trajectories, reducing the solution space for subsequent model-based algorithms, e.g., MPNet~\cite{qureshi2020motion}, MPC-MPNet~\cite{li2021mpc}, NFMPC~\cite{sacks2023learning}, and MPPI~\cite{williams2018information}. Note that trajectory generation is often learned from classical sampling-based approaches.
Lastly, learning algorithms can be used to adjust the hyper-parameters involved in model-based algorithms. For instance, the cost and dynamics terms of MPC can be learned by differentiating parameters with respect to the policy function through the optimization problem~\cite{amos2018differentiable, agrawal2021learning}.

Our framework, NeuPAN, also belongs to the model-based learning approach. However, in contrast to the aforementioned works that are partially interpretable, NeuPAN is \textbf{end-to-end interpretable} from perception to control. This is achieved by solving an end-to-end optimization problem with numerous point-level constraints using the PnP PAN. This makes NeuPAN suitable for dense-scenario collision avoidance with nonholonomic robots, whereas existing results~\cite{zhou2020ego, fan2020distributed, han2023efficient} usually consider holonomic drones or nonholonomic robots in wide-open scenarios. Similar to~\cite{amos2018differentiable, agrawal2021learning}, parameters within the costs and constraints of NeuPAN can be trained via function differentiation in an end-to-end fashion. Hence, NeuPAN enjoys easy-to-deploy and environment-invariant features.

\subsection{Comprehensive Comparison with Existing Solutions}
Comprehensive comparison with celebrated navigation approaches is provided in Table~\ref{related_work}. In particular, TEB~\cite{rosmann2017kinodynamic}, OBCA~\cite{zhang2020optimization}, RDA~\cite{han2023rda}, STT~\cite{han2023efficient}, AEMCARL~\cite{wang2022adaptive}, and Falco~\cite{zhang2020falco} are modular approaches. Their inputs are grids, poses, boxes, or sets generated by a pre-established grid map or a front-end object detector. All these methods involve error propagation. In contrast to AEMCARL which treats the ego robot as a point (or ball), TEB, OBCA, RDA, STT, and Falco consider the shape of the ego robot and obstacles. However, TEB, STT, and Falco involve approximations when computing the distance. OBCA and RDA are the most accurate approaches but require high computational latency, e.g., up to second-level (i.e., less than $1\,$Hz).
On the other hand, Hybrid-RL~\cite{fan2020distributed}, MPC-MPNet~\cite{li2021mpc}, and NeuPAN (ours) are end-to-end approaches. By treating raw lidar points as inputs, these methods are free from error propagation. Hybrid-RL adopts a single network for point input and action output. Despite its high-level integration, Hybrid-RL suffers from poor generalization capability. MPC-MPNet, and NeuPAN adopt two networks, one for encoding and the other for planning: MPC-MPNet uses a neural encoder to map points into voxel features; NeuPAN uses a neural encoder to map points and predictive ego motions into latent distance features. All of them consider the shape of objects; however, due to the discretization from points to voxels, MPC-MPNet cannot compute the exact distances between two shapes. They also lack generalization as the trajectory generator is scenario-dependent and needs retraining for new robots. Our NeuPAN overcomes these drawbacks, at the cost of a slightly higher computational load.

NeuPAN is the first approach that builds the end-to-end mathematical model (i.e., point-in and action-out), and uses model-based learning to solve it.
As such, NeuPAN is fully explainable.
Owing to this new feature, NeuPAN generates very accurate, end-to-end, physically-interpretable motions in real time. This empowers autonomous systems to work in dense unstructured environments, which are previously considered impassable and not suitable for autonomous operations,
triggering new applications such as cluttered-room housekeeper and limited-space parking.
Compared to existing end-to-end methods,
\cite{devo2020towards, fan2020distributed, xiao2023barriernet,hu2023planning,li2022bevformer,mescheder2019occupancy,qureshi2020motion},
NeuPAN provides mathematical guarantee and leads to much lower uncertainty and higher generalization capability.
Compared to existing model-based motion planning methods~\cite{zhang2020optimization, qureshi2020motion, han2023rda,han2023efficient}, NeuPAN is more accurate, since conventional motion planning belongs to modular approaches and involves error propagation.
There also exist other model-based learning methods \cite{lin2019flying,salzmann2023real,qureshi2020motion,sacks2023learning,williams2018information} for robot navigation.
However, these methods are not end-to-end, e.g.,
\cite{lin2019flying,salzmann2023real} consider state-in action-out, and \cite{qureshi2020motion,sacks2023learning,williams2018information} consider point-in trajectory-out.
The work \cite{li2021mpc} considers point-in action-out, by bridging \cite{qureshi2020motion} with a back-end controller.
However, such a bridge loses mathematical guarantee and no longer solve the inherent end-to-end model.
The methods \cite{li2021mpc,qureshi2020motion} also leverage neural encoder like transformer to compress the sensor data, and this neural encoder has no interpretability.

\section{Problem Statement}\label{section 3}

We consider an end-to-end navigation approach with points in and actions out for full-dimensional robots based on the model predictive control (MPC) framework. To reach the goal state through a control sequence, a perception-to-control optimization problem over a receding horizon $\mathcal{H} = \{t, \cdots, t+H\}$ is solved iteratively at each time step $t$.

\emph{1) Robot Kinematics}: Given the control vector $\mathbf{u}_t$, the current state $\mathbf{s}_t$ and subsequent state $\mathbf{s}_{t+1}$ should adhere to the discrete-time kinematic model:
\begin{equation}
    \mathbf{s}_{t + 1} = \mathbf{s}_t + f(\mathbf{s}_t, \mathbf{u}_t)\Delta t,
\end{equation}
where $\Delta t$ is the time slot between two states. We assume that the function $f({\mathbf{s}_t},{\mathbf{u}_t})$ is linear with respect to the state ${\mathbf{s}_t}$ and control vector $\mathbf{u}_t$. In scenarios involving non-linear dynamics, these functions can be approximated using linearization techniques, such as the Taylor series expansion. Accordingly, this constraint can be rewritten in a linear form:
\begin{equation}
    {\mathbf{s}_{t + 1}} = \mathbf{A}_t{\mathbf{s}_t} + \mathbf{B}_t {\mathbf{u}_t} + {\mathbf{c}_t},
    \label{kinematics}
\end{equation}
where $(\mathbf{A}_t, \mathbf{B}_t, \mathbf{c}_t)$ are coefficient matrices at time step $t$. Examples for ackermann and differential models are provided in Appendix A (Supplementary Material). Particularly, the state $\mathbf{\overline{{s}}_{t}}$ at initial time point under the MPC framework can be measured by odometry or provided by the localization system. Due to physical limits, the control vector $\mathbf{u}_t$ should belong to a feasible control that set limits on the absolute values and rates of changes. This yields the following constraints:
\begin{equation}
    \begin{aligned}
            & \mathbf{u}_\text{min} \preceq \mathbf{u}_{h}  \preceq \mathbf{u}_\text{max}, \\
            &\mathbf{a}_{\text{min}} \preceq \mathbf{u}_{h+ 1} - \mathbf{u}_{h}  \preceq \mathbf{a}_{\text{max}},\ \forall h \in \mathcal{H}.
    \end{aligned}\label{control}
\end{equation}
Consequently, we define the kinematic feasible set $\mathcal{F}$ as the set of all states and control vectors that satisfy the above constraints Eq.~\eqref{kinematics}, \eqref{control}\footnote{Note that $\mathbf{{s}}_{t}=\mathbf{\overline{{s}}_{t}}$, where $\mathbf{\overline{{s}}_{t}}$ is the current robot state in each horizon.}

\emph{2) Robot Model}:
The space occupied by the ego robot at the origin of the coordinate system can be represented by a compact convex set $\mathbb{C}$. Based on the conic inequality representation~\cite{boyd2004convex}, $\mathbb{C}$ is given by
\begin{equation}
    \mathbb{C} = \{ \mathbf{x} \in \mathbb{R}^n | \mathbf{G} \mathbf{x} \preceq_\mathcal{K} \mathbf{h}\},
\end{equation}
where, $ \mathbf{G} \in \mathbb{R}^{l \times n} $ and $ \mathbf{h} \in \mathbb{R}^l$ represent the rotations and translations of surfaces with respect to the zero point, respectively, and $l$ is the minimum number of surfaces that can represent the shape of ego robot.
Symbol $\mathcal{K}$ is a proper cone, with its partial ordering on $\mathbb{R}^n$ defined as: $\mathbf{x} \preceq_\mathcal{K} \mathbf{y} \Longleftrightarrow \mathbf{y}-\mathbf{x} \in \mathcal{K}$.
For instance, for a polygonal robot, $\mathcal{K}=\mathbb{R}^n_+$ and $l$ corresponds to the number of hyperplanes. Consequently, given the state $\mathbf{s}_t$, the occupied space at $t$-th time frame, denoted by the convex compact set $\mathbb{Z}_t$\footnote{Nonconvex objects can be represented as unions of convex objects.}
\begin{equation}
    \mathbb{Z}_t({\mathbf{s}_t}) =
    \left\{\mathbf{R}_t({\mathbf{s}_t})\mathbf{x} + \mathbf{t}_t({\mathbf{s}_t})|\mathbf{x}\in\mathbb{C}
    \right\},
    \label{robject}
\end{equation}
where $\mathbf{R}_t({\mathbf{s}_t}) \in \mathbb{R}^{n \times n}$ is the rotation matrix representing the orientation of the robot and $\mathbf{t}_t({\mathbf{s}_t}) \in \mathbb{R}^n$ is the translation vector denoting the position of the robot. For instance, for a state in the 2D space with position and orientation angle \(\mathbf{s}_t = [x_t, y_t, \theta_t]^T\), the rotation and translation matrices can be obtained by:
\begin{equation}
    \mathbf{R}(\mathbf{s}_t) = \begin{bmatrix}
        \cos \theta_t & -\sin \theta_t \\
        \sin \theta_t & \cos \theta_t
    \end{bmatrix}, \quad \mathbf{t}(\mathbf{s}_t) = \begin{bmatrix}
        x_t \\
        y_t
    \end{bmatrix}.
\end{equation}

\emph{3) Point-Level Collision Avoidance Constraints}: We adopt a set of points to represent the obstacles in the environment at time $t$, denoted by $\mathbb{P}_t = \{ \mathbf{p}_t^1, \ldots, \mathbf{p}_t^i, \ldots, \mathbf{p}_t^M \}$, where $\mathbf{p}_t^i \in \mathbb{R}^n$ is the $n$-dimensional position of the $i$-th point in the global coordinate system, and $M$ is the total number of points. The set of obstacle points can be obtained directly from raw lidar scans in real-time. Therefore, conventional operations such as object detection or occupancy grid maps are not required. Consequently, the minimum distance between two non-intersecting sets, the robot $\mathbb{Z}_t(\mathbf{s}_t)$ and the $t$-th $\mathbb{P}_t$ can be expressed as:
    \begin{align}
         & {\bf{dist}}(\mathbb{Z}_t(\mathbf{s}_t), \mathbb{P}_t) = \min \limits_\mathbf{e}  \left\{ {\left. {{{\left\| \mathbf{e} \right\|}_2}} \right|(\mathbb{Z}_t(\mathbf{s}_t) + \mathbf{e}) \cap \mathbb{P}_t \ne \emptyset } \right\}
        \nonumber                                                                                                                                                                                                                           \\
         & =\min
        \{{D_{1,t}}(\mathbb{Z}_t(\mathbf{s}_t), \mathbf{p}_t^1),\cdots,{D_{M,t}}(\mathbb{Z}_t(\mathbf{s}_t), \mathbf{p}_t^M)\},
        \label{dist}
    \end{align}
    where \( \mathbf{e} \) denotes the smallest translation vector that brings these two sets into contact, as introduced in~\cite{schulman2014motion, zhang2020optimization}. $D_{i,t}(\mathbb{Z}(\mathbf{s}_t), \mathbf{p}_t^i)$ denotes the exact minimum distance between the point $\mathbf{p}^i_t$ and the robot $\mathbb{Z}_t(\mathbf{s}_t)$.

The computation of distance $D_{i,t}$ is a convex optimization problem given by:
    \begin{equation}
        \begin{aligned}
            \mathop {{\rm{min}}}\limits_{\mathbf{x}}~ & {\left\| {\mathbf{R}_t(\mathbf{s}_t)\mathbf{x} + \mathbf{t}_t(\mathbf{s}_t) - \mathbf{p}_t^i} \right\|_2^2}
            \\
            \text { s.t. }~                           & ~\mathbf{G}\mathbf{x}{ \preceq _{\mathcal{K}}}\mathbf{h}.
            \label{dis_opt}
        \end{aligned}
    \end{equation}
    We refer to the above convex optimization problem as the primal representation for computing the minimum distance. Accordingly, we have the point-level collision avoidance constraints:
    \begin{equation}
        \begin{aligned}
            {\bf{dist}}(\mathbb{Z}_{t}(\mathbf{s}_{t}), \mathbb{P}_{t}) \geq d_{\min},
            \label{avoidance_label}
        \end{aligned}
    \end{equation}
    where $d_{\min}$ is a safety distance. Generally, these collision avoidance constraints are nonconvex and nondifferentiable~\cite{zhang2020optimization}.

\emph{4) Objective Function}: Given the start state $\mathbf{s}_\text{start}$ and the goal state $\mathbf{s}_\text{goal}$, our objective is to find a control sequence $\mathcal{U} = \{\mathbf{u}_0, \cdots, \mathbf{u}_{T}\}$ and associated trajectory $\mathcal{S} = \{\mathbf{s}_0, \cdots, \mathbf{s}_{T}\}$, such that ${\{{\mathcal{S}, \mathcal{U} }\} \in \mathcal{F}}$, $\mathbf{s}_0 = \mathbf{s}_{\text{start}}$ and $\|\mathbf{s}_T - \mathbf{s}_{\text{goal}}\|_2 \leq \epsilon$, where $\epsilon$ is the navigation tolerance, while avoiding the obstacles in the environment. We adopt the naive straight line connecting $\mathbf{s}_{\text{start}}$ and $\mathbf{s}_{\text{goal}}$ (same as \cite{schulman2014motion}) as the initialization~\footnote{ For Ackermann kinematics, Dubins or Reeds-Shepp paths can be adopted}, with a list of interpolating points $\{\mathbf{s}_0^\diamond, \mathbf{s}_1^\diamond, \cdots\}$ on the line. In the case of multiple goal states, the initial path contains multiple line segments. A desired speed $\mathbf{u}_{\text{speed}}^\diamond$ is used as the initial value for the robot to maintain. In our case, the cost function $C_{0}(\mathcal{S}, \mathcal{U})$ is constructed as the distance between the output and naive trajectories over the horizon $\mathcal{H} = \{t, t+1, \cdots, t+H\}$:
\begin{align}
    C_{0}(\mathcal{S}, \mathcal{U}) & = \sum_{h=t}^{t+H}\Big(\left\|\mathbf{q} \circ (\mathbf{s}_{h+1} - \mathbf{s}_{h+1}^\diamond)\right\|_2^2 \nonumber
    \\
                                    & + \left\|\mathbf{p} \circ (\mathbf{u}_{\text{speed}, h} - \mathbf{u}_{\text{speed}}^\diamond)\right\|_2^2\Big),
    \label{cost0}
\end{align}
where $\{\mathbf{q}, \mathbf{p}\}$ are weighted parameters. A larger value of $\mathbf{q}$ encourages the robot to follow the initialized trajectory more closely, while a larger value of $\mathbf{p}$ control the robot to maintain the desired speed. $\{\mathbf{s}_{h+1}^\diamond, \forall h \in \mathcal{H} \}$ is selected from the naive path according to the closest distance criteria.

\emph{5) Problem Formulation}: The direct point robot navigation problem is formulated as the following model predictive perception and control (MPPC) optimization problem over horizon $\mathcal{H}$:
    \begin{subequations}
        \begin{align}
            \mathsf{P}: \min_{\substack{{\{{\mathcal{S}, \mathcal{U} }\} \in \mathcal{F}}}}~~ & C_{0}(\mathcal{S}, \mathcal{U})  \label{ps_a}                                                                       \\
            \quad \text{subject to}~~~
                                                                                              & {\bf{dist}}(\mathbb{Z}_{h}(\mathbf{s}_{h}), \mathbb{P}_{h}) \geq d_{\min}, ~\forall h \in \mathcal{H}. \label{ps_b}
        \end{align}
        \label{ps}
    \end{subequations}

\textbf{Technical Challenge}: $\mathsf{P}$ represents the direct point robot navigation problem. The input includes the point cloud $\mathbb{P}_t$, naive path $\{\mathbf{s}_0^\diamond, \mathbf{s}_1^\diamond, \cdots\}$, and desired speed $\mathbf{u}_{\text{speed}}^\diamond$. The output is the optimized control vector and associated trajectory $\{\mathcal{S}, \mathcal{U}\}$. The collision avoidance constraints in $\mathsf{P}$ are bilevel, involving inner-level distance computations, and large-scale, as the number of point-level constraints, $MH$, can reach thousands. Existing model-based approaches convert $\mathbb{P}_t$ to convex sets~~\cite{zhang2020optimization, han2023rda}, voxels~\cite{zhang2020falco}, or grids~\cite{rosmann2017kinodynamic} for constraint reduction. However, this would lead to degradation of solution accuracy.

\section{NeuPAN System Architecture}\label{section 4}

\begin{figure*}[t]
    \centering
    \includegraphics[width=0.99\textwidth]{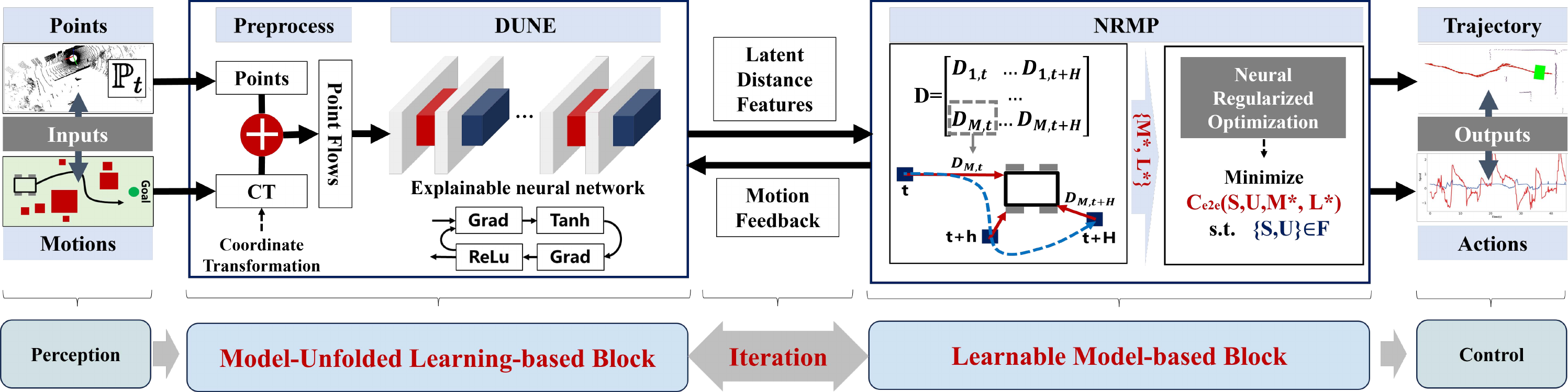}
    \caption{System architecture of NeuPAN, a perception-to-control end-to-end approach for navigation, consists of two main blocks: learning-based and model-based.}
    \label{fig:architecture}
\end{figure*}

To keep the most accurate, complete, and dense information about the environments, we propose to directly process $\mathbb{P}_t$, resulting in an end-to-end perception-to-control approach for unstructured and unknown environments. The system architecture is illustrated in Fig.~\ref{fig:architecture}. Below, we first introduce the mathematical interpretation behind NeuPAN. Then we provide the detailed descriptions of the system.

\subsection{Mathematical Interpretation}
    \emph{1) Strong Duality Transformation}: To overcome the nonconvexity and nondifferentiability of the collision avoidance constraints Eq.~\eqref{avoidance_label}, we transform the exact minimum distance computation problem Eq.~\eqref{dis_opt} to its dual problem under the strong duality property, as in~\cite{zhang2020optimization, han2023rda}:
    \begin{equation}
        \begin{aligned}
             & D_t^i =\mathop {\max }\limits_{\bm{\lambda}_t^i, \bm{\mu}_t^i}~~{\bm{\mu}_t^i}^\top ( \mathbf{G}{{\mathbf{\widetilde{p}}_t^i}(\mathbf{s}_t)} - \mathbf{h} )                                          \\
             & \text{ s.t. }~~{\bm{\mu}_t^i} \succeq _{\mathcal{K^*}} 0, ~ {\left\| \bm{\lambda}_t^i \right\|_*} \preceq 1, {\bm{\mu}_t^i}^\top\mathbf{G} + {\bm{\lambda}_t^i}^\top\mathbf{R}(\mathbf{s}_t)=\bm{0}.
            \label{distance_dual}
        \end{aligned}
    \end{equation}
    where $\mathcal{K^*}$ represents the dual cone of $\mathcal{K}$, and $\left\| \cdot \right\|_*$ denotes the dual norm. ${\mathbf{\widetilde{p}}_t^i}(\mathbf{s}_t)=\mathbf{R}_t(\mathbf{s}_t)^\top[\mathbf{p}_t^i - \mathbf{t}_t(\mathbf{s}_t)]$ represents the position of the obstacle point in the ego robot coordinate system. \(\mathcal{M} = \{\bm{\mu}_t^i \in \mathbb{R}^l\}\) and \(\mathcal{L} = \{\bm{\lambda}_t^i \in \mathbb{R}^n\}\) are introduced as the Lagrange multipliers associated with the inequality and equality constraints respectively. Each point \(\mathbf{\widetilde{p}}_t^i\) is associated with a pair of \(\bm{\mu}_t^i\) and \(\bm{\lambda}_t^i\).

\begin{figure}[t]
    \centering
    \includegraphics[width=0.49\textwidth]{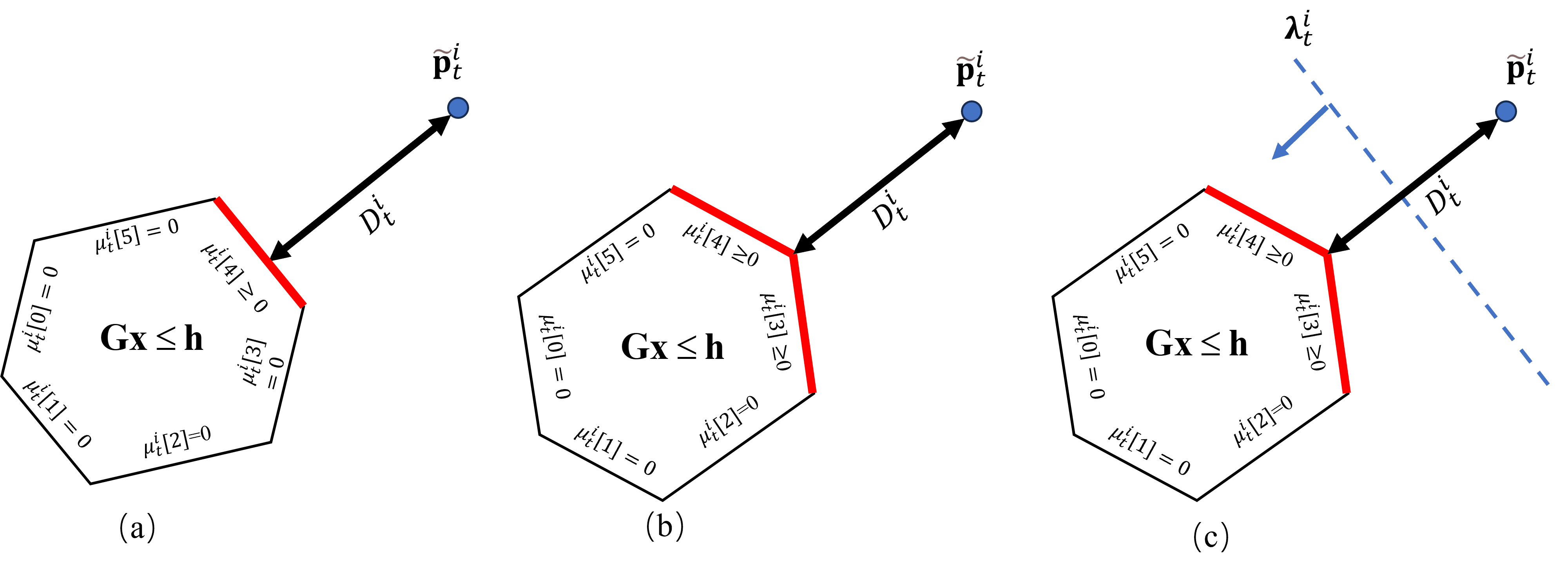}
    \caption{Geometric interpretation of LDFs \(\mathcal{M}\) and \(\mathcal{L}\). Red sides represent the closest robot edge(s) to the obstacle point. \(\bm{\lambda}_t^i\) represents the normal vector of the separation hyperplane. Elements of \(\bm{\mu}_t^i\) with positive value represent the collision related red sides.}
    \label{ldf}
\end{figure}

The geometric representation of \(\mathcal{M}\) and \(\mathcal{L}\) for a polygonal robot is shown in Fig.~\ref{ldf}. It can be seen that the minimum distance \(D_t^i\) is determined by the obstacle point \(\mathbf{\widetilde{p}}_t^i\) and the closest edge(s) of the ego robot (depicted as one or two red sides in the figure). For \(\bm{\mu}_t^i\), the elements corresponding to these red sides (related to collision) have positive values, while the other elements should be zero (unrelated to collision). For \(\bm{\lambda}_t^i\), it represents the normal vector of the separation hyperplane between the obstacle point \(\mathbf{\widetilde{p}}_t^i\) and the ego robot at time step \(t\). Intuitively, (\(\bm{\mu}_t^i\), \(\bm{\lambda}_t^i\)) represent the matching and ranging from each obstacle point to its nearest robot edge. This representation sparsifies the distance computations by pruning unmatched edges for each obstacle point. 
Based on the above intuiation, we define \(\mathcal{M}\) and \(\mathcal{L}\) as latent distance feature (LDF), and the set of all LDFs under the constraints in~Eq.~\eqref{distance_dual} can be denoted by \(\{\mathcal{M},\mathcal{L}\}\in \mathcal{G}\).

Leveraging the dual transformation and LDFs, we can reformulate the original problem $\mathsf{P}$ as an equivalent augmented dual form $\mathsf{Q}$ over $[\mathcal{S}, \mathcal{U}, \mathcal{M}, \mathcal{L}]$, which is a biconvex optimization problem:
    \begin{align}
        \mathsf{Q}:\min_{\substack{\{\mathcal{S},\mathcal{U}\}\in \mathcal{F}, \{\mathcal{M},\mathcal{L}\}\in \mathcal{G}}}~~ &
        \underbrace{C_0(\mathcal{S},\mathcal{U})+
            C_r(\mathcal{S},\mathcal{M},\mathcal{L})}_{:=C_{\mathrm{e2e}}(\mathcal{S},\mathcal{U},\mathcal{M},\mathcal{L})}
        \label{problem_q}
    \end{align}
    where
    \begin{equation}
        \begin{aligned}
            C_r(\mathcal{S},\mathcal{M},\mathcal{L}) & =\frac{\rho_1}{2} \sum_{h=t}^{t+H} \sum_{i=0}^{M}\left\|\min(I(\mathbf{s}_{h},{{\bm{{\mu}}}}_{h}^i,{\bm{\lambda}}_{h}^i), 0)\right\|_{2}^{2} \nonumber \\
                                                     & \quad
            +\frac{\rho_2}{2} \sum_{h=t}^{t+H} \sum_{i=0}^{M}\left\|E(\mathbf{s}_{h},{{\bm{{\mu}}}}_{h}^i,{\bm{\lambda}}_{h}^i)\right\|_{2}^{2}, \label{C_r}
        \end{aligned}
    \end{equation}
    and $\rho_1, \rho_2$ are penalty parameters. The penalty functions $I$ and $E$ are
    \begin{align}
         & I(\mathbf{s}_{h},{{\bm{{\mu}}}}_{h}^i,{\bm{\lambda}}_{h}^i)={\bm{\lambda}_h^i}^\top(\mathbf{t}_h(\mathbf{s}_h) - \mathbf{p}_h^i)
        -{\bm{{\mu}}_h^i} ^\top \mathbf{h}
        - d_{\mathrm{min}}, \label{penaltyI}
        \\
         & E(\mathbf{s}_{h},{{\bm{{\mu}}}}_{h}^i,{\bm{\lambda}}_{h}^i) = {\bm{{\mu}}_h^i}^\top\mathbf{G} + {\bm{\lambda}_h^i}^\top\mathbf{R}(\mathbf{s}_h).
        \label{penaltyH}
    \end{align}
    As shown in Appendix B (Supplementary Material), the optimal solution to $\mathsf{Q}$ is also optimal to $\mathsf{P}$.
    It can be seen that the bilevel collision avoidance constraints of $\mathsf{P}$ are transformed into biconvex distance costs. Intuitively, this transformation assigns a distance cost to these edge(s), which converts constraints into costs, facilitating subsequent problem decomposition.

\begin{figure*}[tb]
    \centering
    \includegraphics[width=0.90\textwidth]{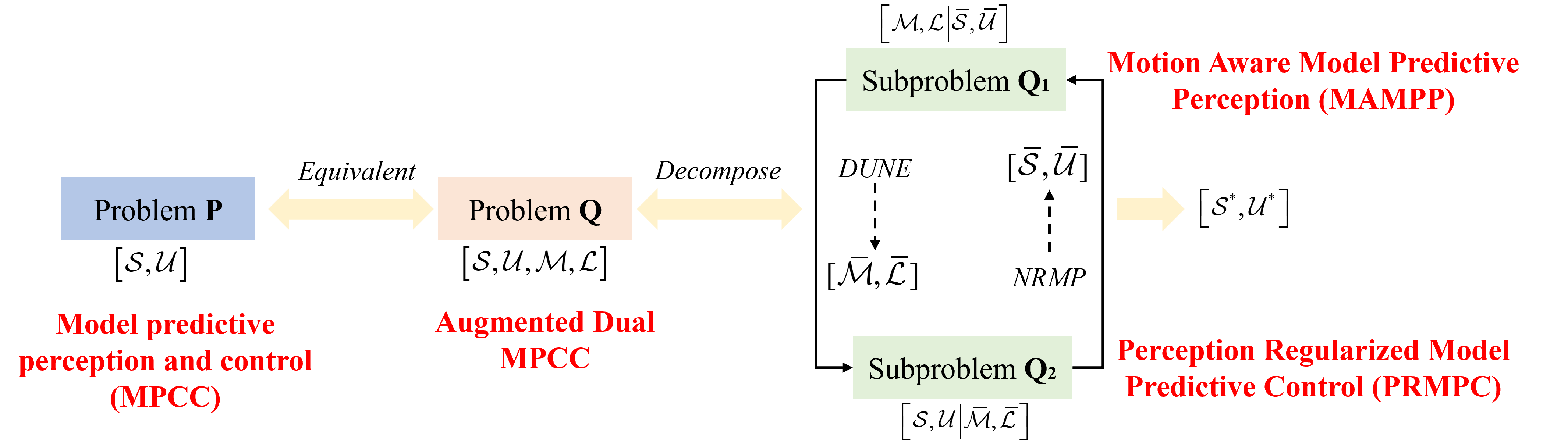}
    \caption{Relationship between problems $\mathsf{P}$, $\mathsf{Q}$, $\mathsf{Q_1}$, and $\mathsf{Q_2}$}
    \label{fig:relationship}
\end{figure*}

\emph{2) Problem Decomposition}: To facilitate the problem solving, $\mathsf{Q}$ is decomposed into two subproblems $\mathsf{Q_1}$ and $\mathsf{Q_2}$:
    \begin{itemize}
        \item $\mathsf{Q}_1$: Involving variables $\mathcal{M}$ and $\mathcal{L}$ while keeping $\mathcal{\overline{S}}$ and $\mathcal{\overline{U}}$ fixed, detailed in Section~\ref{section dune}.
        \item $\mathsf{Q}_2$: Involving variables $\mathcal{S}$ and $\mathcal{U}$ while keeping $\mathcal{\overline{M}}$ and $\mathcal{\overline{L}}$ fixed, detailed in Section~\ref{section nrmp}.
    \end{itemize}
These subproblems are solved alternately in each iteration. Specifically, as illustrated in Fig.~\ref{fig:relationship}, subproblem $\mathsf{Q}_1$ is a motion aware model predictive perception (MAMPP) problem to generate $[\mathcal{\overline{M}}, \mathcal{\overline{L}}]$ for $\mathsf{Q}_2$. Subproblem $\mathsf{Q}_2$ is a perception regularized model predictive control (PRMPC) problem to generate new $[\mathcal{\overline{S}}, \mathcal{\overline{U}}]$. This iterative process continues until convergence, yielding the final solution $[\mathcal{S}^*, \mathcal{U}^*]$. The convergence and complexity analysis of the proposed algorithm are provided in Section~\ref{section convergence}.
Intuitively, $\mathsf{Q}_1$ encodes the point cloud and robot shape into LDFs $[\mathcal{M}, \mathcal{L}]$, which is a high-dimension but easy-to-unfold problem. Meanwhile, $\mathsf{Q}_2$ maps the LDFs from $\mathsf{Q}_1$ to robot action associated with predictive trajectories, which is a low-dimension but complex optimization problem. This decomposition allows for efficient handling of the large number of direct point collision avoidance constraints by dividing the problem into more manageable components.
    
\subsection{NeuPAN System}\label{section_system}
As shown in Fig.~\ref{fig:architecture}, in each MPPC horizon, NeuPAN transforms raw points to point flow \(\widetilde{\mathbb{PF}}\) by the preprocess block, incorporating the motion feedback, then to LDFs by the learning-based block addressing problem $\mathsf{Q}_1$, and finally to robot control actions by the model-based block addressing $\mathsf{Q}_2$. The solution (i.e., containing robot actions and associated trajectories) generated by the model-based block is fed back to the preprocess block for re-generation of point flow, which forms a tightly-coupled closed-loop between perception and control.
Below we present the structure of each block.

\textbf{1) Model-unfolded learning-based block:} We first preprocess the inputs by transforming the coordinates of obstacle points \(\mathbb{P}_t\), from the global coordinate system to the robot's local coordinate system within the MPPC receding horizon framework. Given a set of obstacle points \(\mathbb{P}_t = \{ \mathbf{p}_t^1, \ldots, \mathbf{p}_t^M \}\) and their associated velocities \(\mathbb{V}_t = \{ \mathbf{v}_t^1, \ldots, \mathbf{v}_t^M \}\) at time \(t\), the point flow over the horizon \(H\) in the global coordinate system should be:
\begin{equation}
    \mathbb{PF}_t = \begin{bmatrix}
        \mathbf{p}_t^1     & \cdots & \mathbf{p}_t^M     \\
        \vdots             & \ddots & \vdots             \\
        \mathbf{p}_{t+H}^1 & \cdots & \mathbf{p}_{t+H}^M
    \end{bmatrix},
\end{equation}
where \(\mathbf{p}_{t+1}^i = \mathbf{p}_t^i + \mathbf{v}_t^i \Delta t\), \(i = 1, \ldots, M\), and \(\Delta t\) is the sample time. To transform this point flow \(\mathbb{PF}_t\) into the robot's local coordinate system \(\widetilde{\mathbb{PF}}_t\), we apply the rotation matrix \(\mathbf{R}(\mathbf{s}_t)\) and translation matrix \(\mathbf{t}(\mathbf{s}_t)\), calculated from the ego robot's state \(\mathbf{s}_t\) as introduced in Eq.~\eqref{robject}.
Thus, the point flow \(\widetilde{\mathbb{PF}}_t\) in the robot's local coordinate system should be:
\begin{equation}
    \widetilde{\mathbb{PF}}_t = \begin{bmatrix}
        \widetilde{\mathbf{p}}_t^1     & \cdots & \widetilde{\mathbf{p}}_t^M     \\
        \vdots                         & \ddots & \vdots                         \\
        \widetilde{\mathbf{p}}_{t+H}^1 & \cdots & \widetilde{\mathbf{p}}_{t+H}^M
    \end{bmatrix},
\end{equation}
where \(\widetilde{\mathbf{p}}_{h+1}^i = \mathbf{R}(\mathbf{s}_{h})^{-1} (\mathbf{p}_{h}^i - \mathbf{t}(\mathbf{s}_{h}))\), for \(h = t, \ldots, t+H\). This transformed point flow \(\widetilde{\mathbb{PF}}_t\) is then fed into the deep unfolded neural encoder (DUNE), which leverages a neural encoder to map the point flow \(\widetilde{\mathbb{PF}}_t\) to generate the LDFs \(\mathcal{M} = \{\bm{\mu}_t^i \in \mathbb{R}^l\}\) and \(\mathcal{L} = \{\bm{\lambda}_t^i \in \mathbb{R}^n\}\) with $\{\mathcal{M}, \mathcal{L}\}\in\mathcal{G}$. The encoder is denoted as deep unfolded neural encoder, since it can be interpreted as unfolding PIBCD into the DNN. Consequently, this network is both simple and fast, capable of handling thousands or more input points. This block is detailed in Section~\ref{section dune}.

\textbf{2) Learnable model-based block:}
The model-based block seamlessly incorporates the LDFs into a motion planning optimization problem as a rigorously-derived regularizer, $C_r(\mathcal{S},\mathcal{\overline{M}},\mathcal{\overline{L}})$, imposed on the loss function $C_0$, accounting for the reward of collision avoidance.
With this regularizer, we can safely remove the numerous point-level collision avoidance constraints, which is in contrast to conventional approximation or regularization methods~\cite{zhou2020ego, ding2021epsilon, han2023efficient} that may lead to approximation error.
The resultant problem is neural regularized, hence its associated solver is called neural regularized motion planner (NRMP). 
By leveraging functional differentiation, NRMP is also learnable, supporting auto-tuning of parameters in an end-to-end manner.
This block is detailed in Section~\ref{section nrmp}.

\textbf{3) End-to-End Algorithm:}
Our framework belongs to the generalized end-to-end approach because of the following reasons.
1) Our primal optimization problem is an end-to-end MPPC problem, which takes raw sensor data (i.e., points) as input and directly generates actions as output, aligning with typical end-to-end methods, such as those in~\cite{hu2023planning, liu2021efficient}, where system components are jointly optimized to achieve the final objective. 2) The MPPC problem is solved by intertwined optimization of motion aware model predictive perception (i.e., DUNE) and perception regularzied model predictive control (i.e. NRMP), which is derived from an equivalent transformation of the primal end-to-end MPPC problem. Intuitively, obstacle points are directly mapped to explainable LDFs to generate control commands, without involving any intermediate feature extraction step. 3) The system can be trained by end-to-end backpropagation to find the fine-tuned parameters in optimization block (detailed in Section~\ref{lon_fine_tuning}), aligning with typical end-to-end methods~\cite{hu2023planning, liu2021efficient}. This enables our approach to directly learn from rewards provided by the environments, thereby overcoming the issues of domain variations.

\section{NeuPAN Encoder and Planner Design}\label{section 5}

\subsection{Deep Unfolded Neural Encoder}\label{section dune}

This subsection presents DUNE, which corresponds to solving $\mathsf{Q}_1$ using, for example, $\{\overline{\mathcal{S}}=\mathcal{S}^{[k-1]},\overline{\mathcal{U}}=\mathcal{U}^{[k-1]}\}$ at the $k$-th iteration. The nature of this subproblem is to convert each point to its corresponding LDF.
The benefits of using such a transformation for the front-end are twofold: 
1) LDFs can be directly incorporated into the subsequent NRMP network; 2) Mapping from points to LDFs can be realized by interpretable neural networks, as shown later.
To explain how DUNE works, we will first derive the mathematical model of mapping from points to LDFs. Specifically, to guarantee collision avoidance at any time within the receding horizon of length $H$, it is necessary to compute the optimal $\{{\bm{\mu}_t^i}^*\}$ for any $t\in[t,t+H]$ and $i\in[0, M]$.
The inference mapping from points to LDFs for all points $\{i\}$ and time slots $\{t\}$ is thus to solve $M\times H$ problems in parallel, each having a similar formula as Eq.~\eqref{dual_penalty}.
This leads to a total computational complexity of $\mathcal{O}(MHl^{3.5})$ by using CVXPY ECOS~\cite{bib:Domahidi2013ecos} and $\mathcal{O}(MHl)$ by using a penalty inexact block coordinate descent (PIBCD) algorithm, making it impossible for real-time applications, when $M$ is in the range of thousands or more. To adress this problem, we further unfold PIBCD into an explainable DNN, and obtain the architecture of DUNE, which can be viewed as a neural accelerated version of PIBCD. 

\begin{figure*}[t]
    \centering
        \includegraphics[width=0.95\textwidth]{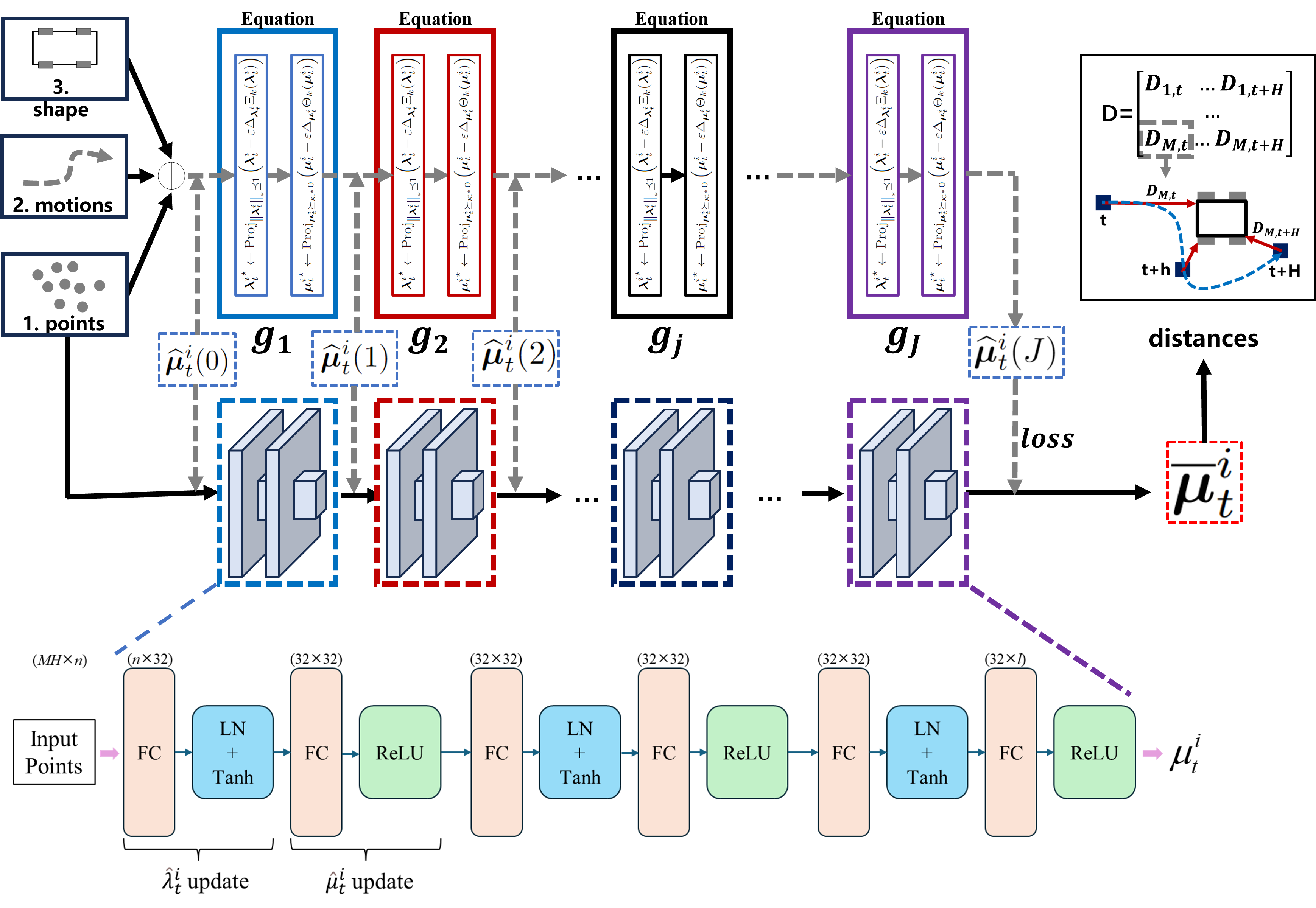}
        \caption{Structure of DUNE, which unfolds the PIBCD algorithm. DUNE is explainable, simple, and fast to train, and can be deployed across various scenarios without retraining, provided the robot's shape remains unchanged.}
        \label{fig:dnn}
\end{figure*}

\subsubsection{PIBCD for MAMPP Problem}
The MAMPP problem $\mathsf{Q}_1$ can be simplified and written as the penalizing form of Eq.~\eqref{distance_dual}:
\begin{equation}
    \begin{aligned}
    &\mathsf{Q}_1:
        \mathop {\min }\limits_{\bm{\mu}_t^i,\bm{\lambda}_t^i}~C_{\mathsf{DUNE}}(\bm{\mu}_t^i,\bm{\lambda}_t^i),~~\\
        &\text {s.t.}~{\bm{\mu}_t^i} \succeq _{\mathcal{K^*}} 0, ~ {\left\|\bm{\lambda}_t^i\right\|_*} \preceq 1,
        \label{dual_penalty}
    \end{aligned}
\end{equation}
in which 
\begin{align}
C_{\mathsf{DUNE}}(\bm{\mu}_t^i,\bm{\lambda}_t^i)
&=-{\bm{\mu}_t^i}^\top ( \mathbf{G}{{\mathbf{\widetilde{p}}_t^i}(\mathbf{\overline{s}}_t)} - \mathbf{h} )\nonumber\\
+ &\underbrace{\rho
        \|{\bm{\mu}_t^i}^\top\mathbf{G}
         + {\bm{\lambda}_t^i}^\top\mathbf{R}_t(\mathbf{\overline{s}}_t)\|_2^2}_{\text{penalty~term}},
\end{align}
where $\rho$ is a sufficiently large penalty parameter, e.g., $\rho=10^3$.
Due to strong convexity, Eq.~\eqref{dual_penalty} can be optimally solved via inexact block coordinate descent optimization, which updates $\bm{\mu}_t^i$ using gradient descent with $\bm{\lambda}_t^i$ fixed, and vice versa. 

At the $j$-th iteration of PIBCD, given the solution 
$\widehat{\bm{\mu}}_t^i(j-1)$ at the $(j-1)$-th iteration, the problem related to $\bm{\lambda}_t^i$ is 
\begin{equation}
    \begin{aligned}
        \mathop {\min }\limits_{\bm{\lambda}_t^i}~
 \Xi_j(\bm{\lambda}_t^i),~~\text { s.t. }~{\left\|\bm{\lambda}_t^i\right\|_*} \preceq 1,
        \label{problemlambda}
    \end{aligned}
\end{equation}
which is a quadratic minimization problem under norm ball constraints, with 
\begin{align}
 \Xi_j(\bm{\lambda}_t^i)={\| \widehat{\bm{\mu}}_t^i(j-1)}^\top\mathbf{G} + {\bm{\lambda}_t^i}^\top\mathbf{R}_t(\mathbf{\overline{s}}_t) \|_2^2. \nonumber
\end{align}
The associated one-step projected gradient descent update is
\begin{equation}
    \begin{aligned}
    &{\bm{\lambda}_t^i}^\star \leftarrow 
\mathrm{Proj}_{{\left\|\bm{\lambda}_t^i\right\|_*} \preceq 1}\left({\bm{\lambda}_t^i} - \varepsilon \Delta_{\bm{\lambda}_t^i}\Xi_j(\bm{\lambda}_t^i)\right),
        \label{gp2}
    \end{aligned}
\end{equation}
where $\varepsilon$ is the step-size. 
With the updated ${\bm{\lambda}_t^i}^\star$, the problem related to $\bm{\mu}_t^i$ is 
\begin{equation}
    \begin{aligned}
        \mathop {\min }\limits_{\bm{\mu}_t^i}~~&
 \Theta_j(\bm{\mu}_t^i)
        \\
        \quad\text { s.t. }~~&{\bm{\mu}_t^i} \succeq _{\mathcal{K^*}} 0,
        \label{problemmu}
    \end{aligned}
\end{equation}
which is a quadratic minimization problem under box constraints, with 
\begin{align}
\Theta_j(\bm{\mu}_t^i)
&=\rho
        \|{\bm{\mu}_t^i}^\top\mathbf{G} + {{\bm{\lambda}_t^i}^\star}^\top\mathbf{R}_t(\mathbf{\overline{s}}_t)\|_2^2 -{\bm{\mu}_t^i}^\top ( \mathbf{G}{{\mathbf{\widetilde{p}}_t^i}(\mathbf{\overline{s}}_t)} - \mathbf{h}). \nonumber
\end{align}
The associated one-step projected gradient descent update is
\begin{equation}
    \begin{aligned}
    &
{\bm{\mu}_t^i}^\star \leftarrow 
\mathrm{Proj}_{{\bm{\mu}_t^i} \succeq _{\mathcal{K^*}} 0}\left({\bm{\mu}_t^i} - \varepsilon \Delta_{\bm{\mu}_t^i}\Theta_j(\bm{\mu}_t^i)\right).
        \label{gp1}
    \end{aligned}
\end{equation}
This completes one iteration.
By setting $\widehat{\bm{\mu}}_t^i(j)={\bm{\mu}_t^i}^\star$, we can proceed to solve the problem for the $j$-th iteration. According to~\cite{yang2019inexact}, the sequence $\{{\widehat{\bm{\mu}}_t^i(0)},{\widehat{\bm{\mu}}_t^i(1)},\cdots \}$ generated by the above iterative procedure is convergent, and converges to the optimal solution of Eq.~\eqref{dual_penalty}.

To address the complexity issue while preserving interpretability, we propose a deep unfolding neural network to implement the point-LDF mapping. The detailed derivation is given below. 

\subsubsection{Deep Unfolding Architecture}
The sequence $\{{\widehat{\bm{\mu}}_t^i(0)},{\widehat{\bm{\mu}}_t^i(1)},\cdots \}$ generated by the PIBCD can be viewed as sequential mappings 
\begin{align}
&{\widehat{\bm{\mu}}_t^i(0)}\stackrel{g_1}{\longrightarrow}
{\widehat{\bm{\mu}}_t^i(1)}
\stackrel{g_2}{\longrightarrow}
{\widehat{\bm{\mu}}_t^i(2)}
\cdots
\stackrel{g_J}{\longrightarrow}
{\widehat{\bm{\mu}}_t^i(J)}
\end{align}
These mappings $\{g_1,g_2,\cdots\}$ are all gradient mappings. 
More specifically, $g$ only consists of gradients of quadratic functions, which are matrix multiplication operations.
Consequently, each $g_i$ can be safely unfolded into neural network layers that also correspond to matrix multiplications. 
The required number of iterations $J$ for PIBCD to converge determines the number of layers in the neural network. 
As such, deep unfolding takes a step towards interpretability by designing DNNs as learnt variations of iterative optimization algorithms.

Based on the above observations, the architecture of the proposed DUNE is shown in Fig.~\ref{fig:dnn}. 
The first layer is an $n\times 32$ fully connected layer that reads the positions of $M\times H$ points in a batch manner from the point flow. Following the first fully connected layer, Layer normalization (LN) and hyperbolic tangent function (Tanh) activation are implemented. This layer is obtained by unfolding Eq.~\eqref{gp2}, which is a one-step gradient descent update projected onto the $l_2$ norm ball. The second layer is an $32\times 32$ fully connected (FC) layer followed by Rectified Linear Unit (ReLU) activation. This layer is obtained by unfolding Eq.~\eqref{gp1}, which is a one-step gradient descent update projected on the positive semidefinite cone (in the case of polygonal robots). By alternately repeating the first and second layers with 32 units alternatively ($J=3$), the subsequent layers are constructed. Finally the output layer is a $32\times l$ FC layer to output $\widehat{\bm{\mu}}_t^i$, and the entire DUNE network is equivalent to unfolding the PIBCD method for solving $\mathsf{Q}_1$. This is because PIBCD is an algorithm that computes Eq.~\eqref{gp2} and Eq.~\eqref{gp1} alternatively, while DUNE is a network that computes unfolded layers of Eq.~\eqref{gp2} and Eq.~\eqref{gp1} alternatively. 
\footnote{$\widehat{\bm{\lambda}}_t^i$ can be derived from output $\widehat{\bm{\mu}}_t^i$ by the relationship in Eq.~\eqref{distance_dual}. $\widehat{\bm{\lambda}}_t^i=-\widehat{\bm{\mu}}_t^i\mathbf{G}^\top\mathbf{R}(\mathbf{s}_t)$.}

\subsubsection{Loss Function Design}  
The neural network is trained using a labeled dataset derived from the PIBCD algorithm. Backpropagation is determined based on a loss function between the optimal solution, $\{\widehat{\bm{\mu}}_t^i\}$, and the neural network's solution, $\{\overline{\bm{\mu}}_t^i\}$. 
A naive choice of the loss function is the mean square error (MSE) between these two values. 
However, since the learnt $\{\overline{\bm{\mu}}_t^i\}$ will be used in the subsequent motion planning as a regularizer for tackling the collision avoidance constraints, 
we must guarantee high accuracy of $\{\overline{\bm{\mu}}_t^i\}$ in different vector directions and incorporate these MSEs into the loss functions.
Based on these considerations, we formulate the loss function as follows:
\begin{equation}
    \begin{aligned}
        &\mathcal{L}(\hat{\bm{\mu}_i}, \overline{\bm{\mu}}_i) =  \|\bm{\widehat{\mu}_i}-\overline{\bm{\mu}}_i\|_2^2 + \|f_o(\bm{\widehat{\mu}_i})- f_o(\overline{\bm{\mu}}_i)\|_2^2 \\
        & \quad + \|f_a(\bm{\widehat{\mu}_i})-f_a(\overline{\bm{\mu}}_i)\|_2^2 + \|f_b(\bm{\widehat{\mu}_i})-f_b(\overline{\bm{\mu}}_i)\|_2^2, 
        \label{loss}
    \end{aligned}
\end{equation}
where, $f_o(\bm{\mu}_i)={\bm{\mu}_t^i}^\top ( \mathbf{G}{\mathbf{\widetilde{p}_t^i}} - \mathbf{h} )$ is the objective function of problem Eq.~\eqref{distance_dual}.
Functions $f_a(\bm{\mu}_i)=-{\bm{\mu}_i}^\top\mathbf{G}\mathbf{R}_t^\top$ and $f_b(\bm{\mu}_i)= {\bm{\mu}_i}^\top\mathbf{G}\mathbf{R}_t^\top{\mathbf{\widetilde{p}_t^i}} - {\bm{\mu}_i}^\top\mathbf{h}$ are associated with the neural regularizer function $C_r$, which will be introduced in Section~\ref{section nrmp}. 

\subsubsection{DUNE Training} 
The hyperparameters for training the DUNE model are listed in Table \ref{hyper}.
For a robot with the given value $[\mathbf{G}, \mathbf{h}]$ (decided by the shape size), the training process starts with generating $N_g$ point positions $\mathbf{\widetilde{p}}$ randomly within a specific range [$\mathbf{r}_l$, $\mathbf{r}_h$] in each axis. Subsequently, $N_g$ convex optimization problems respecting to the variable $\bm{\mu}$ are constructed by the known values of $[\mathbf{G}, \mathbf{h}, \mathbf{\widetilde{p}}^i]$ and solved by the PIBCD algorithm or CVXPY ECOS, resulting in $N_g$ optimal values ${\bm{\mu}^i}^*$. As such, each point position $\mathbf{\widetilde{p}}^i$ has a one-to-one corresponding optimal solution ${\bm{\mu}}^i$, resulting in a labeled dataset $\mathcal{T}=\{\mathbf{\widetilde{p}}^i,\widehat{\bm{\mu}}^i\}$ for neural network training.
The training process is conducted for $e$ epochs with a batch size of $B_n$. The Adam optimizer is leveraged to update the network parameters using a learning rate of $l_r$ and decay rate of $d_r$.
In our experiment, the training process is conducted on a desktop computer with an AMD Ryzen 9 CPU and an NVIDIA GeForce GTX 4090 GPU. The training process takes approximately an hour to complete. The process is listed in algorithm~\ref{dune_algorithm}.

Note that unlike typical deep learning models which require extensive data collection in real-world environments and may struggle with generalization due to the sim-to-real gap, our DUNE model is resilient to these challenges and can be trained rapidly because of its unique point representation and model-unfolded network structure. Furthermore, according to Eq.~\eqref{distance_dual}, the DUNE model is influenced solely by $[\mathbf{G}, \mathbf{h}]$, which are determined by the shape size. Consequently, for a specific robot, the DUNE model can be deployed across various real-world environments without the need for retraining.

\begin{table}[t]
    \centering
    \caption{Hyper-parameters used in DUNE training}
    \label{hyper}
    \scalebox{1.4}{
        \begin{tabular}{c|c|c|c}
            \hline
            Parameters                       & Value     & Parameters & Value \\ \hline
            $N_g$                            & 100000    & $l_r$      & 5e-5  \\ \hline
            [$\mathbf{r}_l$, $\mathbf{r}_h$] & [-25, 25] & $d_r$      & 0.5   \\ \hline
            $B_n$                            & 256       & $e$        & 5000  \\ \hline
        \end{tabular}
    }
\end{table}

\begin{algorithm}[t]
    \caption{Deep Unfolded Neural Encoder (DUNE)}
    \label{dune_algorithm}
    \textbf{Input:} Robot shape $\{\mathbf{G}, \mathbf{h}\}$ and randomly generated point set  ${\mathbf{\widetilde{p}}_t^i}(\mathbf{s}_t)$ within a specific range in the ego-robot coordinate system;\\
    \textbf{Initialize:} $\widehat{\bm{\mu}}_t^i(0)\succeq _{\mathcal{K^*}} 0$;\\
    \For{iteration $j=1, 2,\cdots,J$}
    { Compute ${\bm{\lambda}_t^i}^\star$ using Eq.~\eqref{gp2};\\
      Compute ${\bm{\mu}_t^i}^\star$ using Eq.~\eqref{gp1};\\
      Set $\widehat{\bm{\mu}}_t^i(j)={\bm{\mu}_t^i}^\star$ and $j\leftarrow j+1$.\\
      \If{convergence}{break}
    }
    Set training dataset $\mathcal{T}=\{\mathbf{\widetilde{p}}_t^i,\hat{\bm{\mu}}_t^i(J)\}$;\\  
    \For{epoch $e=1, 2,\cdots,E$}
    {
        Calculate the loss function Eq.~\eqref{loss};\\
        Back propagation using $\mathcal{T}$;\\
      \If{ loss converges }{break}
   }
   \textbf{Output:} DUNE model.
  \end{algorithm}

\subsection{Neural Regularized Motion Planner}\label{section nrmp}
This subsection presents NRMP, which corresponds to solving the PRMPC problem $\mathsf{Q}_2$ with fixed $\overline{\mathcal{M}}=\{{\overline{\bm{\mu}}}_t^i\}$ and $\overline{\mathcal{L}}=\{{\overline{\bm{\lambda}}}_t^i\}$ generated from the upstream DUNE. Owing to the efficient exact distance computation from DUNE, we can easily sort the importance of the obstacle points according to the distance. As such, we can reduce the computational complexity by only considering the $M'$ closest points for the regularizer in each time step. This leads to the following problem
\begin{align}
\mathsf{Q}_2:\min_{\substack{\{\mathcal{S},\mathcal{U}\} \in \mathcal{F}}}~
    &C_{0}(\mathcal{S},\mathcal{U})+\underbrace{C_{r}(\mathcal{S},\overline{\mathcal{M}},\overline{\mathcal{L}})}_{\text{neural~regularizer}}
    \nonumber\\
    &
    +\underbrace{\frac{b_k}{2} \sum_{h=t}^{t+H}\|\mathbf{s}_{h}-\mathbf{\overline{s}}_{h}\|_2^2}_{\text{proximal~term}},
\end{align}
where $\overline{\mathbf{s}}$ is short for $\mathbf{s}^{[k-1]}$ obtained from the last iteration of NeuPAN, and $b_k$ is the associated proximal coefficient at NeuPAN iteration $k$.
This subproblem aims to generate collision-free robot actions efficiently by LDF-integrated regularizers. 

\subsubsection{Neural Regularizer}
The regularization function based on Eq.~\eqref{problem_q} is
\begin{equation} 
\begin{aligned}
C_{r}(\mathcal{S},\overline{\mathcal{M}},\overline{\mathcal{L}})&=\frac{\rho_1}{2} \sum_{h=t}^{t+H} \sum_{i=0}^{M'}\left\|\min(I(\mathbf{s}_{h},{{\overline{\bm{{\mu}}}}}_{h}^i,{\overline{\bm{\lambda}}}_{h}^i), 0)\right\|_{2}^{2} \nonumber \\ 
  &\quad
  +\frac{\rho_2}{2} \sum_{h=t}^{t+H} \sum_{i=0}^{M'}\left\|E(\mathbf{s}_{h},{{\overline{\bm{{\mu}}}}}_{h}^i,{\overline{\bm{\lambda}}}_{h}^i)\right\|_{2}^{2}, \label{C_r}
  \end{aligned}
\end{equation}

In particular, $C_r$ has the following properties: 
\begin{itemize}
\item[(i)] $C_r$ is a nonlinear but low-dimensional convex function of $\{\mathbf{s}_t\}$. 
\item[(ii)] $C_r$ is a function of $\overline{\mathcal{M}}=\{{\overline{\bm{\mu}}}_t^i\}$ and $\overline{\mathcal{L}}=\{{\overline{\bm{\lambda}}}_t^i\}$ generated from the upstream neural encoder DUNE. That is why we call $C_r$ ``neural regularizer''. 
\item[(iii)] In practice, based on the relationship between $\bm{\lambda}_t^i$ and $\bm{\mu}_t^i$ guaranteed by DUNE, $\bm{\lambda}_t^i=-\bm{\mu}_t^i\mathbf{G}^\top\mathbf{R}(\mathbf{s}_t)$, $C_r$ can be simplified by removing the penalty function $E$ (set $\rho_2=0$)
\end{itemize}

Problem $\mathsf{Q}_2$ is a low-dimensional convex optimization program due to: convexity of norm function $C_0$ in Eq.~\eqref{cost0}, linearity of $\mathcal{F}$, and property (i) of $C_r$. 
Consequently, $\mathsf{Q}_2$ can be optimally solved via the off-the-shelf software for solving convex problems, e.g., cvxpy ECOS. In practice, a static safety distance $d_{\mathrm{min}}$ may not be suitable for time-varying environments. As a consequence, $d_{\mathrm{min}}$ in $\mathsf{Q}_2$ needs to be dynamically adjusted.
To achieve this goal, we propose to use a variable distance $d_t$ to replace $d_{\mathrm{min}}$ in penalty function $I$, where $d_t$ is within the range $[{d_{\min }},{d_{\max }}]$.
To encourage $d_t$ to approach $d_{\min}$ in highly-confined spaces and to approach $d_{\max}$ in highly-dynamic environments, 
we propose a sparsity-induced distance regularizer $C_{1}(\mathbf{d})=- \eta \sum_{h = t}^{t+H} \|\mathbf{d}_{h}\|_1$, where $\eta$ is a weighting factor. As such, $C_{1}(\mathbf{d})$ introduces adaptation to collision~\cite{han2023rda}.  
Consequently, this would involve manually tuning parameters $\mathcal{P}=\{\mathbf{q},\mathbf{p},d_{\mathrm{min}}, {d_{\max }}, \eta\}$ in $C_0$, $C_r$, and $C_1$. 
The next subsection will derive a learnable optimization method to fine tune $\mathcal{P}$ automatically. 

\subsubsection{Learnable Optimization Network}\label{ol}
The weight parameters in optimization problems are usually manually tuned, which is time-consuming and may not be optimal for challenging scenarios. In order to enable automatic calibration of parameters in $\mathsf{Q}_2$ under new environments, we present a \emph{learnable optimization network} (LON) approach for solving $\mathsf{Q}_2$, which can learn from failures to find the proper parameters. 
The crux of LON is that it leverages differentiable convex optimization layers (cvxpylayers)~\cite{agrawal2019differentiable}. 
As such, in contrast to traditional optimization solvers, LON offers the ability to differentiate through disciplined convex programs. 
Within these programs, parameters can be mapped directly to solutions, facilitating backpropagation calculations. 
Moreover, with cvxpylayers, the NRMP can be seamlessly integrated with the DUNE, as both of them support back propagation, thereby enabling end-to-end training of the entire NeuPAN system. 

However, solving $\mathsf{Q}_2$ using LON is nontrivial, since cvxpylayers require the optimization problem to satisfy disciplined parametrized program (DPP) form. 
To this end, this paper proposes the following problem reconfiguration that recasts $\mathsf{Q}_2$ into a DPP-friendly problem.
In particular, to satisfy the DPP formats in \cite{agrawal2019differentiable}, we reconfigure the non-DPP function $C_0$ by introducing a set of DPP parameters $\{^a\!\bm{\gamma}_t^i,^b\!\bm{\gamma}_t^i\}$, which gives
\begin{align}
    C_{0}^{\text{DPP}}(\mathcal{S},\mathcal{U}) = &\sum_{h=t}^{t+H}\Bigg(\left\|\mathbf{q}\circ{{\mathbf{s}_{h}} - ^a\!\bm{\gamma}_{h}^i}\right\|_2^2 +\left\|\mathbf{p}\circ\mathbf{u}_{h} - ^b\!\bm{\gamma}_{h}^i\right\|_2^2\Bigg),
    \label{cost0_dpp}
\end{align}
where $^a\!\bm{\gamma}_{h}^i\leftarrow\mathbf{q}\circ{\mathbf{s}_{h}}^\diamond$ and $ ^b\!\bm{\gamma}_{h}^i\leftarrow\mathbf{p}\circ\mathbf{u}_{h}^\diamond$ are introduced as DPP parameters for back propagation\footnote{The symbol $\leftarrow$ represents parameter substitution}.
On the other hand, we reformulate $C_r$ in $\mathsf{Q}_2$ as 
\begin{align}
     C_{r}^{\text{DPP}}(\mathcal{S},\mathcal{U},\mathcal{L})&=\frac{\rho_1}{2} \sum_{h=0}^{H-1} \sum_{i=0}^{M}\left\|\min(I^{\text{DPP}}(\mathbf{s}_{h},{{\bm{{\mu}}}}_{h}^i,{\bm{\lambda}}_{h}^i), 0)\right\|_{2}^{2} \nonumber \\ 
  &
  +\frac{\rho_2}{2} \sum_{h=0}^{H-1} \sum_{i=0}^{M}\left\|E^{\text{DPP}}(\mathbf{s}_{h},{{\bm{{\mu}}}}_{h}^i,{\bm{\lambda}}_{h}^i)\right\|_{2}^{2},
\end{align}
where the DPP-reformulated $I$ and $E$ are
\begin{align}
    I^{\text{DPP}} &= ^c\!\bm{\gamma}_t^i \mathbf{t}_t(\mathbf{s}_t) - ^a\!\xi_t^i - d_{t}, \\
    E^{\text{DPP}} &= ^d\!\bm{\gamma}_t^i +  ^c\!\bm{\gamma}_t^i\mathbf{R}_t(\mathbf{s}_t),
\end{align}
with vector $^c\!\bm{\gamma}_t^i\leftarrow{\bm{\lambda}_t^i}^\top, ^d\!\bm{\gamma}_t^i\leftarrow{\bm{\mu}_t^i}^\top\mathbf{G}$, 
and scalar $^a\!\xi_t^i\leftarrow{\bm{\lambda}_t^i}^\top\mathbf{p}_t^i + {\bm{\mu}_t^i} ^\top \mathbf{h}$. 
Thus, the DPP parameters are $\mathcal{P}=\{\mathbf{q}, \mathbf{p}, d_{\mathrm{min}}, d_{\mathrm{max}}, \eta, \mathbf{\overline{s}}_{h}, 
    ^a\!\bm{\gamma}_t^i,^b\!\bm{\gamma}_t^i,
    ^c\!\bm{\gamma}_t^i,^d\!\bm{\gamma}_t^i,
    ^a\!\xi_t^i\}.$
Considering the impact on the generated behavior, we selected $\mathcal{P'}=\{\mathbf{q}, \mathbf{p}, d_{\mathrm{min}}, d_{\mathrm{max}}, \eta\}.$  as the learnable parameters in LON.

With the above reconfiguration, problem $\mathsf{Q}_2$ becomes
\begin{align}
\mathsf{Q}_2^{\text{DPP}}:\min_{\substack{\{\mathcal{S},\mathcal{U}\} \in \mathcal{F}}}~
    &C_{0}^{\text{DPP}}(\mathcal{S},\mathcal{U})+C_{r}^{\text{DPP}}(\mathcal{S},\overline{\mathcal{M}},\overline{\mathcal{L}}) + C_{1}(\mathbf{d})
    \nonumber\\
    &
    +\frac{b_k}{2} \sum_{h=0}^{H-1}\|\mathbf{s}_{h}-\mathbf{\overline{s}}_{h}\|_2^2
    . \label{cvxpy_layers}
\end{align}
It can be proved that both $C_{0}^{\text{DPP}}$, $C_{r}^{\text{DPP}}$, and $C_{1}(\mathbf{d})$ satisfy the DPP regulation. 
Adding to the fact that other functions in set $\mathcal{F}$ are DPP-friendly, 
problem $\mathsf{Q}_2^{\text{DPP}}$ is DPP compatible, whose parameters can be optimized using the back propagation. 
Specifically, a parameter, say $x\in\mathcal{P'}$, can be revised utilizing the gradient descent method, facilitated by a loss function $L(\mathcal{P})$, as follows:
\begin{equation}
    \begin{aligned}
        &x^{(k+1)} = x^{(k)} - \alpha \frac{\partial {L}}{\partial x} 
    \end{aligned}
\end{equation}
where, $\alpha$ is the learning rate. 
The gradient is clipped within a range $[g_\text{min}, g_\text{max}]$, with the maximum value of $1.0$ to avoid the gradient explosion.  

\subsubsection{Loss Function}
The loss function $L(\mathcal{P})$ is designed based on the cost function $C_0, C_r$, and $C_{1}$ presented in $\mathsf{Q}_2^{\text{DPP}}$ and can be selected depending on a specific task. Particularly, the final loss function could be the combination of several atom loss functions: $L(P')=\sum_{i=0}^{}a_i L_i(P')$, where $a_i$ is the weight of the $i$-th loss function.  
The key idea is to allow the LON to learn from failures to update the learnable parameters. Here, we provide three examples of the atom loss functions:
\begin{equation}
    \left\{
    \begin{aligned}
        L_1(\mathcal{P'})&=\|{\mathbf{s}}- \mathbf{s}^\diamond \|_2^2, \nonumber\\
        L_2(\mathcal{P'})&=\|{\mathbf{u}}- \mathbf{u}^\diamond \|_2^2 \nonumber\\
        L_3(\mathcal{P'})&= - \eta \sum_{h = t}^{t+H} \|\mathbf{d}_{h}\|_1, 
        \label{learn}
    \end{aligned}
    \right.
    \end{equation}
\begin{itemize}
    \item If the robot strays from the destination, the $L_1(\mathcal{P'})$ function can be used to yield a larger $\mathbf{q}$, and encourage the robot to return to the naive path. 
    \item If the robot gets stuck in a congested environment, the $L_2(\mathcal{P'})$ and $L_3(\mathcal{P'})$ can be combined to yield a larger $\mathbf{p}$ but a smaller $d_{\mathrm{max}}$ which creates an incentive for robot motion. 
    \item If the robot collides with obstacles, the $L_1(\mathcal{P'})$ and $L_3(\mathcal{P'})$ can be combined to yield a smaller $\mathbf{q}$, allowing the robot to temporarily leave the naive path, and a larger $d_{\mathrm{max}}$ producing more conservative movements. 
\end{itemize}

The entire procedure of NRMP is summarized in Algorithm~\ref{diff_alg}. 

\begin{algorithm}[t]
    \caption{DPP-friendly NRMP}
    \label{diff_alg}
    \textbf{Input:} LDFs $\{\mathcal{\overline{M}},\mathcal{\overline{L}}\}$ from DUNE;\\
    \qquad  \quad  Robot Geometry information $[\mathbf{G}, \mathbf{h}]$; \\
    Construct parametrized convex optimization problem; \\
    Initialize the learnable parameters $\mathcal{P'}=\{\mathbf{q}, \mathbf{p}, d_{\mathrm{min}}, d_{\mathrm{max}}, \eta\}.$\\
    Solve the problem Eq.~\eqref{cvxpy_layers} via DPP solver; \\
    \If{ failures returns True }{Update NRMP via back propagation by loss function $L(P')$}
    \textbf{Output:} State and action sequence $[\mathcal{S}, \mathcal{U}]$.
  \end{algorithm}

  \begin{algorithm}[t]
    \caption{ NeuPAN}
    \label{NeuPAN_alg}
    \textbf{Input:} Robot geometry information $[\mathbf{G}, \mathbf{h}]$; \\
    \qquad  \quad   Current robot states $\mathbf{ \overline{s}}_t$; \\
    \qquad  \quad   Current point cloud $\mathbb{P}_t$; \\
    \qquad  \quad   Well-trained neural encoder DUNE;\\
    \qquad  \quad   Parametrized optimization network NRMP;\\
    \qquad  \quad   Naive path $\{s_\text{start}, \mathbf{s}_0^\diamond, \mathbf{s}_1^\diamond, \cdots, s_\text{goal}\}$

    Initialize $\mathcal{S}^{[0]}$ and $\mathcal{U}^{[0]}$.

    \For{iteration $k=0,1, 2,\cdots$}
    {
    Generate point flow $\widetilde{\mathbb{PF}}$ from the current scan $\mathbb{P}_t$ and robot motion $\mathcal{S}^{[k]}$;\\
    Generate the latent distance features $\{\overline{\mathcal{M}},\overline{\mathcal{L}}\}$ via DUNE;\\
    Generate the state-action variables $\{\overline{\mathcal{S}},\overline{\mathcal{U}}\}$ via NRMP;\\
    Update $\{\mathcal{M}^{[k+1]}=\overline{\mathcal{M}}, \mathcal{L}^{[k+1]}=\overline{\mathcal{L}}\}$; \\
    Update $\{\mathcal{S}^{[k+1]}=\overline{\mathcal{S}}, \mathcal{U}^{[k+1]}=\overline{\mathcal{U}}\}$;
    \\

    \If{stopping criteria are satisfied,}
    {
        break;\\
    }
    Update $k\leftarrow k+1$;
    }
    \textbf{Output:} Control vector and associated trajectory $\{\mathcal{U}, \mathcal{S}\}$
\end{algorithm}

\subsection{Convergence and Complexity Analysis} \label{section convergence}
  The entire NeuPAN procedure is summarized in Algorithm~\ref{NeuPAN_alg}, which can be viewed as iterations between learning-based (i.e., DUNE) and model-based (i.e., NRMP) algorithms. 
  Specifically, given $\{\mathcal{S}^{[k]},\mathcal{U}^{[k]}\}$ of state-action space at the $k$-th iteration, NeuPAN generates the next-round solution $\{\mathcal{S}^{[k+1]},\mathcal{U}^{[k+1]}\}$ as follows:
  \begin{align}
  &\{\mathcal{S}^{[k]},\mathcal{U}^{[k]}\}
  \stackrel{\mathsf{DUNE}}{\longrightarrow}
  \{\mathcal{M}^{[k+1]},\mathcal{L}^{[k+1]}\}
  \stackrel{\mathsf{NRMP}}{\longrightarrow}
  \{\mathcal{S}^{[k+1]},\mathcal{U}^{[k+1]}\}.
  \nonumber
  \end{align}
  Consequently, starting from an initial guess $\{\mathcal{S}^{[0]},\mathcal{U}^{[0]}\}$, the sequence generated by NeuPAN is 
  \begin{align}
  \{\mathcal{S}^{[0]},\mathcal{U}^{[0]}\}
  &\stackrel{\mathsf{DUNE}}{\longrightarrow}
  \{\mathcal{M}^{[1]},\mathcal{L}^{[1]}\}
  \stackrel{\mathsf{NRMP}}{\longrightarrow}
  \{\mathcal{S}^{[1]},\mathcal{U}^{[1]}\}
  \nonumber\\
  &
  \stackrel{\mathsf{DUNE}}{\longrightarrow}
  \{\mathcal{M}^{[2]},\mathcal{L}^{[2]}\}
  \stackrel{\mathsf{NRMP}}{\longrightarrow}
  \{\mathcal{S}^{[2]},\mathcal{U}^{[2]}\}
  \nonumber\\
  &
  \stackrel{\mathsf{DUNE}}{\longrightarrow}
  \{\mathcal{M}^{[3]},\mathcal{L}^{[3]}\}
  \cdots \label{sequence}
  \end{align}
  Such model-based deep learning framework can be trained via end-to-end back propagation. It can naturally exploit additional training data by adding neurons to DUNE and prior knowledge by adding constraints to NRMP.

  To see why NeuPAN works, the following theorem is established.

\begin{theorem}
The sequence $\left[\{\mathcal{S}^{[0]},\mathcal{U}^{[0]}\},\{\mathcal{S}^{[1]},\mathcal{U}^{[1]}\},\cdots\right]$ satisfies the following conditions:

\noindent(i) Monotonicity: $
C_{\mathrm{e2e}}^{[0]}\geq C_{\mathrm{e2e}}^{[1]}\geq C_{\mathrm{e2e}}^{[2]}
\geq\cdots,
$
where $C_{\mathrm{e2e}}^{[k]}=C_{\mathrm{e2e}}(\mathcal{S}^{[k]},\mathcal{U}^{[k]},\mathcal{M}^{[k]},\mathcal{L}^{[k]})$.

\noindent(ii) Convergence: $\|C_{\mathrm{e2e}}^{[k+1]}-C_{\mathrm{e2e}}^{[k]}\|_2\rightarrow 0$ as $k\rightarrow \infty$.

\noindent(iii) Convergence to a critical point of $\mathsf{Q}$.

\end{theorem}
\begin{proof}
See Appendix C (Supplementary Material). 
\end{proof}

Part (i) of \textbf{Theorem 1} states that adding the tightly-coupled feedback from the NRMP (i.e., motion) to the DUNE (i.e., perception) is guaranteed to improve the performance of the NeuPAN system (under the criterion of solving $\mathsf{P}$). 
Part (ii) of \textbf{Theorem 1} means that we can safely iterate the loop, since the outputs converge and do not explode. 
Part (iii) of \textbf{Theorem 1} implies that the states and actions generated by NeuPAN must be at least a local optimal solution to $\mathsf{Q}$, which is an equivalent problem to $\mathsf{P}$.
All the above insights indicate that the proposed algorithm has end-to-end mathematical guarantee, which is in contrast to all existing approaches.

Based on \textbf{Theorem 1}, we can terminate NeuPAN until convergence. In practice, to save the computation time and enable real-time end-to-end robot navigation, we terminate the NeuPAN iteration when the number of iterations reach the limit, e.g., $K=3$.

Finally, we present the complexity analysis of NeuPAN. Depending on the procedure listed in Algorithm~\ref{NeuPAN_alg}, in each iteration, the scan flow is first generated from the scan data with a complexity of $\mathcal{O}(MH)$. Subsequently, DUNE consists of multiple neural layers at a computational cost of $\mathcal{O}(MHN_n)$, where $N_n$ is the number of neurons in the DUNE neural network. Next is the NRMP, which solves a DPP based optimization problem with a complexity of $\mathcal{O}(H(n + 2))^{3.5}$. In summary, with the number of iterations required to converge being $K$, the total complexity of NeuPAN is $\mathcal{O}(K(MH(N_n+1)+ (H(n + 2))^{3.5}))$. It can be seen that the complexity is linear with $M$ which corroborates the facts that NeuPAN can tackle thousands of points in real-time.

\section{Experiments}\label{section 6}

In this section, we present numerical results in an open source lightweight robot simulator, $\textit{Ir-sim}$~\cite{han2026ir}, to analyze the effectiveness and efficiency of NeuPAN.
To further demonstrate the effectiveness and robustness of NeuPAN under practical settings, we also evaluate the performance of NeuPAN on different robot platforms in high-fidelity simulation environments and real-world test tracks.
 
\begin{figure}[t]
  \centering
  \includegraphics[width=0.48\textwidth]{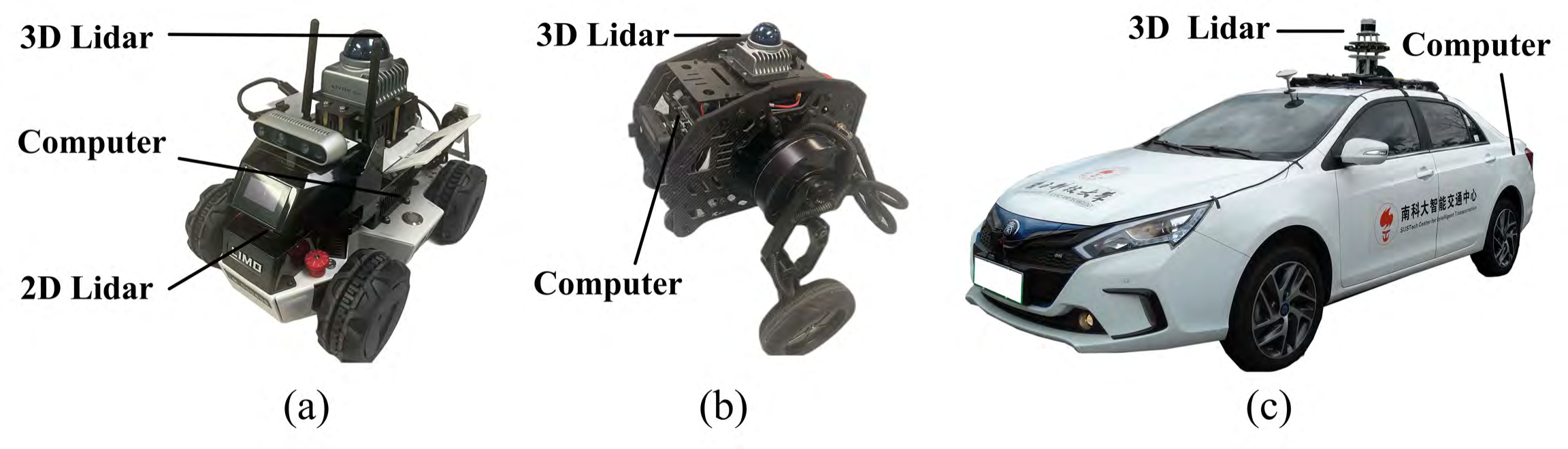}
  \caption{ Three types of robot platforms used in our experiments: (a) ground mobile robot; (b) wheel-legged robot; (c) autonomous driving vehicle. }
  \label{platform}
\end{figure}

The adopted robot platforms include ground mobile robot, wheel-legged robot, and passenger autonomous vehicle, as depicted in Fig.~\ref{platform}.
\begin{itemize}
    \item The ground mobile robot shown in Fig.~\ref{platform}(a) is a customized, multi-modal, small-size robot platform, whose motion kinematics can be switched between the differential and Ackermann modes.
    \item The wheel-legged robot shown in Fig.~\ref{platform}(b) is a mid-size platform that integrates the functionalities of both wheeled and legged robots, thus enjoying the joint benefits of high-mobility and terrain-adaptability in complex scenarios.
    Compared to the ground robot in Fig.~\ref{platform}(a), wheel-legged robots involve higher motion uncertainties (e.g., body oscillation), and thus requires more precise control to ensure stability.
    \item The passenger autonomous vehicle shown in Fig.~\ref{platform}(c) is a $4.675\,\mathrm{m}\times1.77\,\mathrm{m}\times1.5\,\mathrm{m}$ large-size vehicle platform for urban driving.
\end{itemize}
All these platforms are equipped with lidar systems (2D or 3D), enabling them to obtain the point representation of the environment.
Localization of these robots in unknown environments is realized by Fast-lio2~\cite{xu2022fast} or Lego-Loam~\cite{shan2018lego, gkim-2018-iros}.
All the experiments are conducted without any prior map, relying solely on goal positions, desired speed, and onboard lidar for autonomous navigation.
Note that in real-world experiments, the moving obstacles (e.g. humans) have low speed (less than $1\,$m/s), and we treat these obstacle points from laser scans as fixed during each MPPC horizon.\footnote{
In such a case, the movements of obstacles within a short MPPC horizon (i.e., tens of milliseconds) is negligible, and treating obstacles as fixed objects does not impact the system performance. 
Furthermore, our high control frequency enables the robot to perform agile behaviors by reacting to changes in the positions of low-speed obstacles.
}

We adopt the following performance metrics for evaluations.
\begin{itemize}
  \item \textbf{Success rate:} Success is defined as the robot completing the navigation task without any collision. This includes two conditions, i.e., route completion and collision avoidance. Route completion is considered a failure if the navigation time exceeds a predefined threshold (indicating that the robot gets stuck). Meanwhile, any collision between the robot and obstacles is also considered a failure. A similar definition of success is presented in~\cite{chen2019crowd}. The success rate is the ratio of successful cases out of the total number of test trials.
  \item \textbf{Navigation time:} Navigation time refers to the amount of time for the robot to successfully complete the navigation task. It is measured by counting the time steps in the simulator or recording the timestamps in real-world experiments.
  \item \textbf{Average speed:} Average speed is calculated over the entire navigation process, as recorded by the simulator or the odometry in real-world experiments.
\end{itemize}
Higher success rate indicates better collision avoidance ability and effectiveness. Shorter navigation time and higher average speed indicate higher mobility and efficiency.
To quantify the level of navigation difficulty in cluttered environments, we define a metric termed the \textbf{Degree of Narrowness ($\mathsf{DoN}$)}, as follows:
\begin{align}
  \mathsf{DoN} = \frac{\mathrm{robot~width}}{\mathrm{minimum~passable~space~width}}
\end{align}
Higher $\mathsf{DoN}$ (closer to 1) means a narrower space and a higher difficulty for robot pass through; and vice versa.

\begin{figure*}[t]
  \centering
  \begin{subfigure}[t]{0.32\textwidth}
      \includegraphics[width=0.96\textwidth]{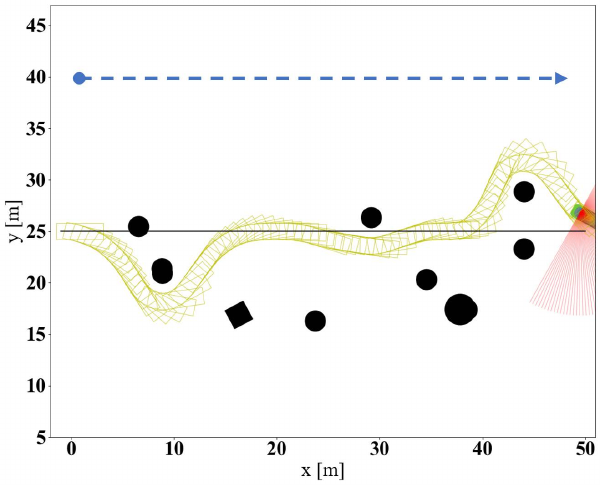}
      \caption{ Convex.}
      \label{convex_obs}
  \end{subfigure}
  \hfill
  \begin{subfigure}[t]{0.32\textwidth}
      \includegraphics[width=0.96\textwidth]{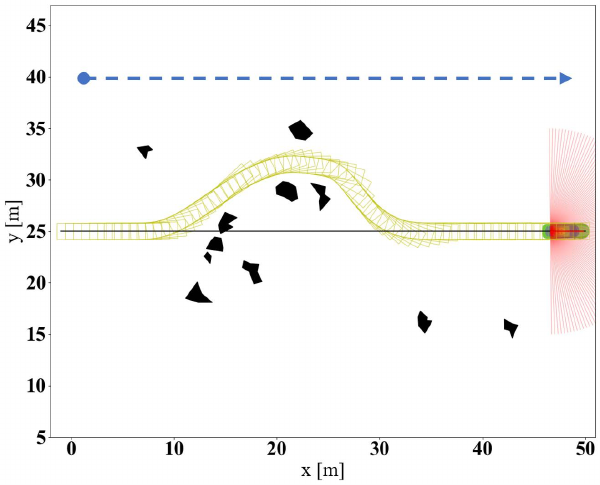}
      \caption{ Nonconvex.}
      \label{nonconvex_obs}
  \end{subfigure}
  \hfill
  \begin{subfigure}[t]{0.32\textwidth}
      \includegraphics[width=0.96\textwidth]{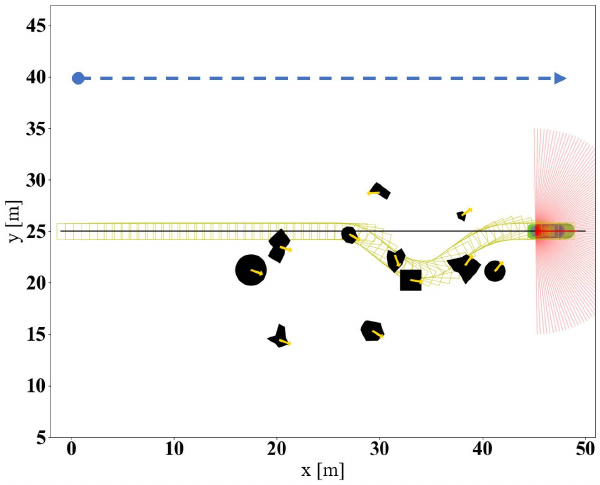}
      \caption{Dynamic.}
      \label{dynamic_numerical_scenarios}
  \end{subfigure}
  \caption{Scenarios with randomly generated obstacles used to evaluate NeuPAN and RDA.}
  \label{numerical_scenarios}
\end{figure*}

\begin{table*}[t]
  \caption{ Performance comparison of NeuPAN and RDA in the Ir-sim simulation}
  \label{irsim}
  \centering
  \resizebox{0.95\textwidth}{!}{%
      \begin{tabular}{cccccccccc}
          \hline
          \multirow{2}{*}{Kinematics}   & \multirow{2}{*}{Scenarios} & \multicolumn{2}{c}{Success Rate} &             & \multicolumn{2}{c}{Average Navigation Time} &                 & \multicolumn{2}{c}{Average Speed}                           \\ \cline{3-4} \cline{6-7} \cline{9-10}
                                        &                            & Neupan                           & RDA         &                                             & Neupan          & RDA                               &  & Neupan        & RDA  \\ \hline
          \multirow{3}{*}{Ackermann}    & Convex Obstacles                    & 86                               & \textbf{90} &                                             & \textbf{136.99} & 145.87                            &  & \textbf{3.80}  & 3.74 \\
                                        & Nonconvex Obstacles                 & \textbf{73}                      & 59          &                                             & \textbf{134.93} & 146.20                             &  & \textbf{3.82} & 3.73 \\
                                        & Dynamic Obstacles          & \textbf{90}                      & 88          &                                             & \textbf{130.49} & 160.42                            &  & \textbf{3.97} & 3.56 \\ \hline
          \multirow{3}{*}{Differential} & Convex Obstacles                     & \textbf{97}                      & 97          &                                             & \textbf{135.18} & 146.75                            &  & \textbf{3.97} & 3.76 \\
                                        & Nonconvex Obstacles                  & \textbf{82}                      & 71          &                                             & \textbf{133.11} & 151.34                            &  & \textbf{3.97} & 3.71 \\
                                        & Dynamic Obstacles          & \textbf{92}                      & 88          &                                             & \textbf{130.51} & 153.95                            &  & \textbf{3.97} & 3.68 \\ \hline
      \end{tabular}%
  }
\end{table*}

\subsection{Experiment 1: Verification of NeuPAN in Ir-sim} 
\subsubsection{Comparison with RDA}
We test and compare the performance of NeuPAN and RDA~\cite{han2023rda} using differential and Ackermann robots in three different scenarios composed of $11$ randomly generated convex, nonconvex, and dynamic obstacles (speed: 1\,m/s). 
To ensure fair comparisons between NeuPAN and RDA, the robots equipped with 2D lidar are given the same start (i.e., [-1, 25]) and goal (i.e., [50, 25]) positions, with a desired speed of $4\,$m/s, as shown in Fig.~\ref{numerical_scenarios}. 
Quantitative comparison results over $100$ trials are listed in Table~\ref{irsim}. We evaluate the success rate, average navigation time, and average speed for both NeuPAN and RDA.
The results demonstrate that when handling the convex obstacles, the success rates of RDA and NeuPAN are close to each other under both differential and Ackermann kinematics.
However, when handling nonconvex obstacles, RDA achieves the success rates of $59\%$ and $71\%$, while NeuPAN achieves the success rates of $73\%$ and $82\%$, under Ackermann and differential kinematics, respectively. 
This indicates that NeuPAN outperforms RDA by over $19\%$ in terms of the success rate for nonconvex obstacles. 
This result demonstrates the importance of directly mapping raw points to actions via NeuPAN. 
This also quantitatively illustrates how much navigation performances could be improved by representing obstacles as dense points.
In addition, NeuPAN achieves higher moving speed ($6.0\%$ improvement) and less navigation time ($11.27\%$ reduction) in all the simulated scenarios as shown in Table~\ref{irsim}, which verifies the high efficiency of NeuPAN.
More demonstrations in Ir-sim can be found in Appendix D (Supplementary Material).

\subsubsection{Impact of Obstacle Velocity}
To show the effectiveness of our approach in handling moving obstacles with known point velocities, we compare the performance of NeuPAN and NeuPAN-vel (incorporating point velocities) in the dynamic scenario shown in Fig.~\ref{dynamic_numerical_scenarios}.
We consider four different scenarios, with the velocities of obstacles ranging from $1$\,m/s to $4$\,m/s, and evaluate the performance metrics of success rate, average navigation time, and average speed. 
All the results in Table.~\ref{dynamic_obstacle} are obtained by averaging 100 trials.
It can be seen that the performance of NeuPAN and NeuPAN-vel is comparable in lower-speed scenarios (e.g., obstacles with velocities $1$\,m/s and $2$\,m/s). 
However, in the high-speed scenarios (e.g., obstacles with velocities $3$\,m/s and $4$\,m/s), NeuPAN-vel achieves higher success rates (with $10.96\%$ and $35.42\%$ improvements in the $3$\,m/s and $4$\,m/s cases, respectively) and shorter navigation time (with $1.09\%$ and $3.47\%$ reductions in the $3$\,m/s and $4$\,m/s cases, respectively), which demonstrates the effectiveness of incorporating point velocities into NeuPAN for handling moving obstacles. 

\begin{table*}[t]
  \caption{Performance comparison of NeuPAN and NeuPAN-vel in the Ir-sim simulation.}
  \label{dynamic_obstacle}
  \centering
  \resizebox{0.95\textwidth}{!}{%
      \begin{tabular}{cccccccccc}
          \hline
          \multirow{2}{*}{\begin{tabular}[c]{@{}c@{}}Dynamic Obstacle\\ Speed (m/s) \end{tabular} } & \multicolumn{2}{c}{Success Rate} &               & \multicolumn{2}{c}{Average Navigation Time} &                 & \multicolumn{2}{c}{Average Speed}                                    \\ \cline{2-3} \cline{5-6} \cline{8-9}
                                                                                                    & Neupan                           & NeuPAN-vel    &                                             & Neupan          & NeuPAN-vel                        &  & Neupan        & NeuPAN-vel    \\ \hline
          1                                                                                         & 0.92                             & \textbf{0.93} &                                             & 129.02          & \textbf{128.83}                   &  & 3.97          & 3.97          \\
          2                                                                                         & \textbf{0.90}                    & 0.89          &                                             & \textbf{131.10} & 131.55                            &  & \textbf{3.97} & 3.96          \\
          3                                                                                         & 0.73                             & \textbf{0.81} &                                             & 139.78          & \textbf{138.26}                   &  & 3.96          & 3.96          \\
          4                                                                                         & 0.48                             & \textbf{0.65} &                                             & 147.98          & \textbf{142.85}                   &  & 3.96          & \textbf{3.99} \\\hline
      \end{tabular}%
  }
\end{table*}

\subsubsection{Verification of Fine-Tuning}\label{lon_fine_tuning}
\begin{figure}[t]
  \centering
  \begin{subfigure}[t]{0.48\textwidth}
      \centering
      \includegraphics[width=0.99\textwidth]{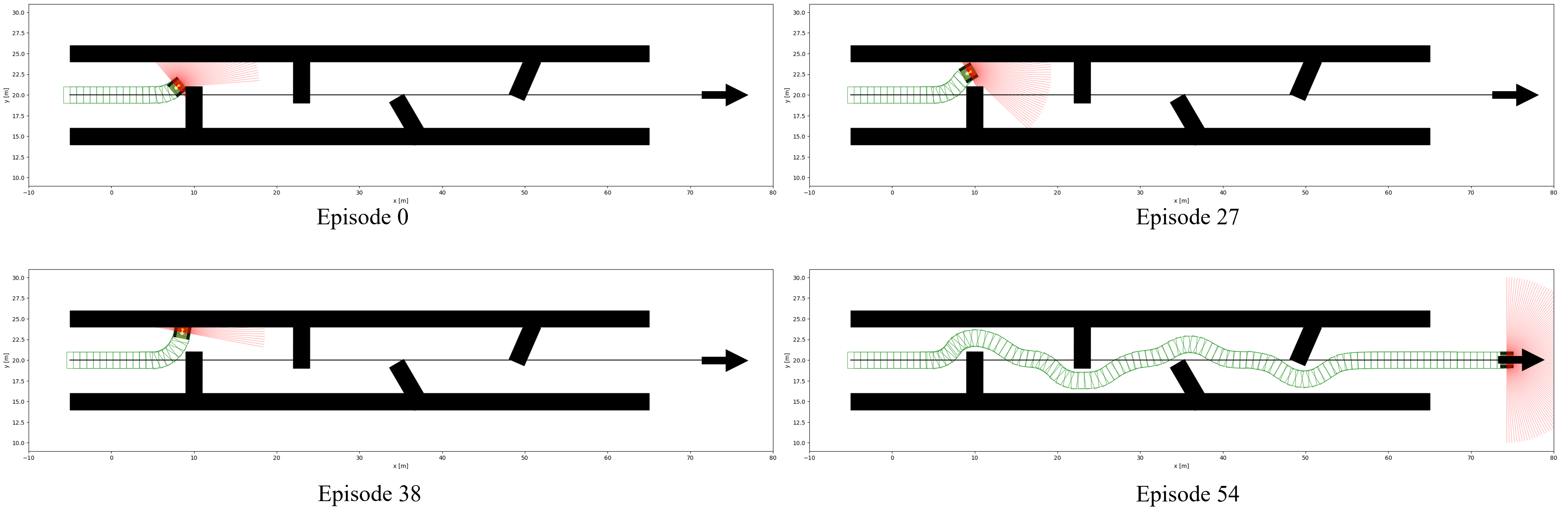}
      \caption{ Robot trajectories with a noise-free equipped lidar. }
      \label{lon2_no_noise}
  \end{subfigure}
  \hfill
  \begin{subfigure}[t]{0.48\textwidth}
      \centering
      \includegraphics[width=0.99\textwidth]{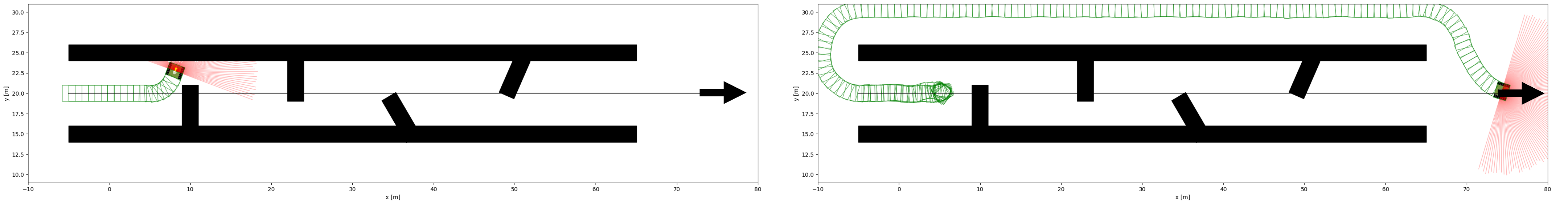}
      \caption{ Robot trajectories with equipped lidar range subjected to Gaussian noise $(0, 0.2)$. }
      \label{lon2_noise_0.2}
  \end{subfigure}
  \begin{subfigure}[t]{0.48\textwidth}
      \centering
      \includegraphics[width=0.99\textwidth]{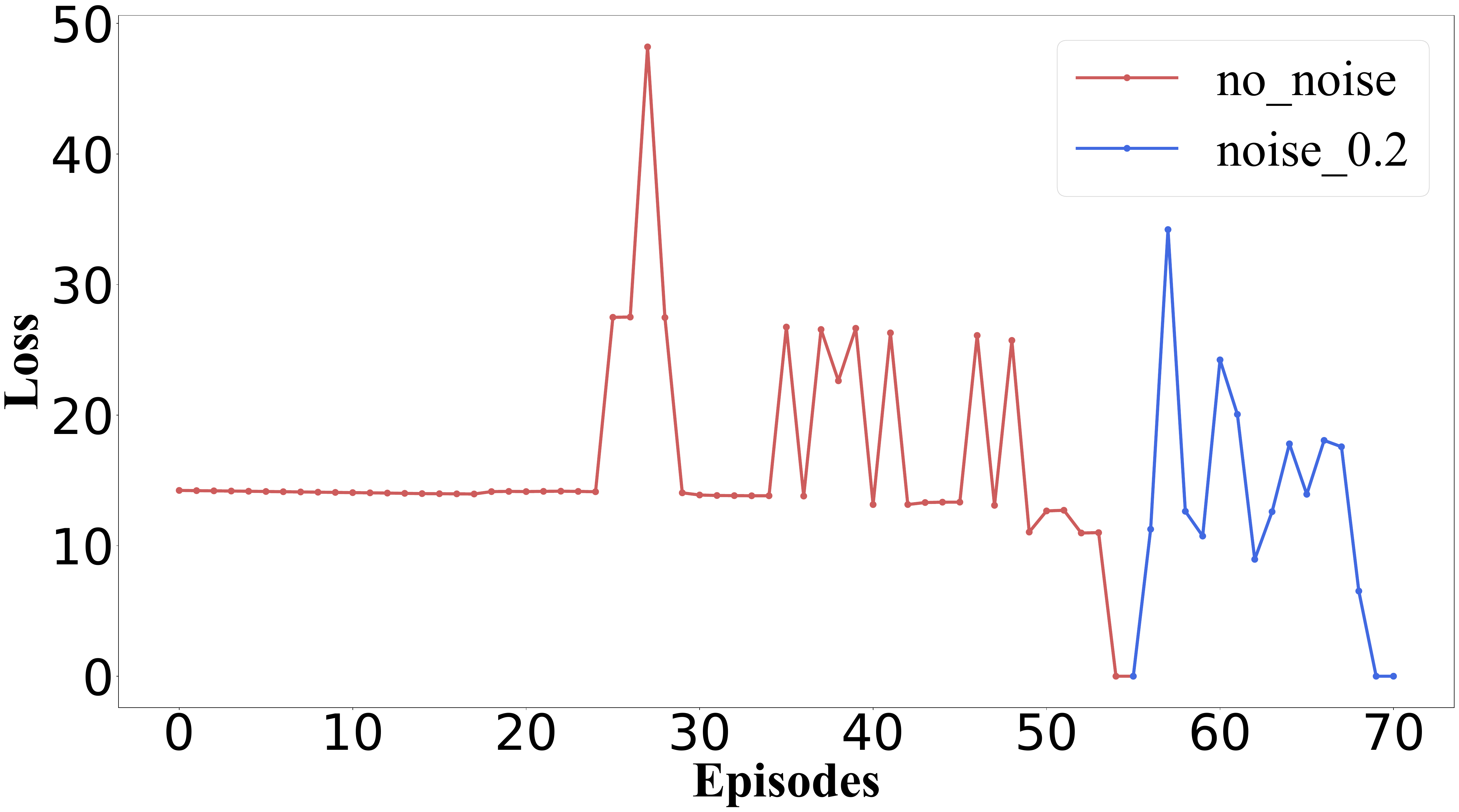}
      \caption{ Change in loss over episodes. }
      \label{error_change_noises}
  \end{subfigure}
  \caption{Robot trajectories and training loss change over episodes.}
  \label{fig_loss_noise}
\end{figure}

The proposed NeuPAN is able to learn from rewards provided by the environments, thereby handling domain variations (e.g., sensor noises) through auto-tuning. 
To see this, we consider a corridor scenario with $\mathsf{DoN}$=0.67. Given start and goal positions $[-5, 20]$ and $[75, 20]$, respectively, the associated result is shown in Fig.~\ref{fig_loss_noise}. 
Initially, we configure NeuPAN with improper parameters and the robot 
experiences a collision at the episode $0$ (i.e., the first subfigure of Fig.~\ref{lon2_no_noise}). 
By back-propagation of failure using the loss function $L_3(\mathcal{P'})$, 
the robot is able to make progress after $27$ and $38$ training episodes (i.e., the second and third subfigures of Fig.~\ref{lon2_no_noise}). 
After $54$ episodes of training, the robot successfully navigates through the scenario with proper parameters (the last subfigure of Fig.~\ref{lon2_no_noise}).

Subsequently, we test the well-trained NeuPAN in the same corridor scenario, but the robot lidar now involves sensor noises (i.e., Gaussian noise with standard deviation of $0.2$ added to the ranging measurements). While the well-trained NeuPAN has the ability to tackle the noise-free case and pass the corridor, it struggles under the sensor noises and collides with obstacles, as shown in the left hand side of Fig.~\ref{lon2_noise_0.2}. 
However, if we continue training NeuPAN for another $12$ episodes, the robot again navigates through the corridor. 
Even if the sensor noise is too large, causing the inner corridor becomes impassable, NeuPAN can still find an alternative outer path (right hand side of Fig.~\ref{lon2_noise_0.2}). This demonstrates the adaptability of NeuPAN in handling domain variations.
    
The average training loss over the number of episodes is shown in Fig.~\ref{error_change_noises}.
It can be seen that the loss decreases rapidly and converges to the minimal value under both perfect and noisy cases, which corroborates Fig.~\ref{lon2_no_noise} and Fig.~\ref{lon2_noise_0.2}, and shows that the training speed is fast. This is due to the model-based learning nature of NeuPAN, which tunes only a small number of learnable parameters through back propagation as mentioned in Section~\ref{section nrmp}.

\begin{figure*}[t]
  \centering
  \begin{subfigure}[t]{0.24\textwidth}
    \centering
    \includegraphics[width=0.99\textwidth]{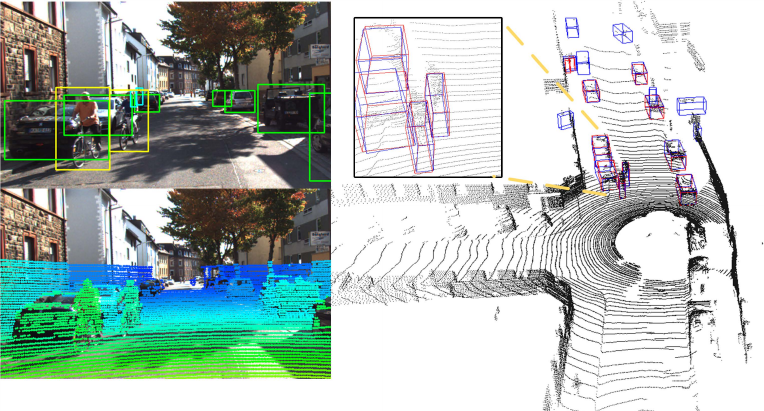}
    \caption{KITTI dataset }
    \label{kitti:vis}
  \end{subfigure}
  \begin{subfigure}[t]{0.24\textwidth}
    \centering
    \includegraphics[width=0.99\textwidth]{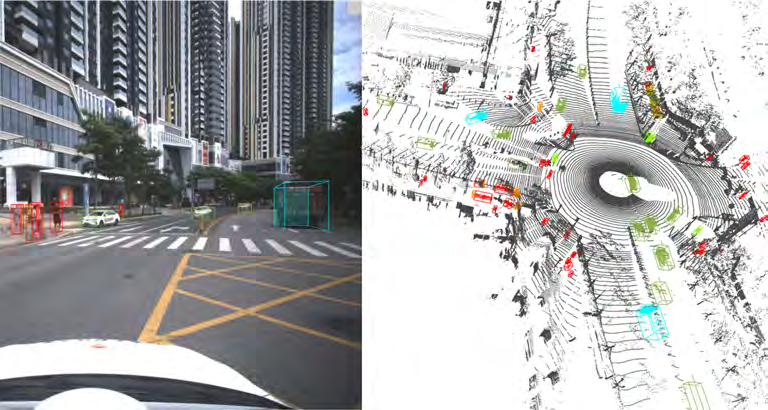}
    \caption{SUSCAPE dataset }
    \label{byd:vis}
  \end{subfigure}
  \begin{subfigure}[t]{0.24\textwidth}
    \centering
    \includegraphics[width=0.99\textwidth]{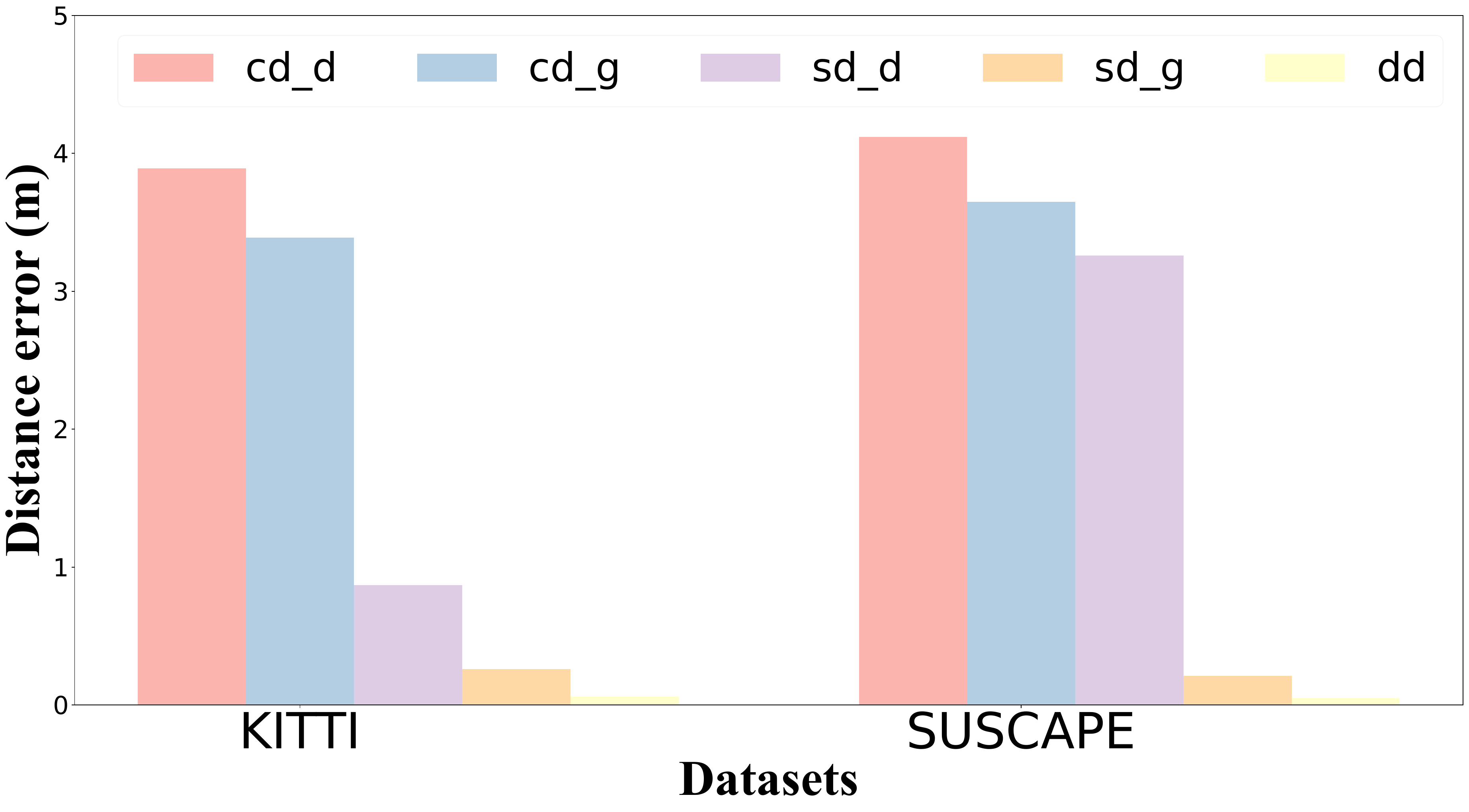}
    \caption{Distance errors comparison}
    \label{byd:error}
  \end{subfigure}
  \begin{subfigure}[t]{0.24\textwidth}
    \centering
    \includegraphics[width=0.99\textwidth]{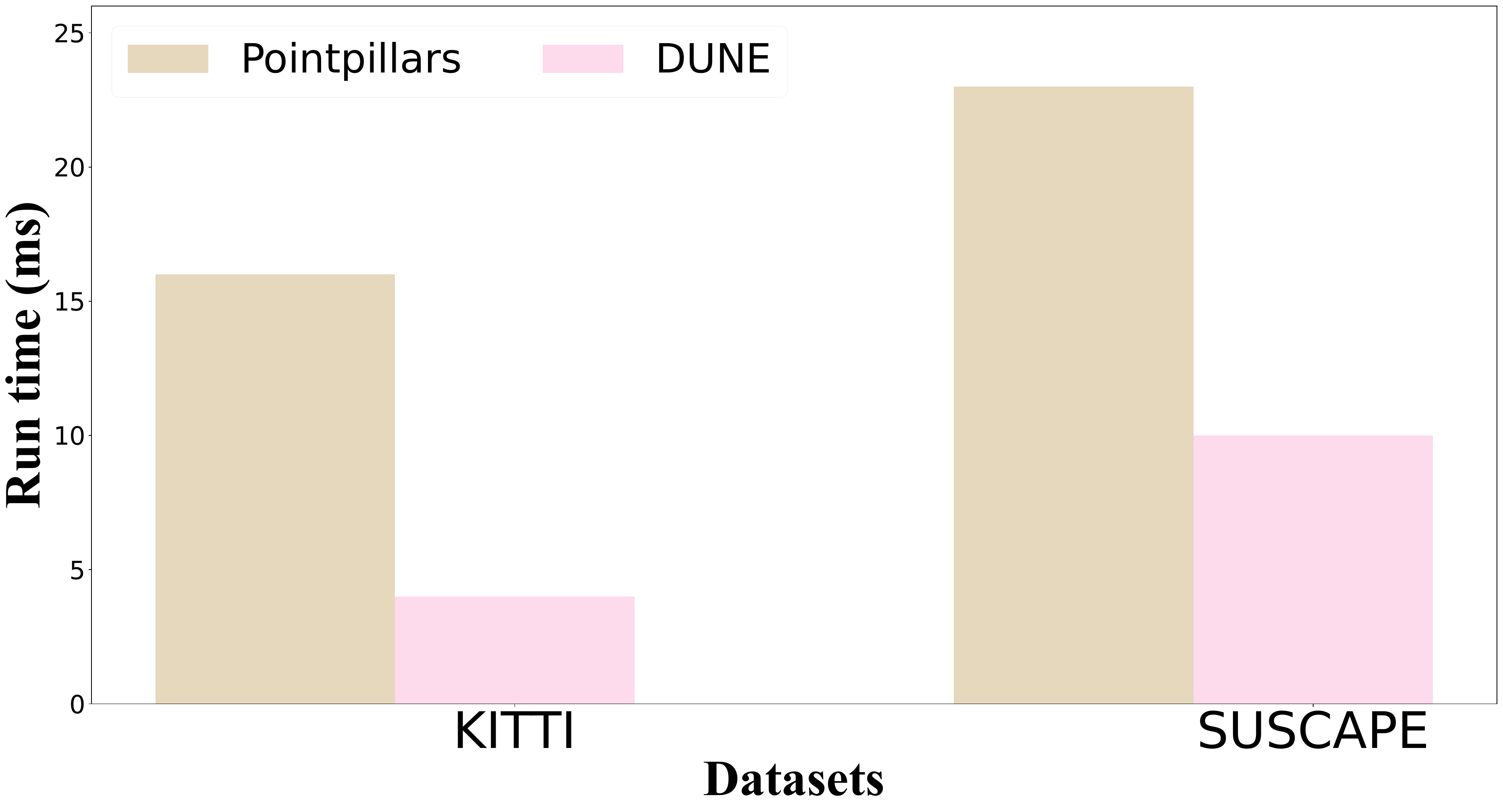}
    \caption{Inference time comparison}
    \label{byd:infer}
  \end{subfigure}
  \caption{Comparison of distance errors on KITTI and SUSCAPE datasets. (a) Object detection results by Pointpillars on KITTI. Blue and red boxes denote the detected result and the ground truth, respectively. (b) Perfect object detection results on SUSCAPE, where all boxes are labeled by human experts. (c) Average distance errors of cd, sd, and dd on KITTI and SUSCAPE. (d) Inference time comparison between DUNE and Pointpillars.}
  \label{distance_error}
\end{figure*}

\subsection{Experiment 2: Validation of DUNE on Open Datasets}

In this subsection, real-world experiments are presented to verify the efficacy of our proposed DUNE block of NeuPAN as well as its advantage over existing methods adopting inexact distances ( Supplementary Material Appendix E). We consider two open-source datasets: 1) KITTI~\cite{Geiger2012CVPR}, which is a popular real-world urban driving dataset (see Fig.~\ref{kitti:vis}); 2) SUSCAPE~\footnote{Online. Available: \url{https://suscape.net/home.}}, which is a large-scale multi-modal dataset with more than $1000$ scenarios.
This SUSCAPE dataset is collected and processed based on the autonomous vehicle platform in Fig.~\ref{platform}(c).
We calculate the minimum distance between the ego-vehicle and all surrounding objects based on three different rules: center point distance (cd), full shape set distance (sd), and our DUNE distance (dd). The computations of cd and sd based on the boxes of the object detectors are denoted by cd\_d and sd\_d.
Here we choose a celebrated point based detector Pointpillars~\cite{lang2019pointpillars}, whose accuracy achieves $74.31$\% (moderate) for the car detection task. The distance error is defined by the difference between the calculated distance and the ground truth distance (computed based on the raw point cloud provided by the dataset).
It can be seen from Fig.~\ref{kitti:vis} and~\ref{byd:error} that the distance error of DUNE block of NeuPAN is significantly smaller than the errors of the other approaches, which concisely quantify the benefit brought by direct point robot navigation. As such, we can effectively overcome the detection inaccuracies inherent in conventional object detection algorithms.

Qualitative results in the SUSCAPE dataset are shown in Fig.~\ref{byd:vis} and the associated quantitative results are shown in Fig.~\ref{byd:error}. The results show that our DUNE distance still has the smallest error. In addition, the cd\_d and sd\_d errors increase compared with the result in KITTI. This is due to the lack of the generalizability of the learning based object detector. In contrast, our DUNE block of NeuPAN can be directly applied across different datasets.
Finally, one may wonder whether we can reduce this distance error by replacing Pointpillars with other highly accurate object detectors.
To this end, we conduct another experiment of considering a perfect object detector with $100\,\%$ accuracy, i.e., we directly use the ground truth bounding boxes labeled by human experts as its outputs. The computations based on this detector can be denoted by cd\_g and sd\_g.
It can be seen from Fig.~\ref{byd:error} that the errors of cd\_g, sd\_g are smaller than that of cd\_d and sd\_d due to the improvement in object detection. However, these errors are still much larger than the dd error. This is because there exist a gap between the shapes of a box and a nonconvex object.

The inference latencies of DUNE block of NeuPAN and Pointpillars on two datasets (tested using the same computer platform) are shown in Fig.~\ref{byd:infer}. It can be seen that the inference time of DUNE achieves $4\,$ms in KITTI dataset and $10\,$ms in SUSCAPE dataset, respectively, which is much faster than the $16\,$ms and $23\,$ms inference time of Pointpillars. This is because DUNE is a fully explainable lightweight neural network with only six fully connected layers, which is much more efficient than Pointpillars network with deep convolution layers.

\begin{figure}[t]
  \centering
  \includegraphics[width=0.45\textwidth]{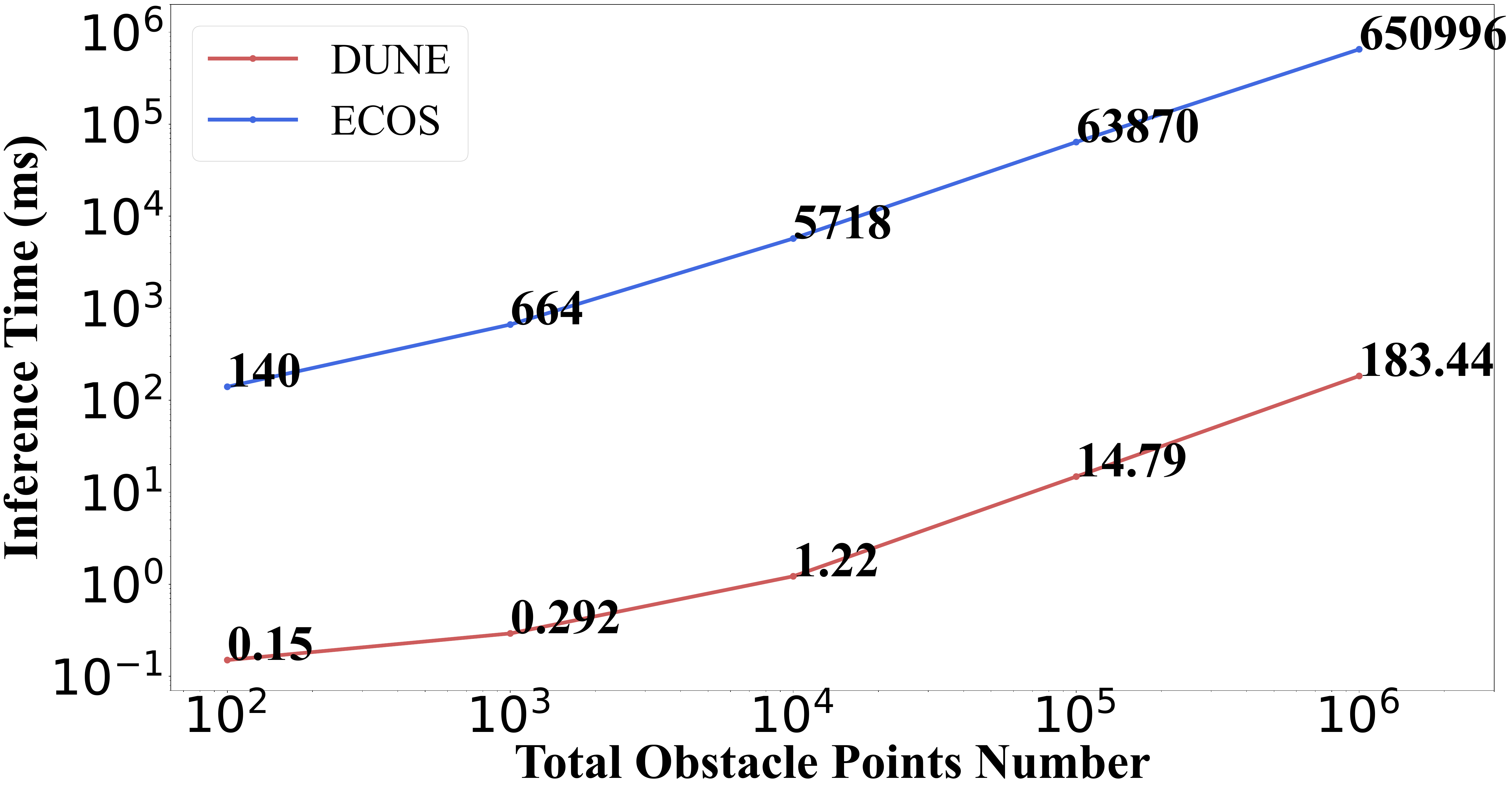}
  \caption{Inference time comparison of DUNE and the optimization solver ECOS with varying number of input points.}
  \label{fig:times}
\end{figure}
To analyze the scalability of DUNE block of NeuPAN, its computation time over the number of points is shown in Fig.~\ref{fig:times}. For comparison, we also implement a benchmark scheme ECOS~\cite{bib:Domahidi2013ecos}, which is an off-the-shelf optimization solver for solving problem Eq.~\eqref{distance_dual}.
All the tests are performed on a computer equipped with an AMD Ryzen 9 CPU. It can be seen that the proposed DUNE is able to handle $1$ million points within $0.2$\,second. In contrast, the baseline approach ECOS requires more than $650$\,seconds for handling $1$ million points, which is significantly higher than that of DUNE. This demonstrates that DUNE reduces the computation time by over $1000$x when the number of input points is in the range of thousands or more. Such significant acceleration is due to the model-based learning nature of DUNE, which corroborates the theory and methodology in Section~\ref{section dune}.

\subsection{Experiment 3: Ground Mobile Robot Navigation}

\begin{table*}[t]
  \centering
  \caption{Performance comparison of NeuPAN, RDA, and AEMCARL in Gazebo}
  \label{comparison_gazebo}
  \resizebox{\textwidth}{!}{%
    \begin{tabular}{cc|ccclccclccc}
      \hline
      \multicolumn{2}{c|}{\multirow{2}{*}{Scenarios}}           & \multicolumn{3}{c}{Success Rate} &                         & \multicolumn{3}{c}{Navigation Time} &      & \multicolumn{3}{c}{Average Speed}                                                                                                                                                              \\ \cline{3-5} \cline{7-9} \cline{11-13}
      \multicolumn{2}{c|}{}                                     & \multicolumn{1}{l}{NeuPAN}       & \multicolumn{1}{l}{RDA} & \multicolumn{1}{l}{AEMCARL}         &      & \multicolumn{1}{l}{NeuPAN}        & \multicolumn{1}{l}{RDA} & \multicolumn{1}{l}{AEMCARL} &       & \multicolumn{1}{l}{NeuPAN} & \multicolumn{1}{l}{RDA} & \multicolumn{1}{l}{AEMCARL}         \\ \hline
      \multicolumn{1}{c|}{\multirow{3}{*}{Cylinder Obstacles}}  & 10                               & \textbf{0.93}           & 0.89                                & 0.89 &                                   & \textbf{35.61}          & 38.51                       & 53.45 &                            & \textbf{0.285}          & 0.280                       & 0.259 \\
      \multicolumn{1}{c|}{}                                     & 15                               & \textbf{0.90}           & 0.88                                & 0.82 &                                   & \textbf{36.71}          & 42.19                       & 57.71 &                            & \textbf{0.279}          & 0.263                       & 0.257 \\
      \multicolumn{1}{c|}{}                                     & 20                               & \textbf{0.88}           & 0.82                                & 0.73 &                                   & \textbf{38.17}          & 46.03                       & 61.21 &                            & \textbf{0.264}          & 0.246                       & 0.221 \\ \hline
      \multicolumn{1}{c|}{\multirow{3}{*}{Nonconvex Obstacles}} & 10                               & \textbf{0.90}           & 0.70                                & -    &                                   & \textbf{37.91}          & 45.25                       & -     &                            & \textbf{0.273}          & 0.266                       & -     \\
      \multicolumn{1}{c|}{}                                     & 15                               & \textbf{0.88}           & 0.62                                & -    &                                   & \textbf{39.87}          & 48.36                       & -     &                            & \textbf{0.261}          & 0.259                       & -     \\
      \multicolumn{1}{c|}{}                                     & 20                               & \textbf{0.85}           & 0.59                                & -    &                                   & \textbf{42.47}          & 50.09                       & -     &                            & \textbf{0.255}          & 0.228                       & -     \\ \hline
    \end{tabular}%
  }
\end{table*}

\begin{figure*}[tb]
  \centering
  \begin{subfigure}[t]{0.24\textwidth}
    \centering
    \includegraphics[width=0.98\textwidth]{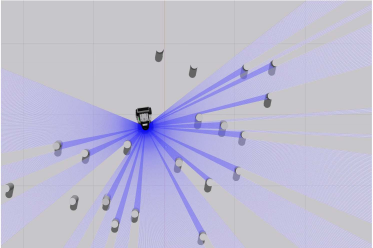}
    \caption{ Cylinder moving obstacles }
    \label{gazebo_s1}
  \end{subfigure}
  \begin{subfigure}[t]{0.24\textwidth}
    \centering
    \includegraphics[width=0.98\textwidth]{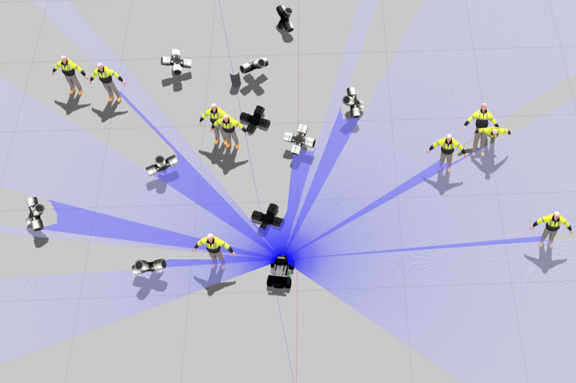}
    \caption{ Nonconvex moving obstacles }
    \label{gazebo_s2}
  \end{subfigure}
  \begin{subfigure}[t]{0.40\textwidth}
    \centering
    \includegraphics[width=0.95\textwidth]{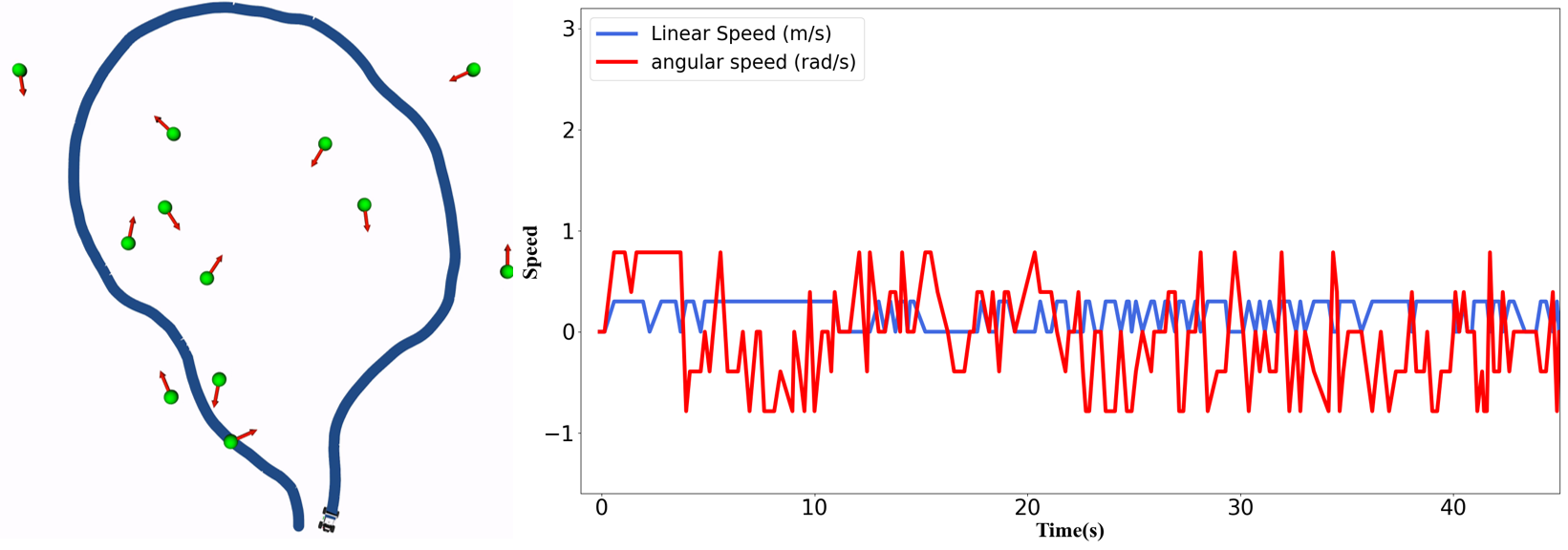}
    \caption{AEMCARL (cylinder) }
    \label{traj_aem1}
  \end{subfigure}
  \hfill
  \begin{subfigure}[t]{0.24\textwidth}
    \centering
    \includegraphics[width=0.95\textwidth]{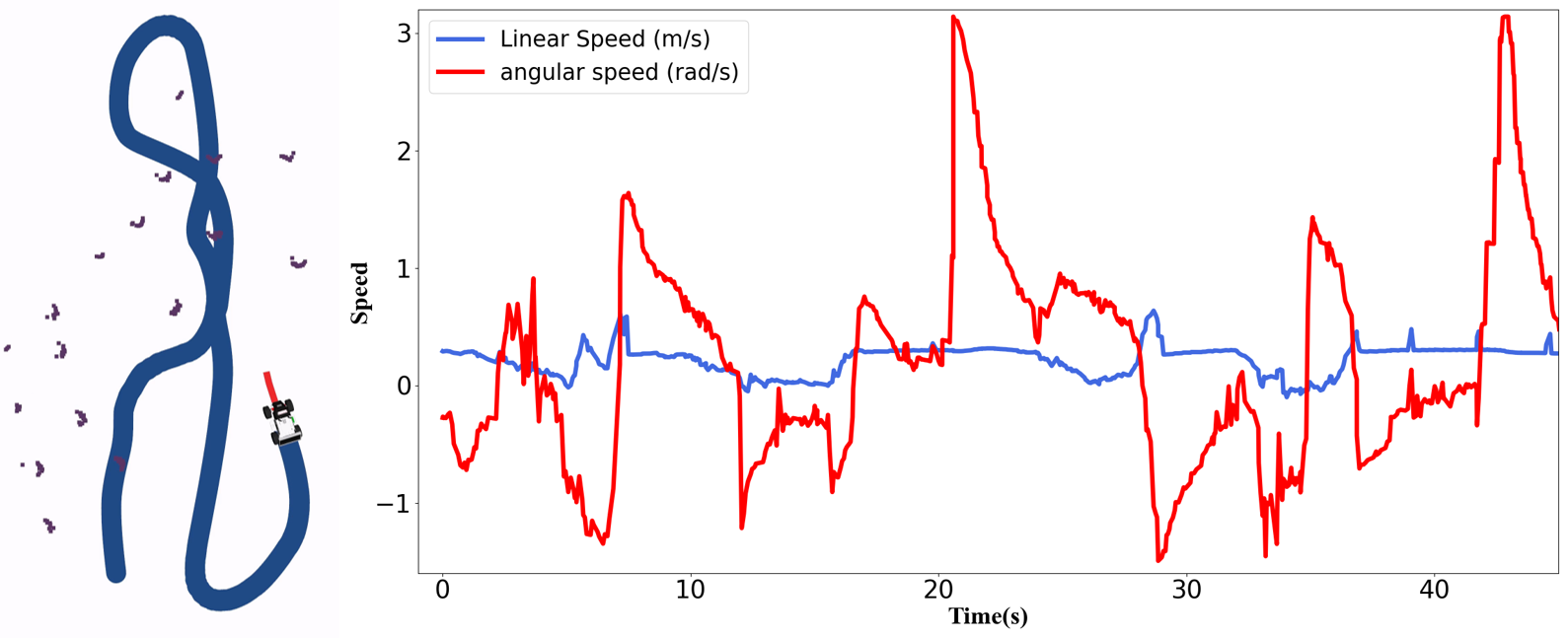}
    \caption{NeuPAN (cylinder)}
    \label{traj_neupan1}
  \end{subfigure}
  \begin{subfigure}[t]{0.24\textwidth}
    \centering
    \includegraphics[width=0.95\textwidth]{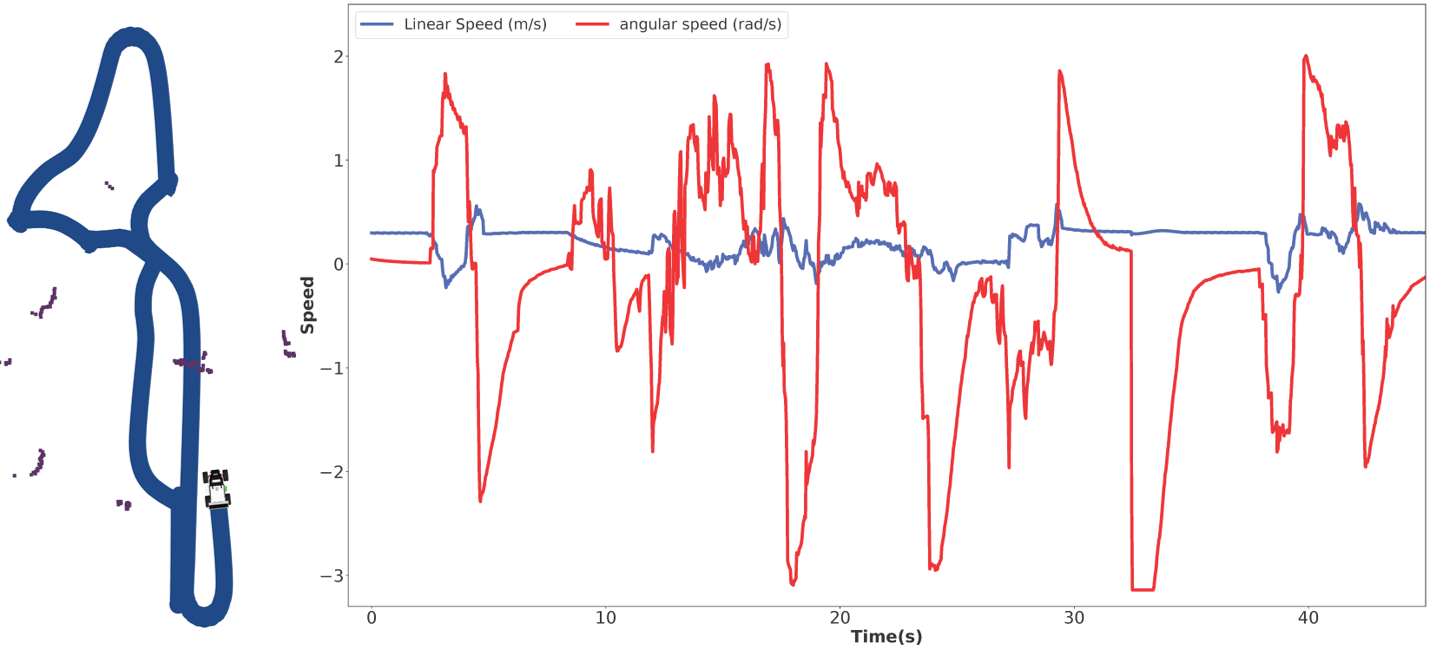}
    \caption{NeuPAN (nonconvex) }
    \label{traj_neupan2}
  \end{subfigure}
  \begin{subfigure}[t]{0.24\textwidth}
    \centering
    \includegraphics[width=0.95\textwidth]{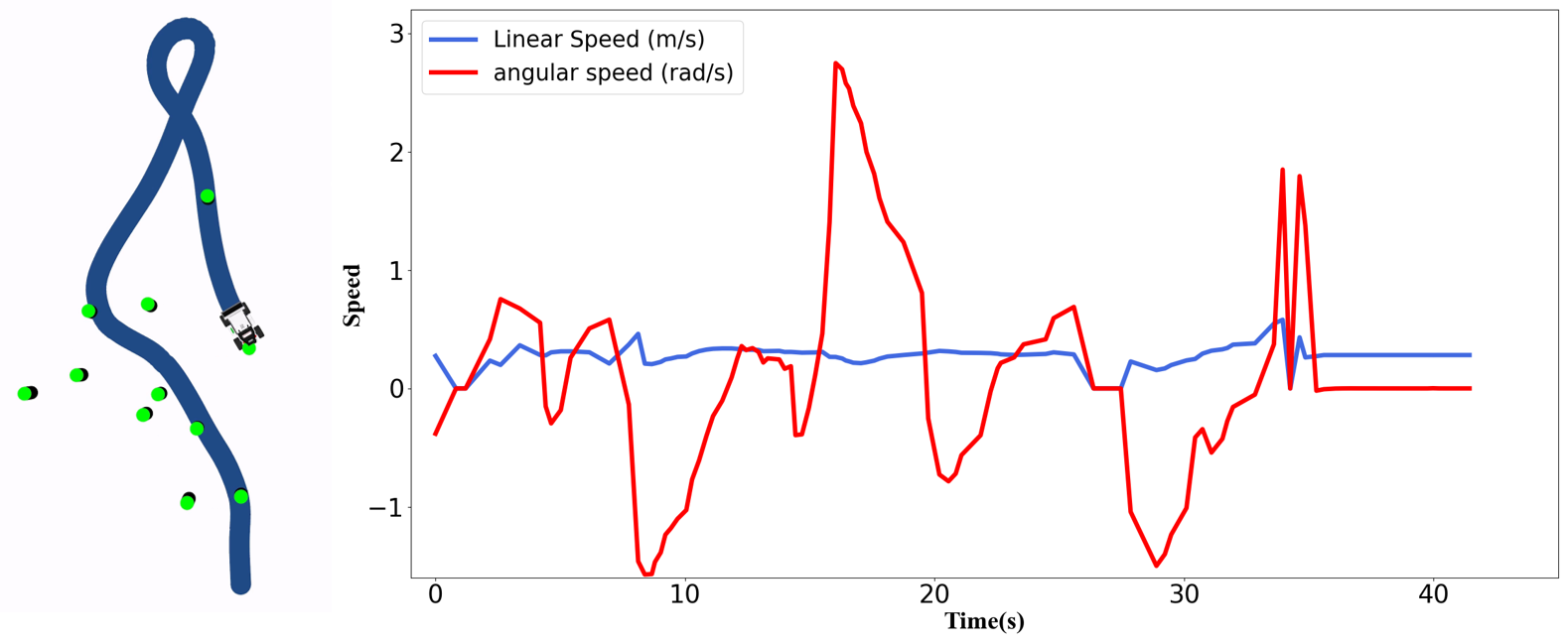}
    \caption{RDA (cylinder) }
    \label{traj_rda1}
  \end{subfigure}
  \begin{subfigure}[t]{0.24\textwidth}
    \centering
    \includegraphics[width=0.95\textwidth]{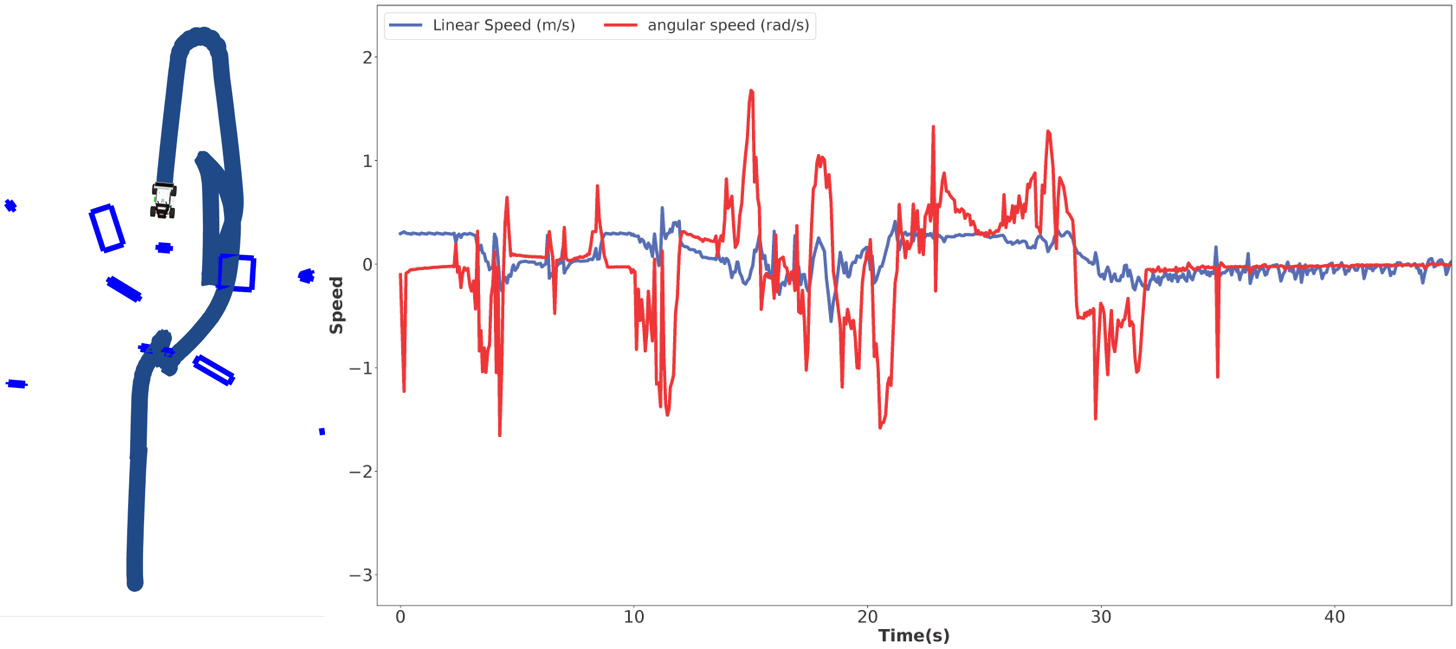}
    \caption{RDA (nonconvex) }
    \label{traj_rda2}
  \end{subfigure}
  \caption{ Trajectory and speed profile comparisons of NeuPAN, RDA, and AEMCARL in Gazebo dynamic environment with cylinder and nonconvex moving obstacles.}
  \label{limo_gazebo}
\end{figure*}

\subsubsection{Dynamic Environment}
We adopt Gazebo~\cite{koenig2004design}, a widely recognized open-source 3D robotics simulator, to demonstrate the dynamic collision avoidance capability of our approach by using onboard 2D lidar.
The experimental setup is illustrated in Fig.~\ref{gazebo_s1}-~\ref{gazebo_s2}, where the robot, operating in differential steering mode, must patrol between two given checkpoints as fast as possible while preventing itself from being surrounded and collided by massive adversarial moving obstacles.
The obstacles with cylinder and nonconvex shapes are randomly placed within a region of interest and exhibit reciprocal collision avoidance behavior with speed adjustments (max speed: 0.1\,m/s). 
We compare our method with the reinforcement learning based method AEMCARL~\cite{wang2022adaptive} and optimization based method RDA~\cite{han2023rda}.
All these methods adopt a desired speed of $0.3\,$m/s.

Experimental results based on $50$ trials with the number of cylinder and nonconvex obstacles ranging from 15 to 20 are presented in Table~\ref{comparison_gazebo}. It can be seen that for scenarios with cylinder obstacles, NeuPAN improves the success rates by $4.69\%$ and $11.60\%$ compared with RDA and AEMCARL.
Moreover, NeuPAN reduces the navigation time by $12.53\%$ and $35.80\%$, and improves the mobility speed by $5.07\%$ and $12.36\%$, relative to RDA and AEMCARL respectively. 
On the other hand, for scenarios with nonconvex obstacles, NeuPAN improves the success rate by $38.19\%$, reduces the navigation time by $16.33\%$, and improves the mobility speed by $5.08\%$, compared to RDA.
Note that AEMCARL fails in this situation, since AEMCARL is developed based on the point-mass model (point with inflated radius), which is not applicable to nonconvex obstacles.

It can be seen that none of the simulated schemes achieves $100$\,\% success rate.
These failures happen due to two reasons. First, the robot's laser scanner has a limited field of view (FoV), which may fail to detect moving obstacles outside its coverage area. Consequently, the robot might collide with these undetected obstacles before it can react. Second, obstacles are densely packed and moving within a confined area. Their behaviors are randomized and they do not actively avoid the ego robot, as demonstrated in~\cite{chen2019crowd}. As a result, there are scenarios where the robot has no viable escape path. For example, if the robot is completely surrounded by obstacles, a collision is inevitable even if the robot remains stationary, as the obstacles would converge toward it.

The robot trajectories and speed profiles of the three schemes are shown in Fig.~\ref{limo_gazebo}. It can be seen that AEMCARL tends to take detours in front of crowded obstacles due to its conservative neural policy and its over-simplified model treating obstacles as balls, resulting in longer navigation times (Fig.~\ref{traj_aem1}). On the other hand, RDA tends to collide with obstacles since it computes inexact robot-obstacle distances based on the polygons converted from laser scans, which introduces inevitable errors in representing nonconvex obstacles, leading to higher failure rates (Fig.~\ref{traj_rda1}-~\ref{traj_rda2}). Lastly, NeuPAN utilizes the raw points to represent obstacles, which gets rid of the distance errors involved in AEMCARL and RDA. 
By directly mapping the raw points to actions, NeuPAN achieves better navigation performance in complex environments with arbitrarily-shaped obstacles (Fig.~\ref{traj_neupan1}-~\ref{traj_neupan2}).

\subsubsection{Structured Real-World Testbed}

\begin{figure}[t]
  \centering
  \begin{subfigure}[t]{0.48\textwidth}
    \centering
    \includegraphics[width=0.90\textwidth]{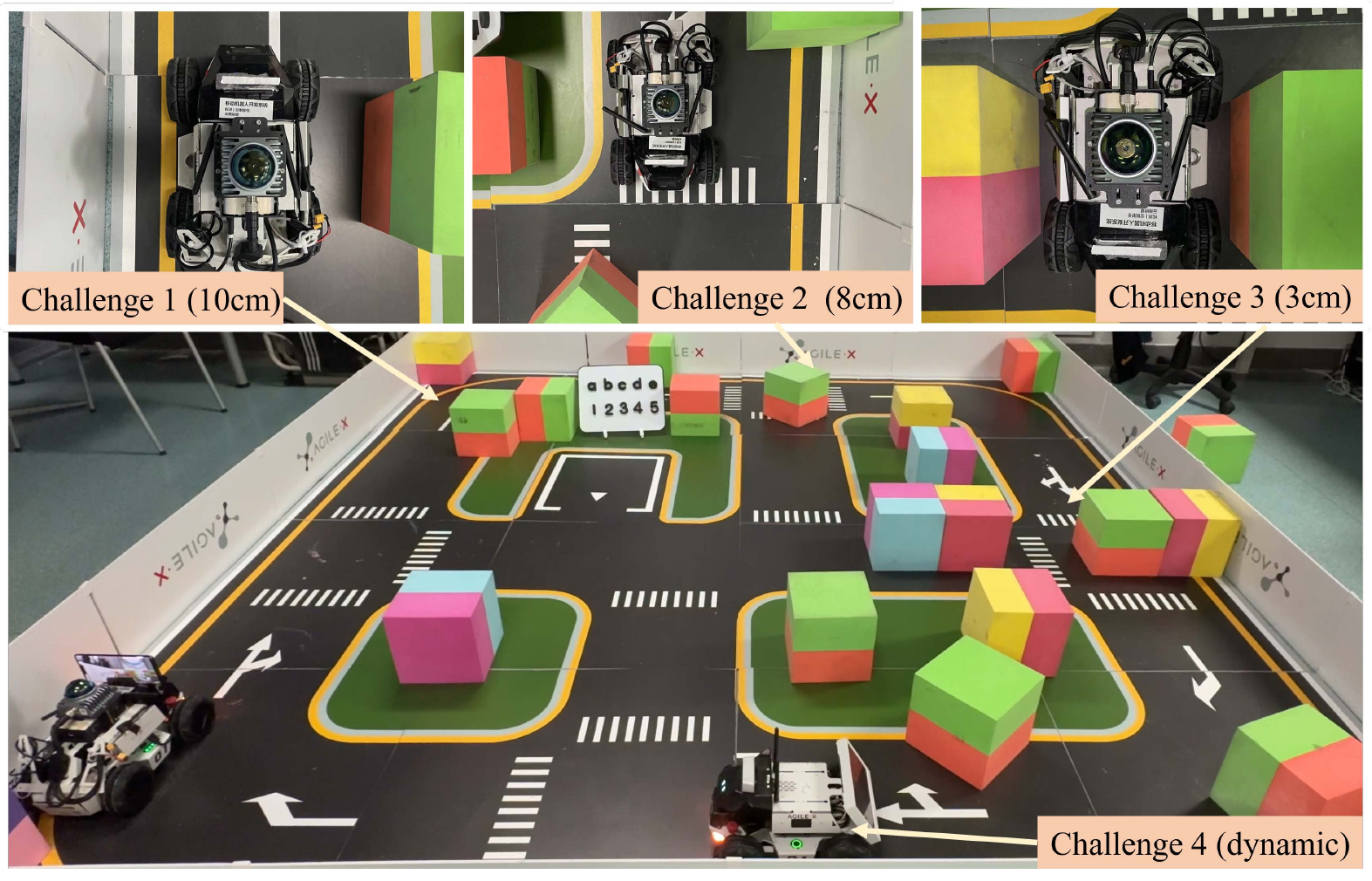}
    \caption{Configurations in the structured real-world testbed.}
    \label{sandbox_challenges}
  \end{subfigure}
  \hfill
  \begin{subfigure}[t]{0.15\textwidth}
    \centering
    \includegraphics[width=0.90\textwidth]{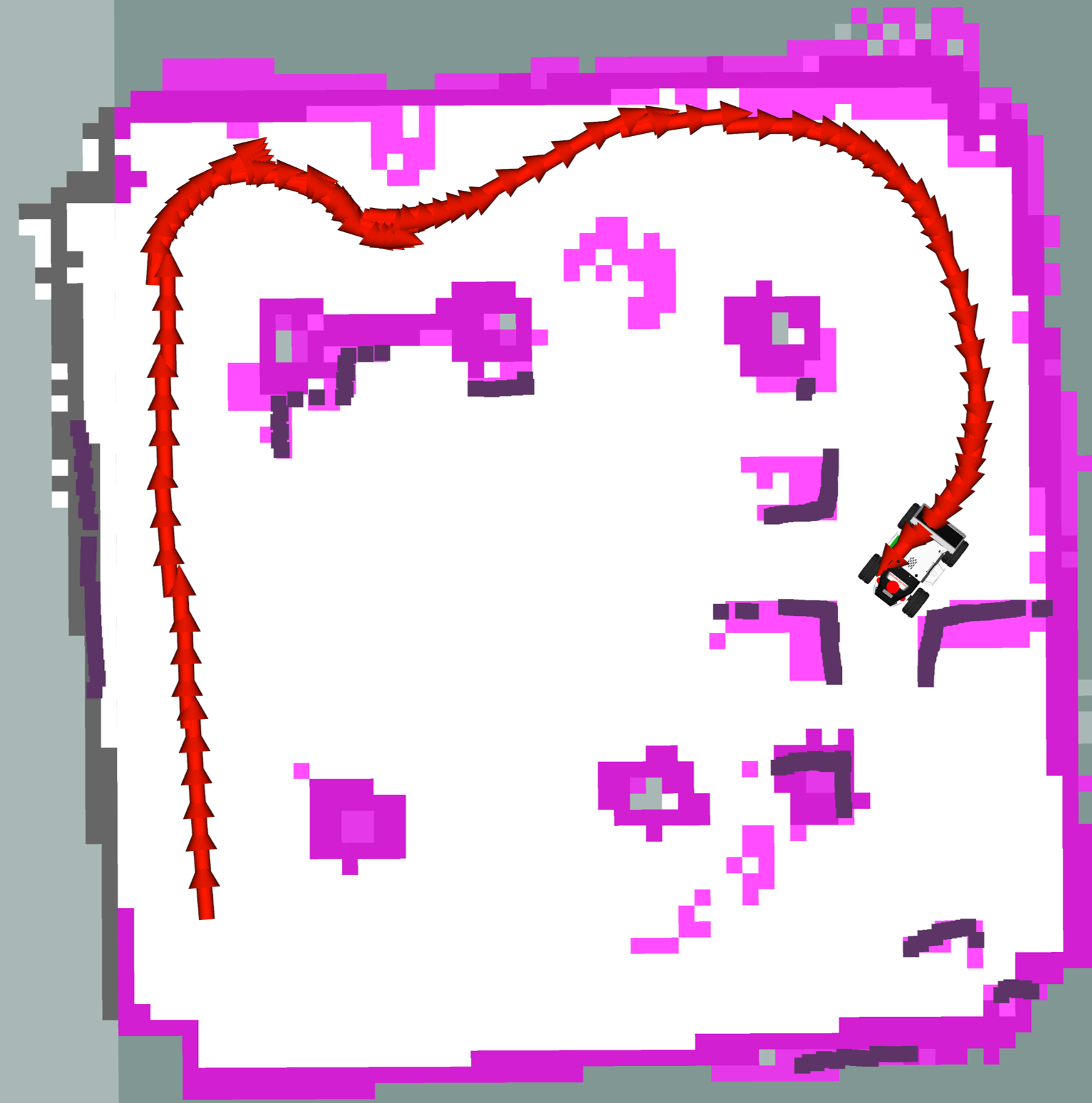}
    \caption{ TEB control }
    \label{teb_limo_sandbox}
  \end{subfigure}
  \begin{subfigure}[t]{0.15\textwidth}
    \centering
    \includegraphics[width=0.90\textwidth]{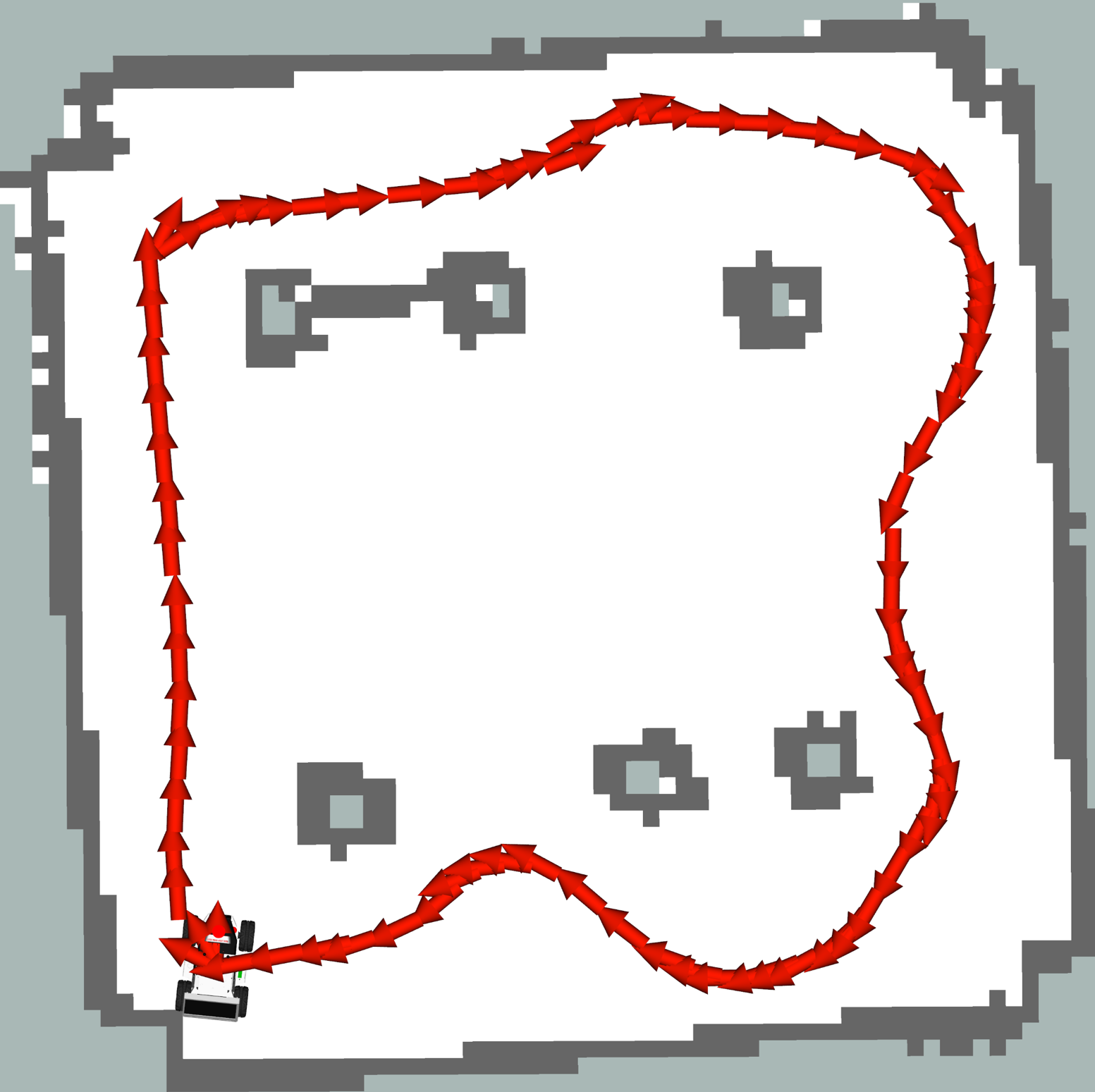}
    \caption{ Manual control }
    \label{human_limo_sandbox}
  \end{subfigure}
  \begin{subfigure}[t]{0.15\textwidth}
    \centering
    \includegraphics[width=0.90\textwidth]{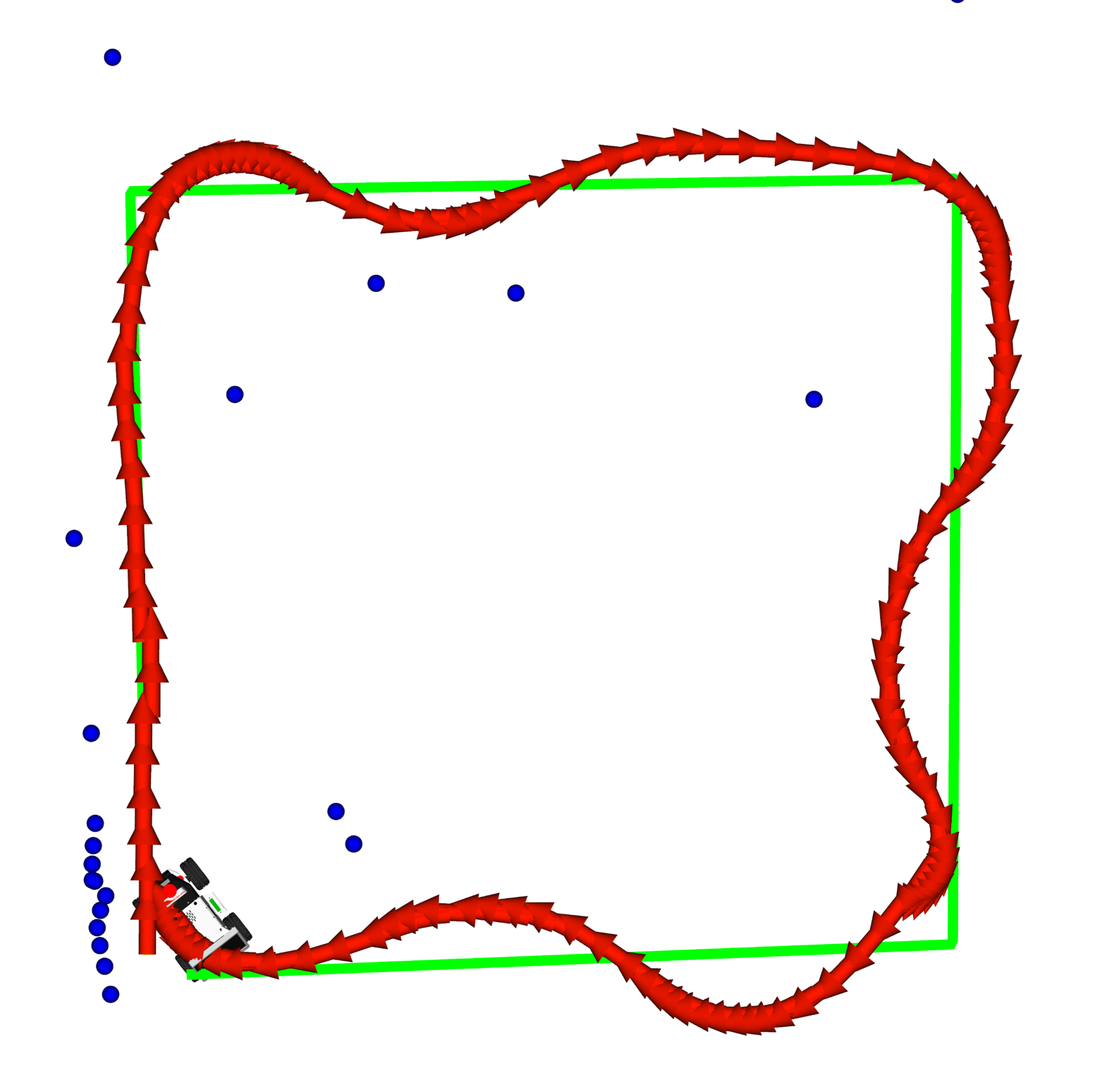}
    \caption{ NeuPAN control }
    \label{neupan_limo_sandbox}
  \end{subfigure}
  \caption{Trajectory comparison of TEB, manual, and NeuPAN control in a structured real-world testbed.}
  \label{limo_sandbox}
\end{figure}

\begin{table*}[t]
  \centering
  \caption{Performance comparison of NeuPAN, TEB, and manual control in the structured real-world testbed}
  \label{racing}
  \resizebox{\textwidth}{!}{
    \begin{tabular}{cccclcclcccl}
      \cline{1-5} \cline{6-12}
      \multirow{2}{*}{Scenario} & \multicolumn{3}{c}{Success} &                & \multicolumn{3}{c}{Navigation Time (s)} &  & \multicolumn{3}{c}{Moving Speed Mean / std (m/s)}                                                                                                                                                                                                                                                                                                             \\ \cline{2-4} \cline{6-8} \cline{10-12}
                                & NeuPAN                      & TEB            & Manual                                  &  & Neupan                                            & TEB   & Manual &  & Neupan                                                                                        & TEB                                                                                   & Manual                                                                                        \\ \cline{1-5} \cline{6-12}
      Challenge 1 (10cm)        & \CheckmarkBold              & \CheckmarkBold & \CheckmarkBold                          &  & \textbf{2.87}                                     & 4.11  & 4.90   &  & \multirow{4}{*}{\begin{tabular}[c]{@{}c@{}} \textbf{0.788} \\ / 0.23 \end{tabular}          } & \multirow{4}{*}{ \begin{tabular}[c]{@{}c@{}}  0.265 \\ / 0.31 \end{tabular}         } & \multirow{4}{*}{\begin{tabular}[c]{@{}c@{}}  0.396 \\ / \textbf{0.165} \end{tabular}        } \\
      Challenge 2 (8cm)         & \CheckmarkBold              & \CheckmarkBold & \CheckmarkBold                          &  & \textbf{6.07}                                     & 14.99 & 10.31  &  &                                                                                               &                                                                                       &                                                                                               \\
      Challenge 3 (3cm)         & \CheckmarkBold              & \XSolidBrush   & \CheckmarkBold                          &  & \textbf{9.25}                                     & -     & 16.98  &  &                                                                                               &                                                                                       &                                                                                               \\
      Challenge 4 (dynamic)     & \CheckmarkBold              & \XSolidBrush   & \CheckmarkBold                          &  & \textbf{12.49}                                    & -     & 23.95  &  &                                                                                               &                                                                                       &                                                                                               \\ \hline
    \end{tabular}}
\end{table*}

We conduct the experiments in the structured real-world testbed (i.e., a sandbox with size $4\,$m$\times4\,$m) to demonstrate the high accuracy of NeuPAN, as shown in Fig.~\ref{limo_sandbox}. The robot is tasked to race within the track as fast as possible without any collision. We set up challenges 1--3 with different $\mathsf{DoN}$s, by varying the positions of $20\,$cm$\times20\,$cm cubes. The narrowest spaces in challenges 1--3 are $32\,$cm, $30\,$cm, and $25\,$cm, respectively. 
Taking the robot width into account, the remaining margins are $10\,$cm, $8\,$cm, $3\,$cm, respectively, corresponding to the $\mathsf{DoN}$ of $0.52$, $0.58$, $0.79$. 
These challenges represent different passing difficulties and are used to evaluate the limit of navigation accuracy. We also set up challenge 4 with a dynamic obstacle, where we adopt another mobile robot to challenge the ego robot inside the track.

We compare the performance of our approach NeuPAN with the celebrated motion planner TEB~\cite{rosmann2017kinodynamic} and manual control in terms of route completion, navigation time, and average speed.
From Fig.~\ref{limo_sandbox}, it is also observed that the TEB method requires a pre-built occupancy grid map. In contrast, our approach only requires the goal state information, i.e., four checkpoints at the corners. The desired speed is set to be $1\,$m/s. Quantitative results are listed in Table.~\ref{racing}.

It can be seen that NeuPAN successfully passes all challenges and finishes the track with the shortest time of $12.49\,$s and highest speed of $0.788\,$m/s, approaching the maximum speed $1\,$m/s of the robot. This indicates that NeuPAN achieves a limit of $3\,$cm control accuracy in this task. In contrast, TEB gets stuck in front of challenge 3 ($3\,$cm-margin passage), failing in completing the trip. This is because TEB relies on occupancy grid maps and the grid representation of obstacles leads to degradation of navigation accuracy, making it inadequate for highly-accurate navigation. By incrementally varying the margin in challenge $3$, the TEB method passes challenge 3 at a margin of $7\,$cm, which represents its navigation accuracy limit. It can be seen that our method improves the accuracy by over $2$x compared to TEB.   
Lastly, by adopting manual control, it is also possible to control the robot through all these challenges. But due to the need to carefully avoid collision at narrow spaces, manual control leads to longer navigation time ($23.95\,$s) and lower moving speed ($0.396\,$m/s / 0.165), compared to our method.
The results corroborate to the effectiveness of our explainable end-to-end framework in the real-world experiment with ubiquitous uncertainties.

\subsubsection{Unstructured Real-World Environment}

\begin{figure*}[t]
  \centering
  \includegraphics[width=0.90\textwidth]{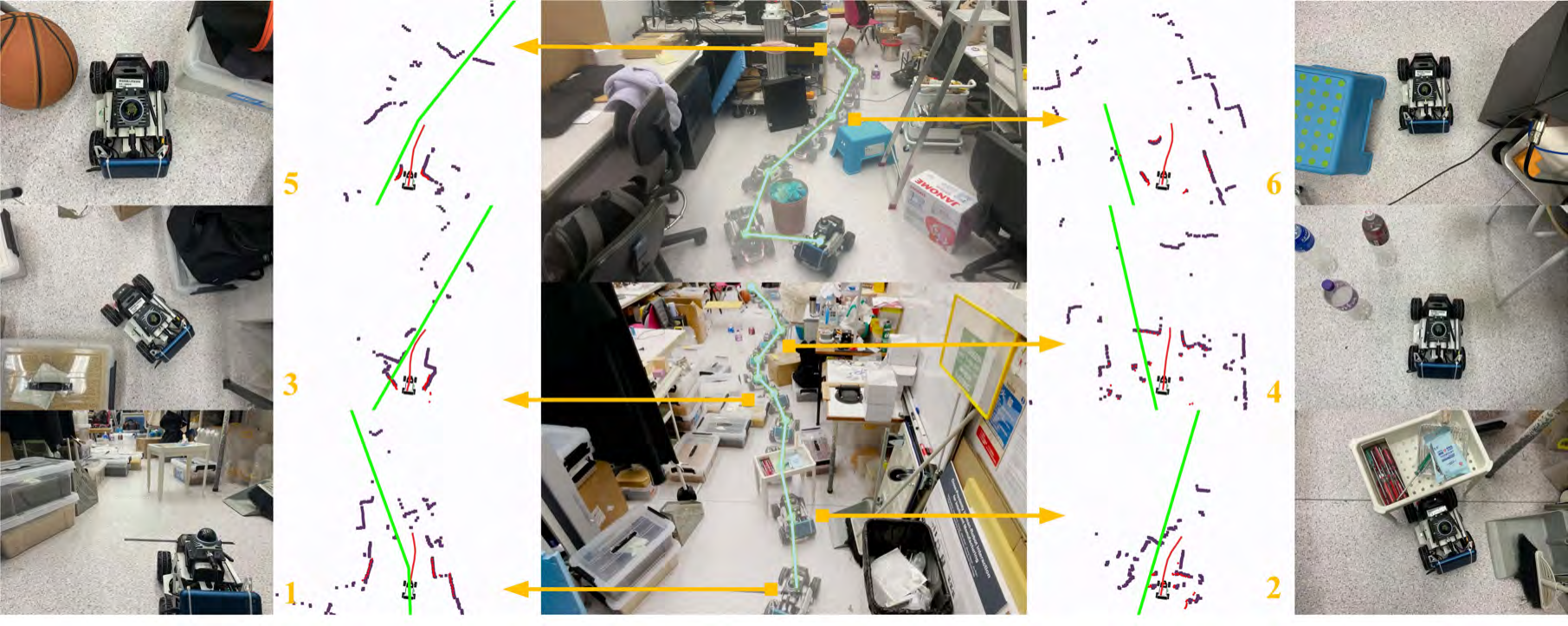}
  \caption{ Robot navigation in an unstructured cluttered environment. Green straight line is the naive path. Purple points are the raw 2D lidar data. Red line is the optimal receding path. Red points are the input obstacle points. 3D lidar is used for localization. The narrowest space (e.g. (5)) has only about 3-centimeters tolerance with $\mathsf{DoN}=0.88$.}
  \label{unstructured}
\end{figure*}

\begin{figure}[t]
  \centering
  \begin{subfigure}[t]{0.48\textwidth}
    \centering
    \includegraphics[width=0.95\textwidth]{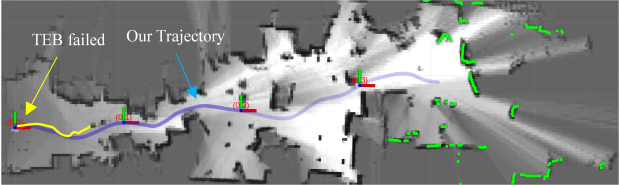}
    \caption{ Trajectories in the grip map. }
    \label{limo_carto}
  \end{subfigure}
  \begin{subfigure}[t]{0.48\textwidth}
    \centering
    \includegraphics[width=0.95\textwidth]{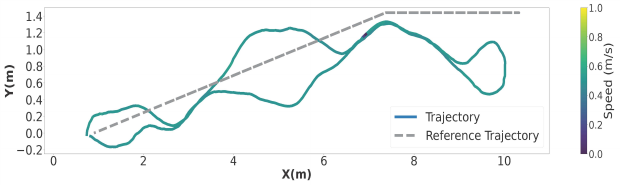}
    \caption{ Trajectory with speed. }
    \label{limo_traj}
  \end{subfigure}
  \begin{subfigure}[t]{0.48\textwidth}
    \centering
    \includegraphics[width=0.95\textwidth]{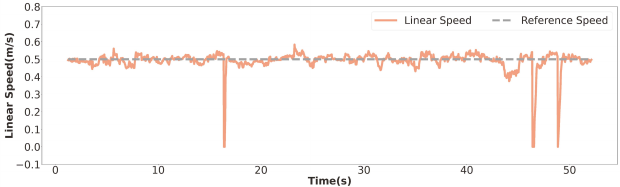}
    \caption{ Linear speed }
    \label{limo_linear}
  \end{subfigure}
  \begin{subfigure}[t]{0.48\textwidth}
    \centering
    \includegraphics[width=0.95\textwidth]{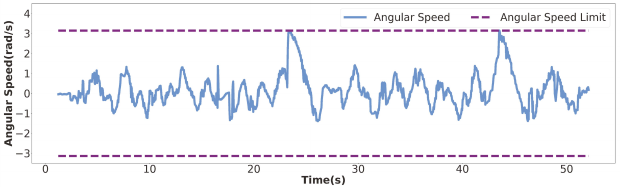}
    \caption{ Angular speed }
    \label{limo_angular}
  \end{subfigure}
  \caption{Trajectory, linear, and angular speed profiles of the agile mobile robot during the office navigation task. }
  \label{limo_traj_sped}
\end{figure}

The above experiments have considered structured environments, where the obstacles are easy to be detected (e.g., box, wall) and classified. However, for extensive real-life applications (e.g., housekeeper machine),
the target scenarios are highly unstructured and disordered.
To this end, we further test NeuPAN in a highly cluttered laboratory as shown in Fig.~\ref{unstructured}. The laboratory is filled with sundries, such as chairs, sandboxes, water bottles, and equipment, which are unrecognizable objects with arbitrary shapes.
The robot needs to patrol in this environment at a desired speed of $0.5\,$m/s without a pre-established map.
First, we invite $10$ human pilots to manually control the robot. But unfortunately, all of them fail, resulting in either robot collisions or time out due to insufficient speed.
Then, we adopt NeuPAN to handle this situation.
This time, the task is accomplished, with the associated trajectories (marked as blue lines) illustrated in Fig.~\ref{unstructured}.
The narrowest place has only about 3-centimeters tolerance with $\mathsf{DoN}=0.88$ (see Fig.~\ref{unstructured}(5)). But owing to the exact distance perception from DUNE and high accurate motion from NRMP, our NeuPAN method is able to navigate the robot through this challenging scenario with only onboard lidar in real time.

The trajectories and speed profiles of NeuPAN and TEB during this task are shown in Fig.~\ref{limo_traj_sped}. We build the occupancy grid map by Cartographer~\cite{hess2016real}, which is a celebrated SLAM algorithm, for TEB to plan its trajectory, as shown in Fig.~\ref{limo_carto}. The TEB trajectory (marked in yellow) fails in the narrow space because of the inevitable errors caused by the limited resolution of grid map. Figs.~\ref{limo_traj}-\ref{limo_angular} illustrate the successful trajectory of NeuPAN (marked in blue) and its associated linear and angular speed profiles. It can be seen that the linear speed fluctuates around the target speed $0.5\,$m/s and the angular speed is rigorously constrained within the range $[-3.14, 3.14]$ rad/s, which achieves a stable and bounded control policy.

\begin{figure}[t]
  \centering
  \begin{subfigure}[t]{0.115\textwidth}
    \centering
    \includegraphics[width=0.95\textwidth]{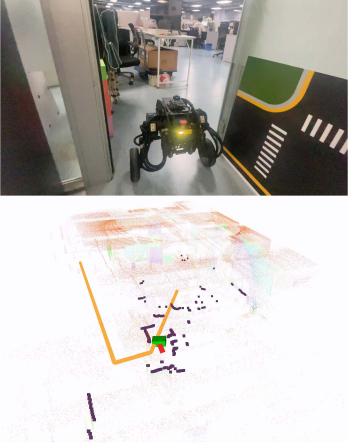}
    \caption{ Doorway.}
    \label{fig:leg1}
  \end{subfigure}
  \begin{subfigure}[t]{0.115\textwidth}
    \centering
    \includegraphics[width=0.95\textwidth]{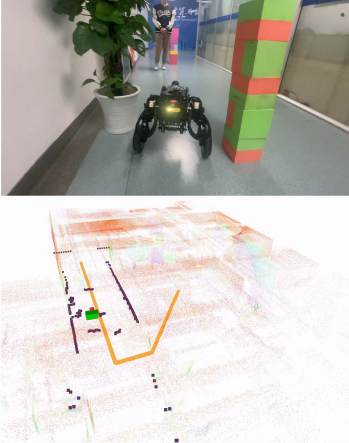}
    \caption{ Gap.}
    \label{leg2}
  \end{subfigure}
  \begin{subfigure}[t]{0.115\textwidth}
    \centering
    \includegraphics[width=0.95\textwidth]{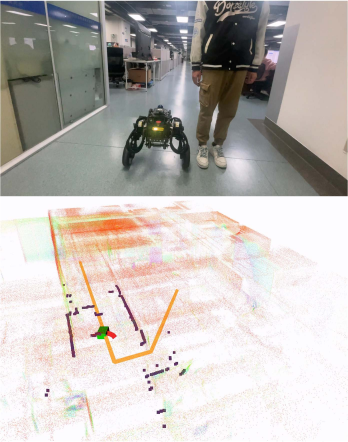}
    \caption{ Blocker.}
    \label{leg5}
  \end{subfigure}
  \begin{subfigure}[t]{0.115\textwidth}
    \centering
    \includegraphics[width=0.95\textwidth]{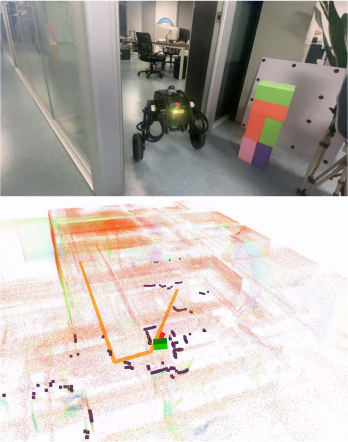}
    \caption{ Return.}
    \label{leg6}
  \end{subfigure}
  \caption{Wheel-legged mapless navigation in the office.}
  \label{exp_leg}
\end{figure}

\begin{figure}[t]
  \centering
  \begin{subfigure}[t]{0.45\textwidth}
    \centering
    \includegraphics[width=0.99\textwidth]{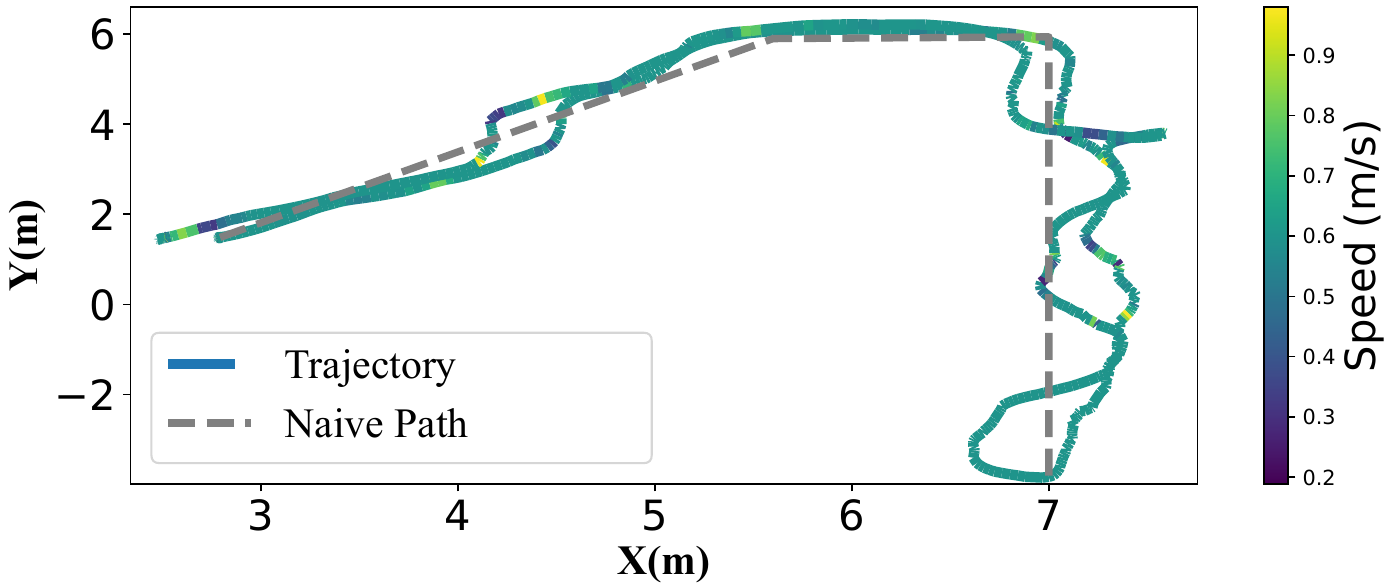}
    \caption{ Trajectory with linear speed change. }
    \label{trajectory}
  \end{subfigure}
  \begin{subfigure}[t]{0.24\textwidth}
    \centering
    \includegraphics[width=0.96\textwidth]{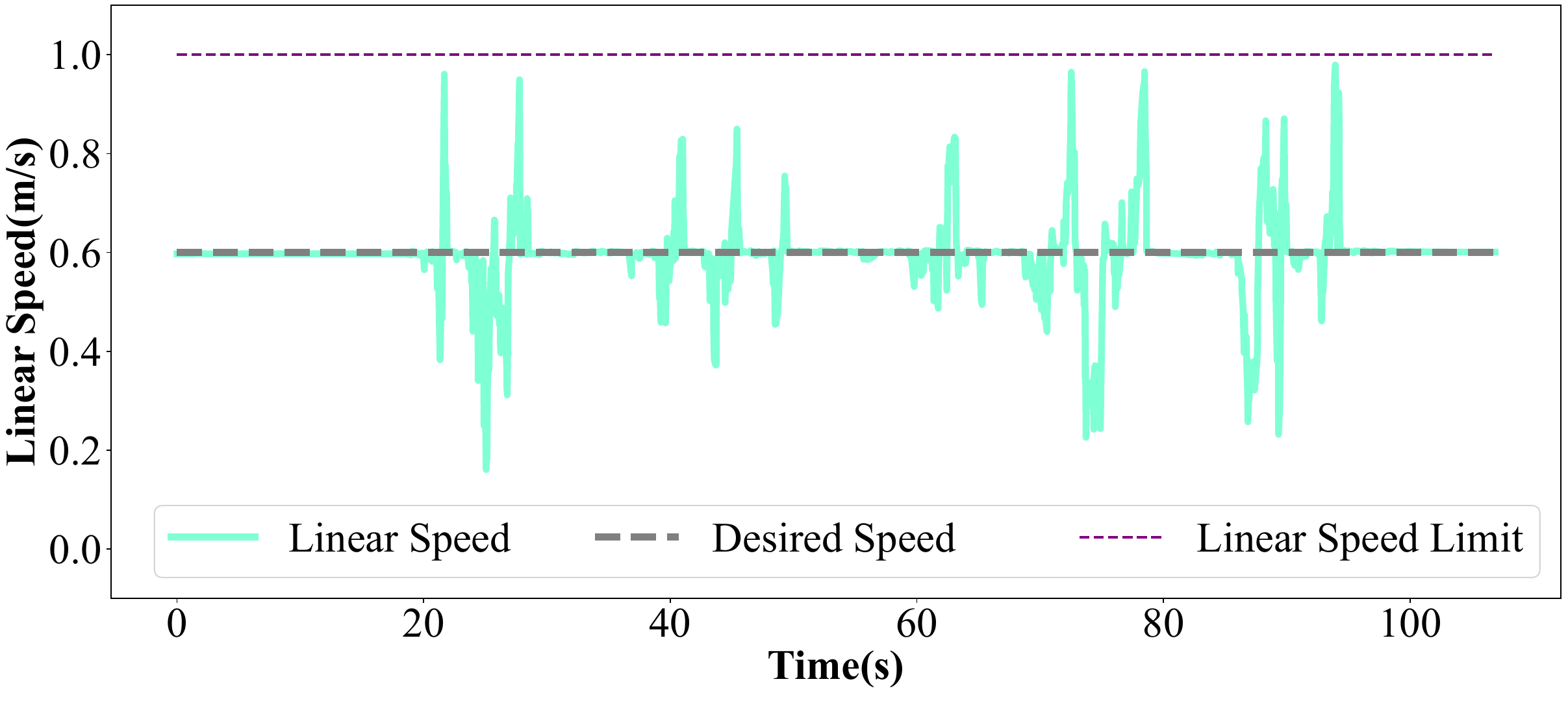}
    \caption{ Linear speed }
    \label{speed}
  \end{subfigure}
  \begin{subfigure}[t]{0.24\textwidth}
    \centering
    \includegraphics[width=0.96\textwidth]{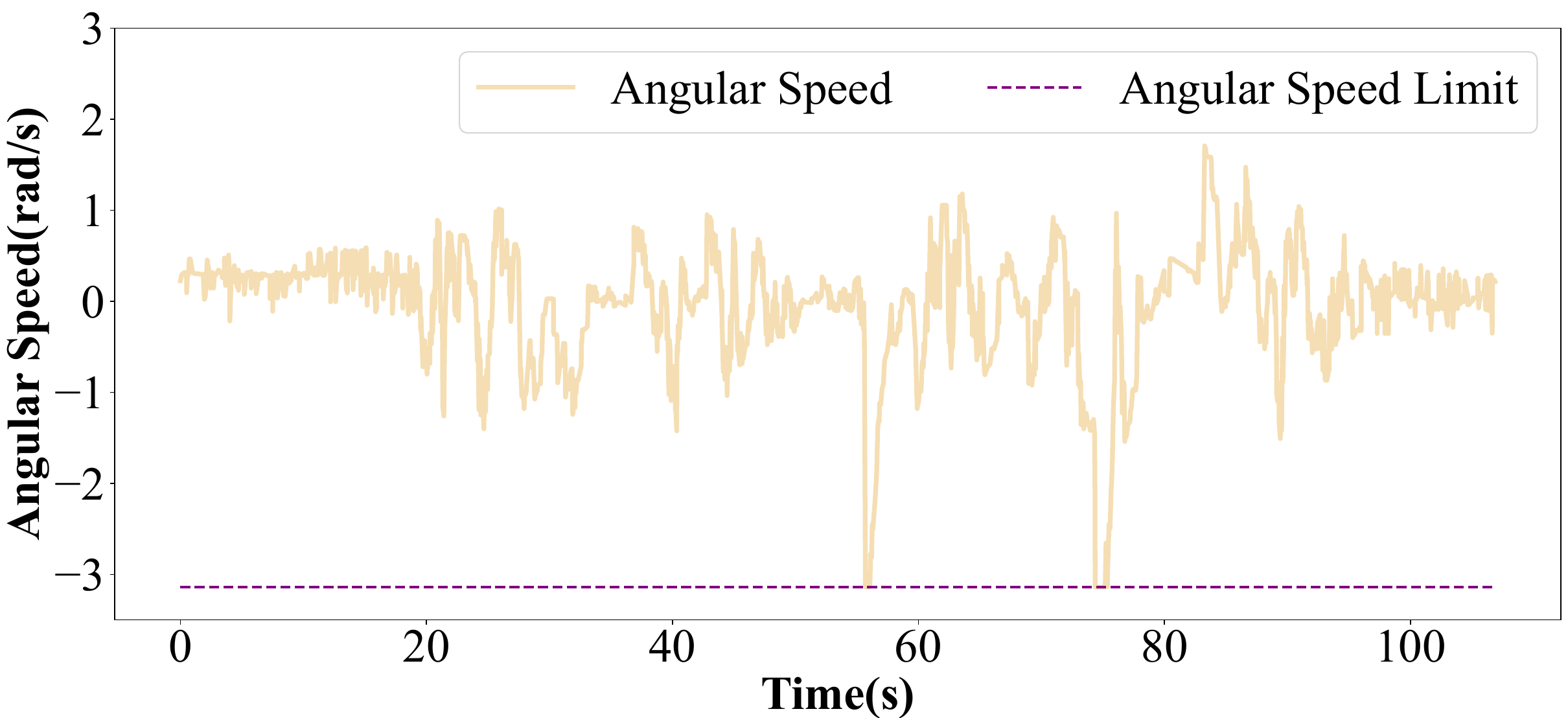}
    \caption{ Angular speed }
    \label{angular_speed}
  \end{subfigure}
  \caption{ Trajectory, linear, and angular speed change of the wheel-legged robot during the mapless navigation task. }
  \label{traj_leg}
\end{figure}

\begin{figure}[t]
  \centering
  \includegraphics[width=0.49\textwidth]{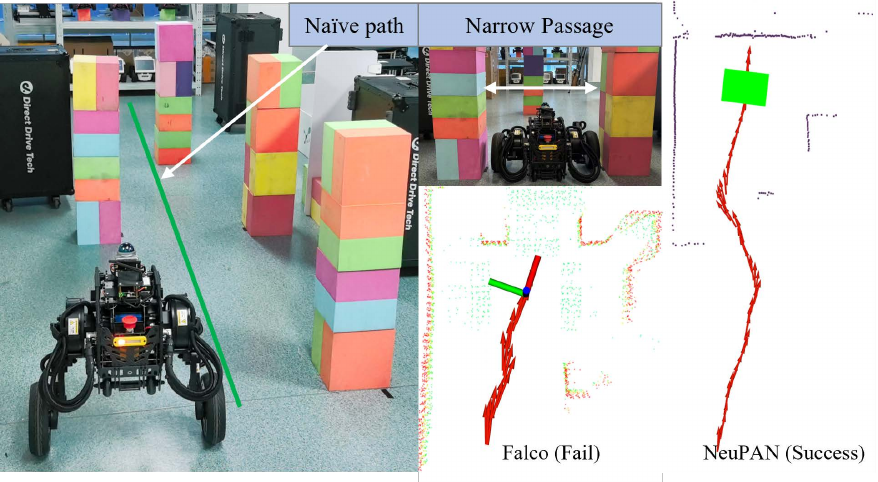}
  \caption{ Trajectory comparison of NeuPAN and Falco in the real-world confined space.}
  \label{qu_falco}
\end{figure}

\begin{table*}[tb]
  \centering
  \caption{Performance comparison of NeuPAN and Falco in the real-world confined space}
  \label{tab:falco}
  \resizebox{0.85\textwidth}{!}{%
    \begin{tabular}{ccclccccc}
      \cline{1-4} \cline{5-9}
      \multirow{2}{*}{Narrow Passage}     & \multicolumn{2}{c}{Passable} &                & \multicolumn{2}{c}{Naivgation Time} &                & \multicolumn{2}{c}{Average Move Speed}                              \\ \cline{2-3} \cline{5-6} \cline{8-9}
                                          & NeuPAN                       & Falco          &                                     & NeuPAN         & Falco                                  &  & NeuPAN          & Falco \\ \cline{1-4} \cline{5-9}
      Falco narrow threshold ($\mathsf{DoN}$: 0.79)  & \CheckmarkBold               & \CheckmarkBold &                                     & \textbf{9.51}  & 13.56                                  &  & \textbf{ 0.477} & 0.36  \\
      Neupan narrow threshold ($\mathsf{DoN}$: 0.89) & \CheckmarkBold               & \XSolidBrush   &                                     & \textbf{10.15} & -                                      &  & \textbf{0.468}  & 0.27  \\ \hline
    \end{tabular}%
  }
\end{table*}

\subsection{Experiment 4: Wheel-legged Robot Mapless Navigation}
In this subsection, we validate NeuPAN on a medium-size robot platform, i.e., the wheel-legged robot as mentioned in Fig.~\ref{fig1}, to conduct autonomous mapless navigation and dense mapping tasks in an unknown office building, as illustrated in Fig.~\ref{exp_leg}. The naive path is generated by connecting the straight line between the manually set goal states without consideration of the obstacles in the environment. Fast-lio2~\cite{xu2022fast} is employed for real-time mapping of the environment and localization of the ego robot. Note that the narrowest space has a $\mathsf{DoN}$ value of $0.92$. It can be seen from Fig.~\ref{exp_leg} that there exist $3$ challenges between the starting and end points (the same position in Fig.~\ref{leg6}): 1) passing a narrow doorway (Fig.~\ref{fig:leg1}); 2) avoiding arbitrary obstacles (the potted plant in Fig.~\ref{leg2}); 3) circumventing adversarial humans who suddenly rush over to block the way (Fig.~\ref{leg5}). Due to the precise actions afforded by NeuPAN, the wheel-legged robot successfully conquers all these challenges. This experiment demonstrates the efficacy and robustness of our approach in enhancing operational capabilities of wheel-legged robots. The changes over time including the trajectory, linear speed, and angular speed, are shown in Fig.~\ref{traj_leg}.
It is clear that the wheel-legged robot succeeds in following the goal states and returning to the starting point. The linear speed fluctuates around the desired speed $0.6\,$m/s, but is rigorously controlled below $1\,$m/s (the maximum speed requested by the user).
Similarly, the angular speed is rigorously controlled between $-3.14\,$rad and $+3.14\,$rad.
This result show that our approach is constraints guaranteed due to the interpretability and hard-bounded constraints in NRMP block of NeuPAN, which is in contrast to learning-based solutions that may output uncertain motions.

To further demonstrate the effectiveness of our approach operating in the real world, we conduct a quantitative experiment to compare the performance of NeuPAN and Falco~\cite{zhang2020falco}. 
The task is to navigate the wheel-legged robot along a straight trajectory in a highly-confined space ($\mathsf{DoN}$=0.89), as shown in Fig.~\ref{qu_falco}.
It can be seen that NeuPAN successfully navigates the robot through the narrow space, reaching the goal smoothly.
In contrast, the Falco method gets stuck in front of the narrow passage.
This is because Falco converts the lidar points into voxels for collision avoidance and determines the trajectories by maximizing the probability to reach the goal. 
The point voxelization and probablistic sampling operations result in degradation of navigation accuracy.
Quantitative results are shown in Table~\ref{tab:falco}. 
It can be seen that NeuPAN improves the $\mathsf{DoN}$ (i.e., navigation accuracy) from $0.79$ to $0.89$, corresponding to over $12.6\%$ $\mathsf{DoN}$ gain and $2$x accuracy improvement compared with Falco.
Furthermore, compared to Falco under the same challenge ($\mathsf{DoN}$=0.79), NeuPAN achieves higher moving speed ($32.5\%$ improvement) and shorter navigation time ($29.9\%$ reduction).   

\begin{figure*}[t]
  \centering
  \begin{subfigure}[t]{0.32\textwidth}
    \centering
    \includegraphics[width=0.99\textwidth]{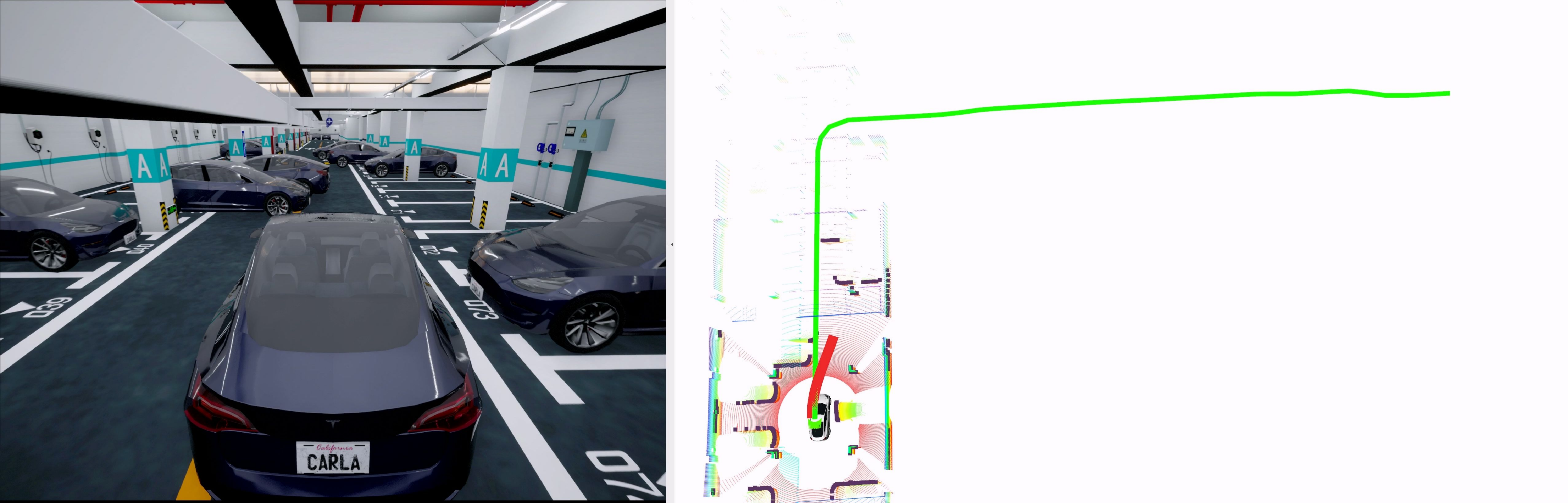}
    \caption{}
    \label{tesla1}
  \end{subfigure}
  \hfill
  \begin{subfigure}[t]{0.32\textwidth}
    \centering
    \includegraphics[width=0.99\textwidth]{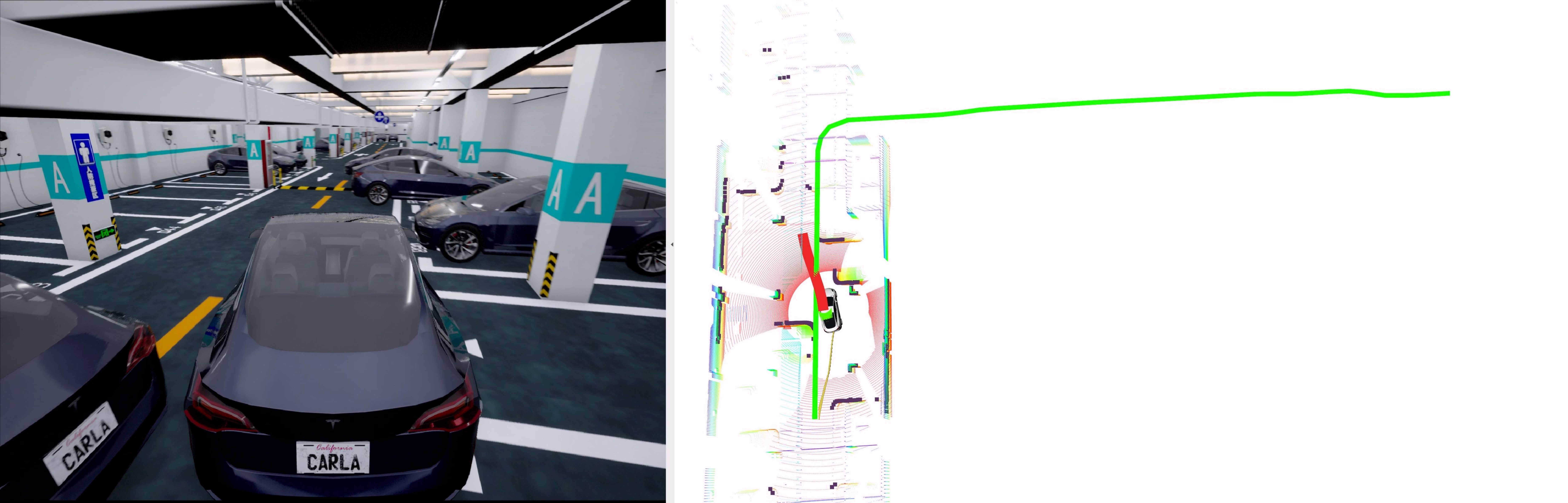}
    \caption{}
    \label{tesla2}
  \end{subfigure}
  \hfill
  \begin{subfigure}[t]{0.32\textwidth}
    \centering
    \includegraphics[width=0.99\textwidth]{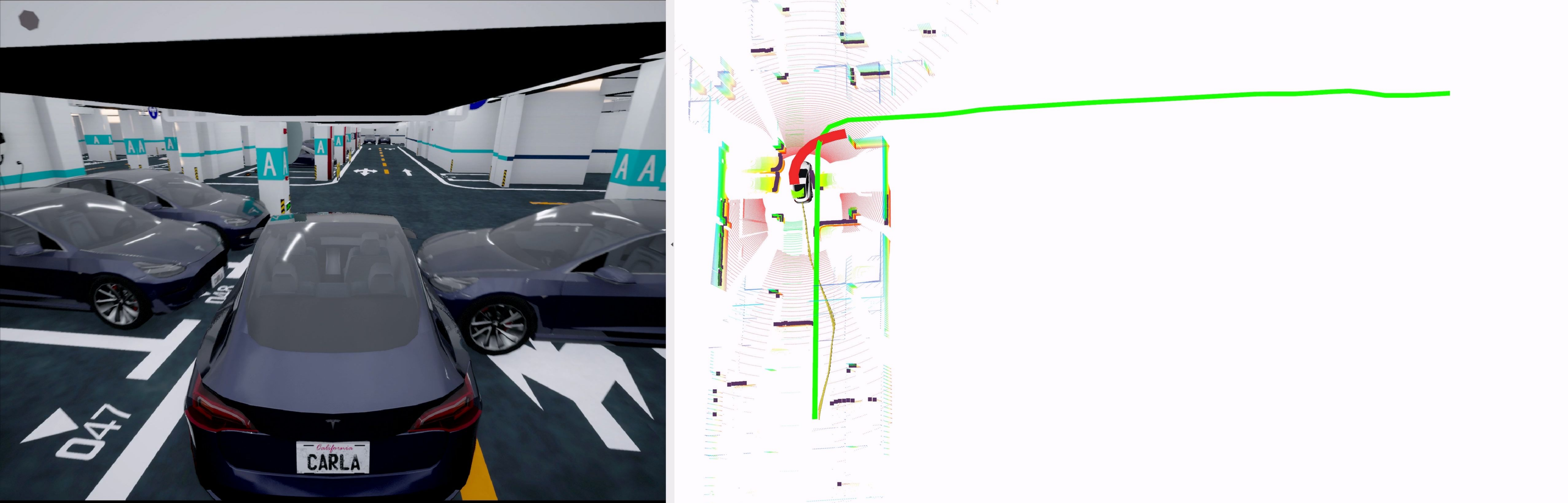}
    \caption{}
    \label{tesla3}
  \end{subfigure}
  \hfill
  \begin{subfigure}[t]{0.32\textwidth}
    \centering
    \includegraphics[width=0.99\textwidth]{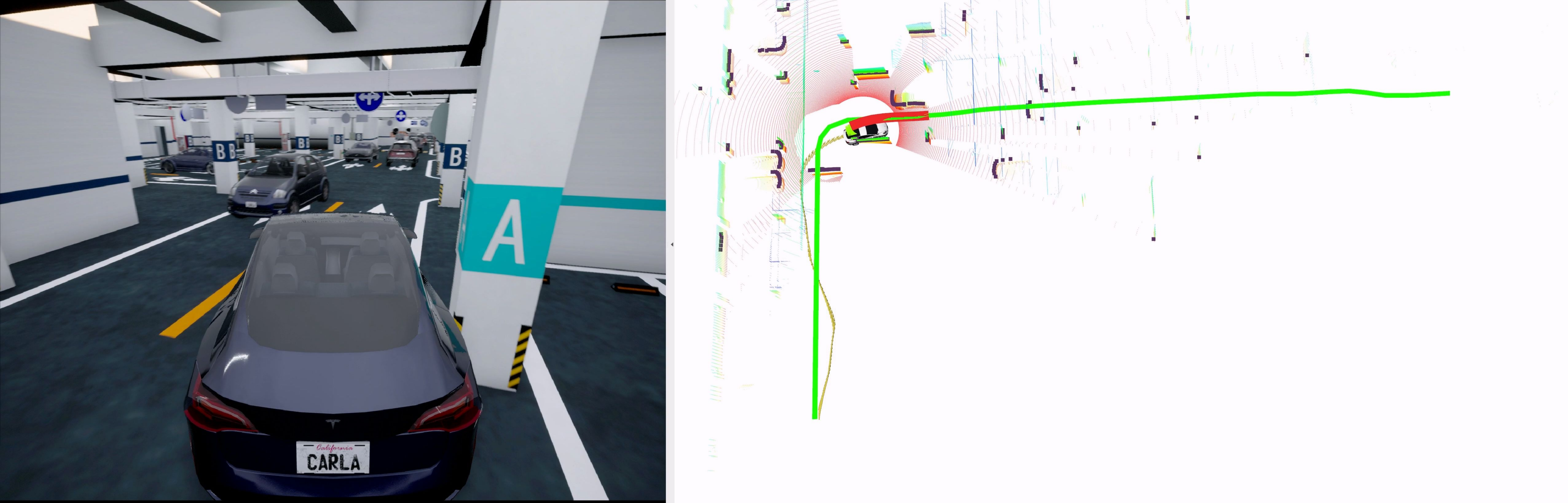}
    \caption{}
    \label{tesla4}
  \end{subfigure}
  \hfill
  \begin{subfigure}[t]{0.32\textwidth}
    \centering
    \includegraphics[width=0.99\textwidth]{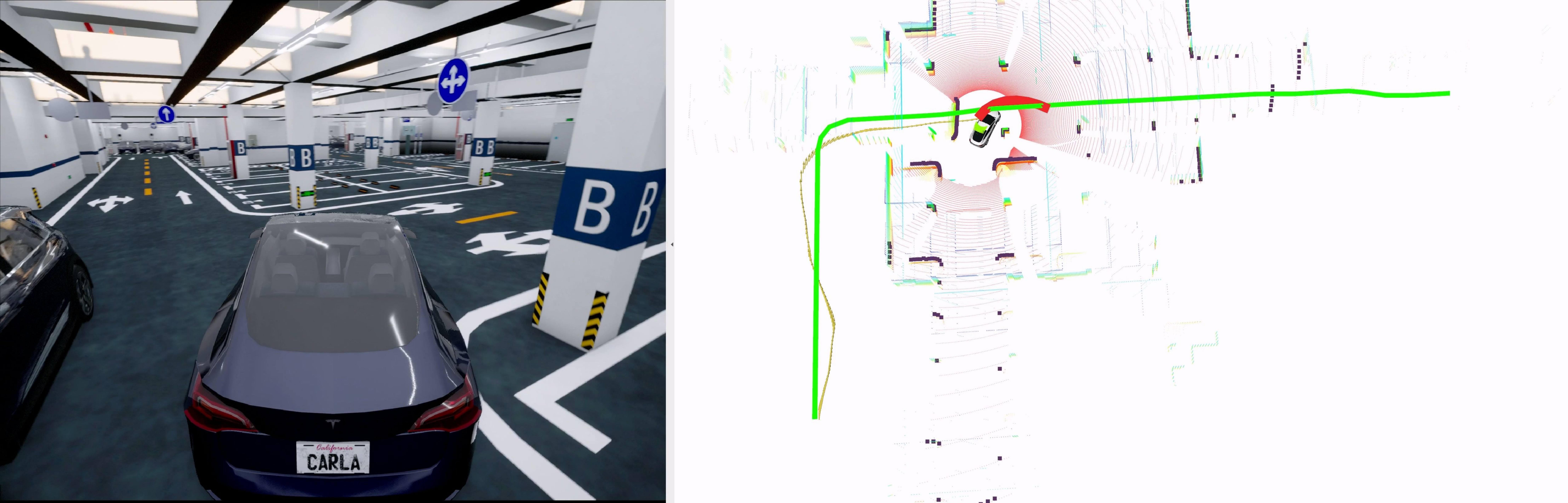}
    \caption{}
    \label{tesla5}
  \end{subfigure}
  \hfill
  \begin{subfigure}[t]{0.32\textwidth}
    \centering
    \includegraphics[width=0.99\textwidth]{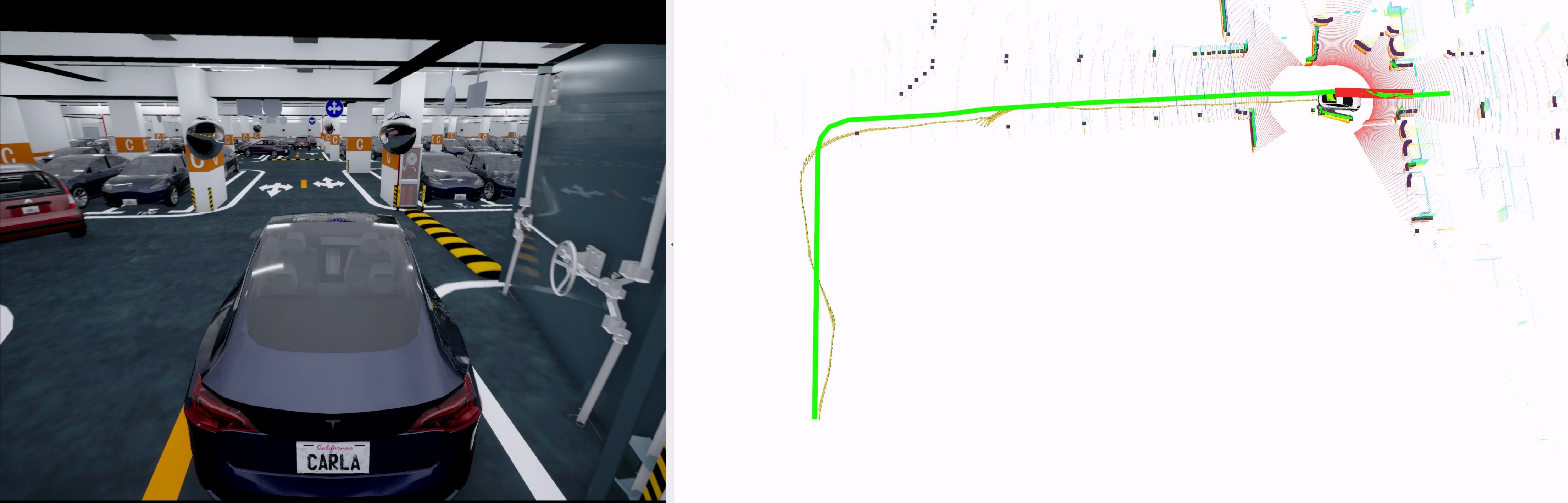}
    \caption{}
    \label{tesla6}
  \end{subfigure}
  \hfill
  \begin{subfigure}[t]{0.32\textwidth}
    \centering
    \includegraphics[width=0.99\textwidth]{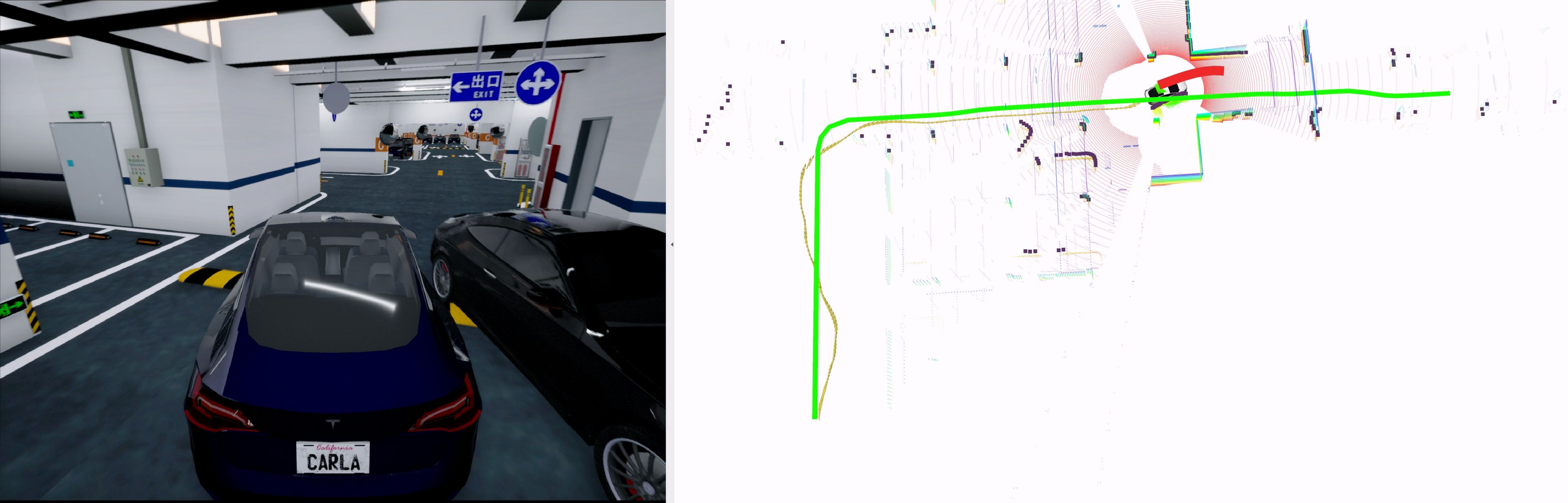}
    \caption{}
    \label{tesla7}
  \end{subfigure}
  \hfill
  \begin{subfigure}[t]{0.32\textwidth}
    \centering
    \includegraphics[width=0.99\textwidth]{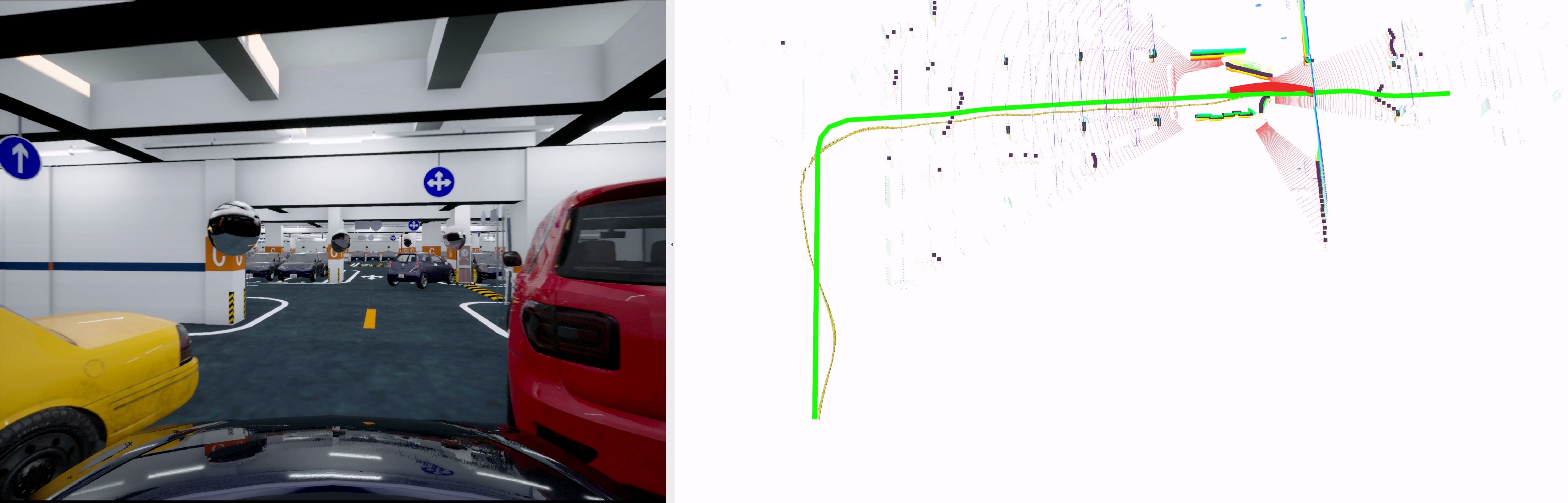}
    \caption{}
    \label{tesla8}
  \end{subfigure}
  \hfill
  \begin{subfigure}[t]{0.32\textwidth}
    \centering
    \includegraphics[width=0.99\textwidth]{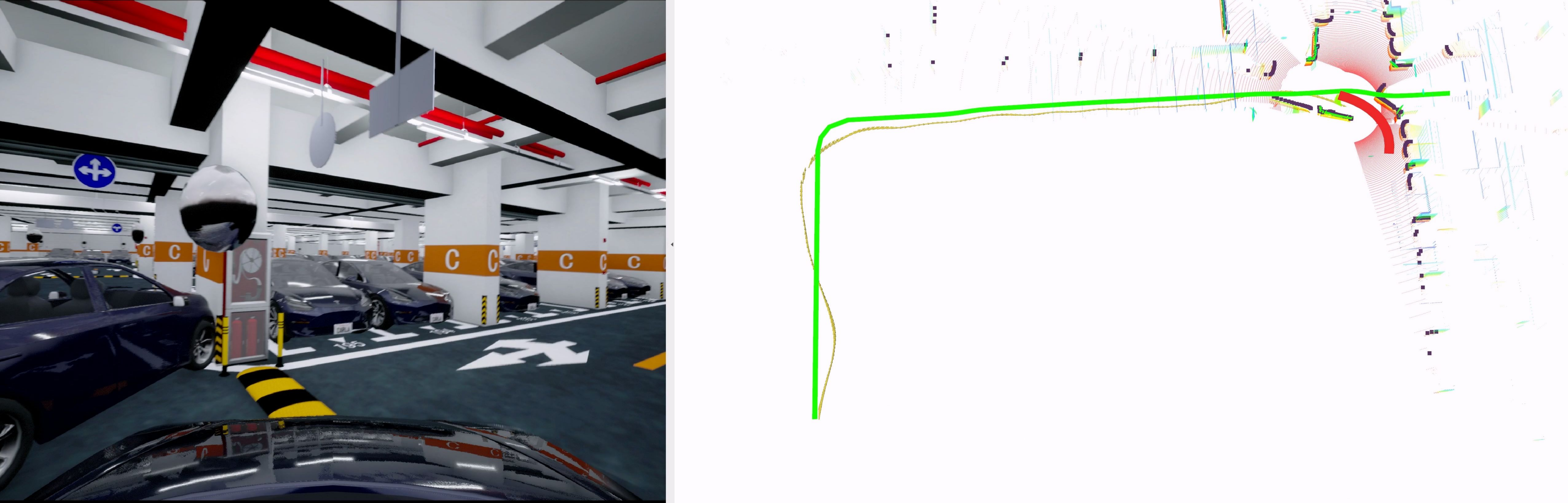}
    \caption{}
    \label{tesla9}
  \end{subfigure}
  \caption{Passenger vehicle navigation in Carla. (a)--(c): Illegally parked vehicles block the road, leaving only a narrow passable space ($\mathsf{DoN}=0.97$). (d)--(f): Randomly generated traffic flow. (g)-(i): Challenging cases in the traffic flow.}
  \label{vehicle}
\end{figure*}

\begin{figure*}[t]
  \centering
  \begin{subfigure}[t]{0.32\textwidth}
    \centering
    \includegraphics[width=0.99\textwidth]{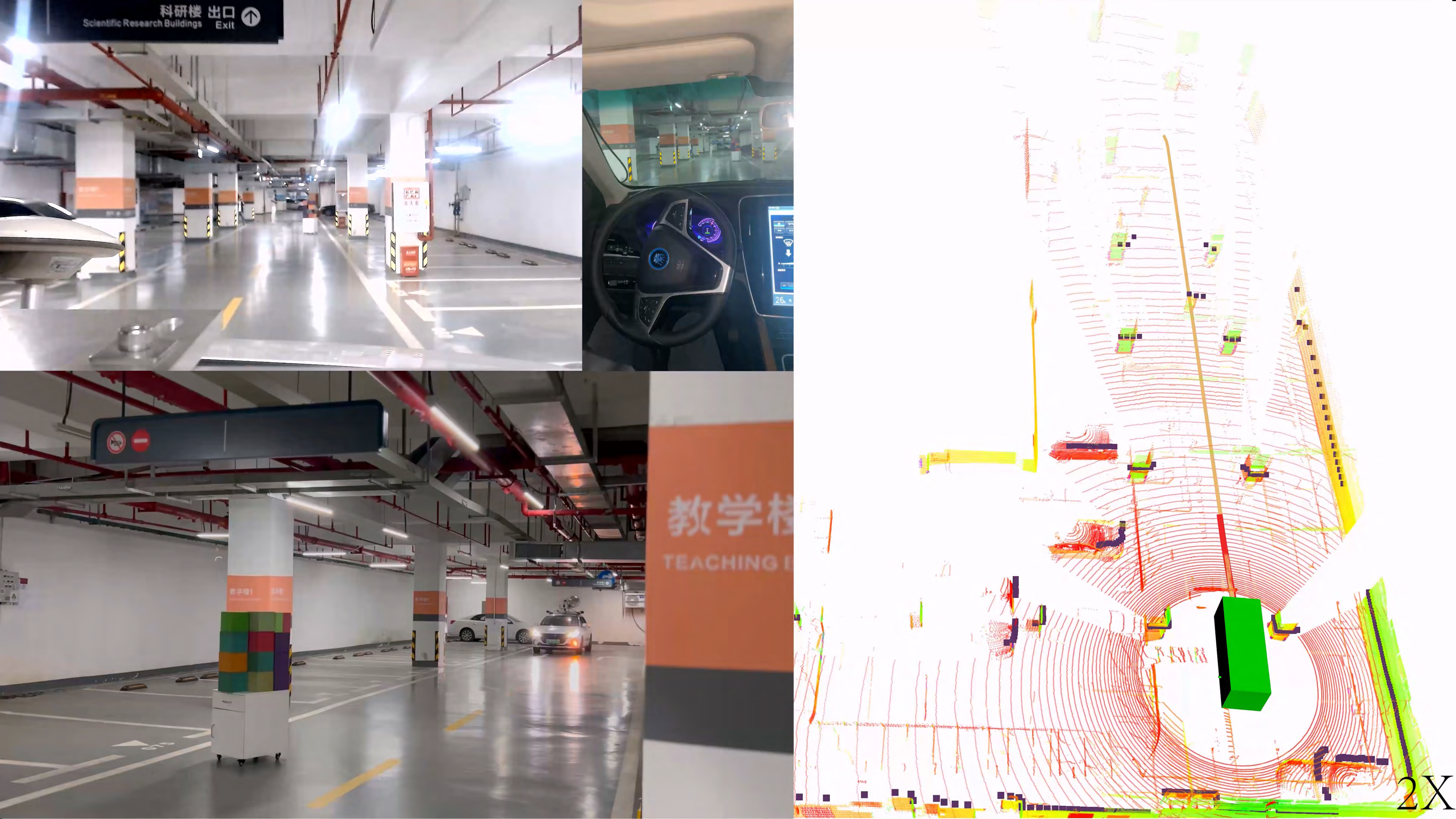}
    \caption{ Start point }
    \label{byd1}
  \end{subfigure}
  \hfill
  \begin{subfigure}[t]{0.32\textwidth}
    \centering
    \includegraphics[width=0.99\textwidth]{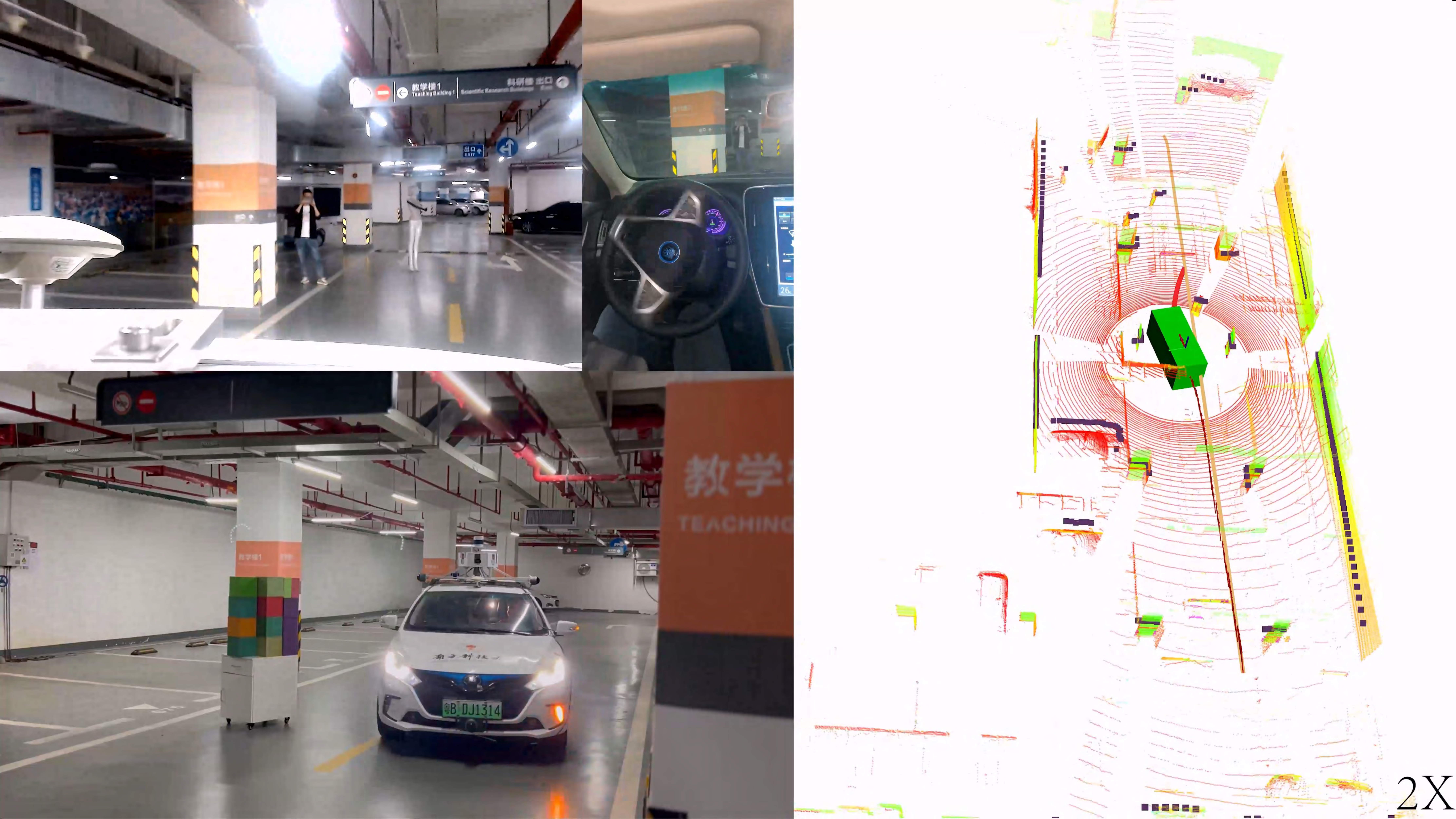}
    \caption{ Avoid boxes }
    \label{byd2}
  \end{subfigure}
  \hfill
  \begin{subfigure}[t]{0.32\textwidth}
    \centering
    \includegraphics[width=0.99\textwidth]{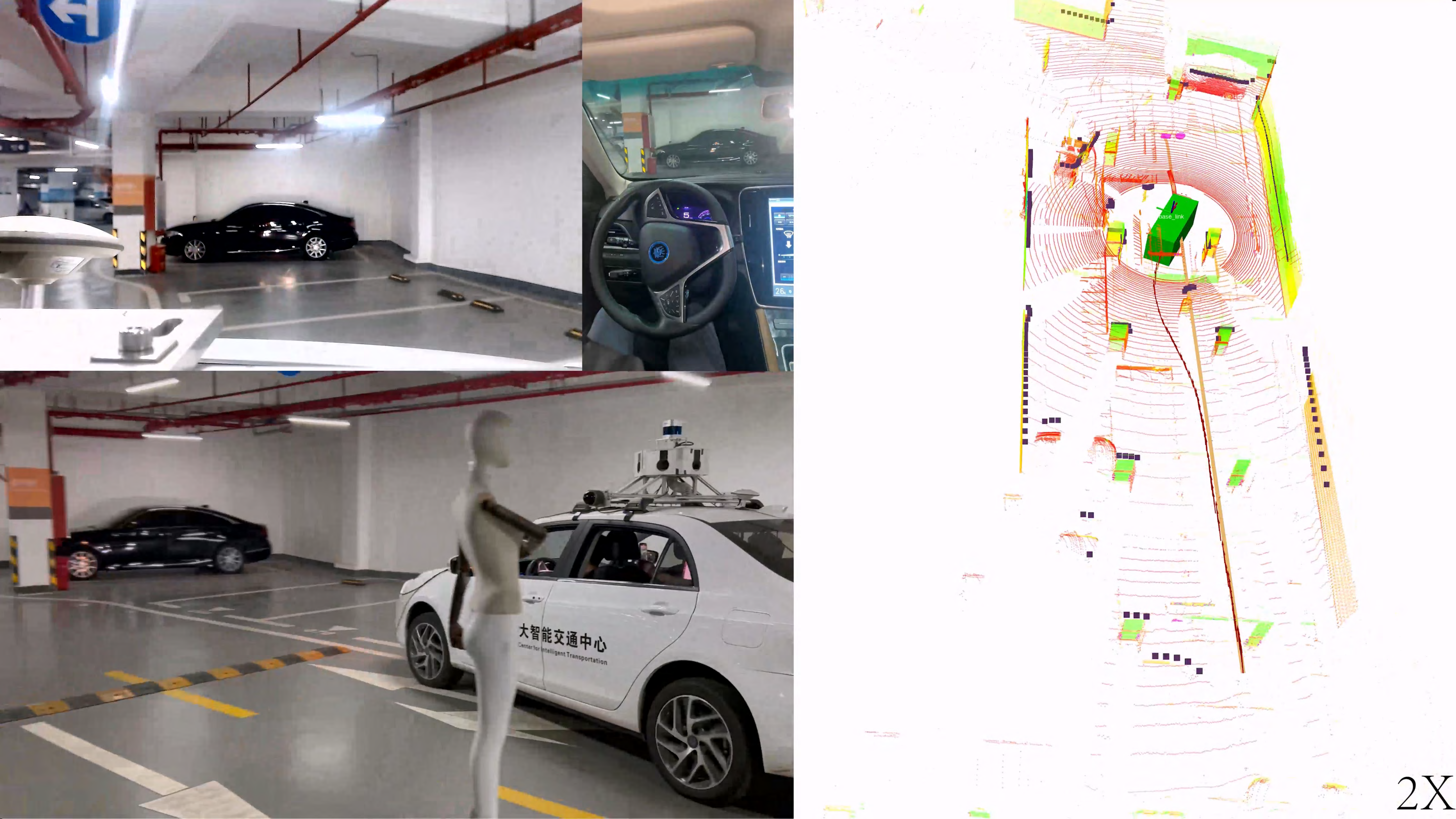}
    \caption{ Avoid mannequins }
    \label{byd3}
  \end{subfigure}
  \caption{ The real-world experiment of passenger vehicle navigation in the parking lot with boxes, mannequins, and other vehicles. }
  \label{byd}
\end{figure*}

\subsection{Experiment 5: Passenger Vehicle Navigation}

In this subsection, we validate NeuPAN on a large-size passenger vehicle platform.
Compared to the aforementioned robots, the vehicle has a larger value of speed (e.g., $10$ to $50\,$km/h) and minimum turning radius (i.e., $6\,$m).
Therefore, existing solutions for passenger vehicles adopt a larger safety distance (e.g. $1\,$m), and vehicles often stop in front of confined spaces.
In the following experiments, we will show that NeuPAN can overcome the above drawbacks.

\subsubsection{Simulated Environment}

We employ CARLA~\cite{dosovitskiy2017carla}, a high-fidelity simulator powered by unreal engine, to create a virtual parking lot scenario in Fig.~\ref{vehicle}.
A 128-line 3D lidar mounted on top of the vehicle provides real-time measurements of the environment.
The task is to navigate the vehicle from zone A to zone B, and consists of two phases.
In the first phase, as shown in Fig.~\ref{tesla3}, illegally parked vehicles block the way with $\mathsf{DoN}=0.95$.
But due to the end-to-end feature, NeuPAN successfully finds the optimal actions under the kinematics constraints in real time, and the successful trajectory is shown in Figs.~\ref{tesla1}-\ref{tesla3}.
In the second phase, as shown in Figs.~\ref{tesla4}--\ref{tesla6}, an adversarial dynamic traffic flow is generated to simulate the real world adversarial or even accidental traffics. For instance, in Fig.~\ref{tesla5}, a dangerous obstacle vehicle crash into our ego vehicle.
After the crash moment, NeuPAN generates an expert-driver-like steering and accelerating action, which rehabilitates the vehicle from a crash and makes a timely turn.
Figs.~\ref{tesla7}--\ref{tesla9} illustrate other challenging cases under the dynamic traffic flows.
Interestingly, Fig~\ref{tesla8} presents an extreme case where the gap is too narrow for any existing methods to pass.
This is because the detected boxes are slightly bigger than the actual objects, making the width of the remaining drivable area smaller than the width of ego vehicle.
However, our methodology makes impassable passable, by maneuvering the vehicle pass through this challenging case successfully, as shown in Fig~\ref{tesla8}.
This already exceeds human-level driving, as tested by our volunteer human drivers.

\subsubsection{Real World Environment}

\begin{figure}[tb]
  \centering
  \begin{subfigure}[t]{0.23\textwidth}
    \centering
    \includegraphics[width=0.99\textwidth]{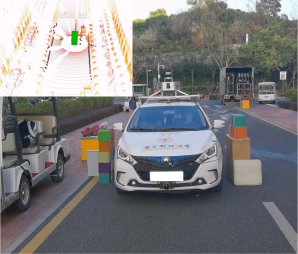}
    \caption{ NeuPAN control (success) }
    \label{byd_m1}
  \end{subfigure}
  \hfill
  \begin{subfigure}[t]{0.23\textwidth}
    \centering
    \includegraphics[width=0.99\textwidth]{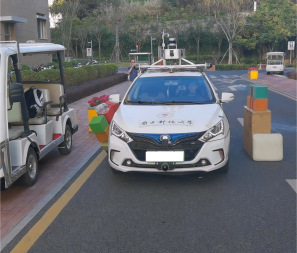}
    \caption{ Human driving (collision) }
    \label{byd_m2}
  \end{subfigure}
  \begin{subfigure}[t]{0.23\textwidth}
    \centering
    \includegraphics[width=0.99\textwidth]{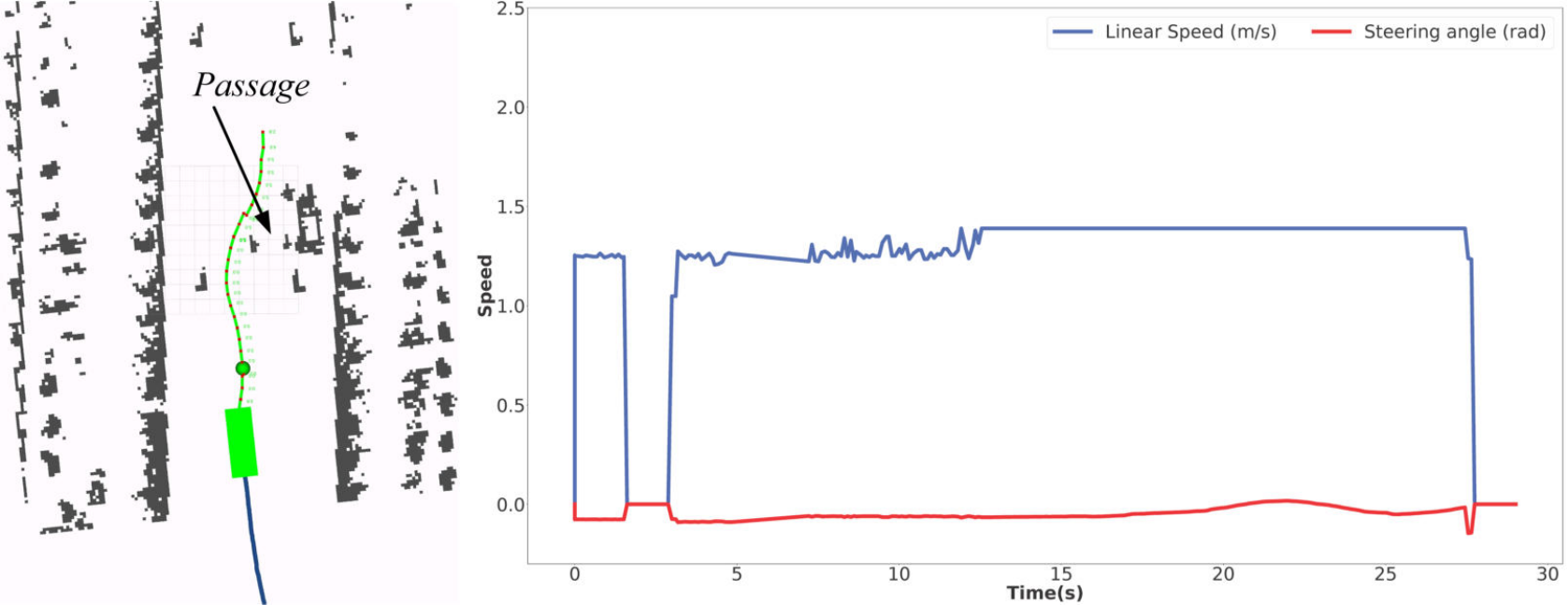}
    \caption{ Hybrid a star (Grid map).}
    \label{byd_astar}
  \end{subfigure}
  \begin{subfigure}[t]{0.23\textwidth}
    \centering
    \includegraphics[width=0.99\textwidth]{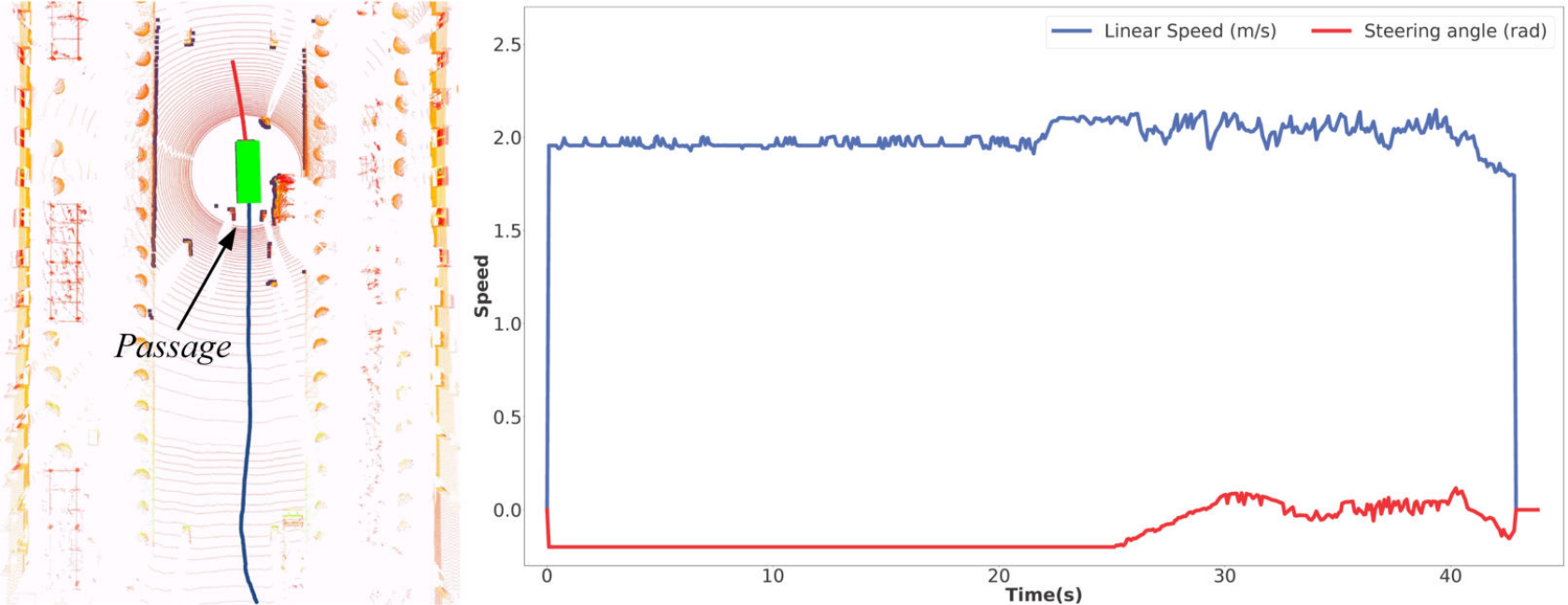}
    \caption{ NeuPAN (Point cloud). }
    \label{byd_neupan}
  \end{subfigure}
  \caption{Real-world experiment of passenger vehicle navigation through a narrow passage ($\mathsf{DoN}=0.97$). (a) NeuPAN succeeds at $7\,\mathrm{km/h}$. (b) Human driver fails at $5\,\mathrm{km/h}$. (c) The trajectory, linear speed, and steering angle of the Hybrid A star guided passenger vehicle. (d) The trajectory, linear speed and steering angle of the NeuPAN guided passenger vehicle.}
  \label{byd_manual}
\end{figure}
In existing works, numerous methodologies, such as end-to-end RL, have demonstrated great potential in simulated environments. Nevertheless, these approaches face difficulties when applied to real-world scenarios due to the sim-to-real gap. Real-world testing can also verify the robustness of NeuPAN since hardware uncertainties (e.g., noises, impairments) are now considered.

First, we test the vehicle with a 128-line lidar in the parking lot, as shown in Fig.~\ref{byd}.
Unlike the wide-open environment on the road, the parking lot leaves less space for the vehicle to pass ($\mathsf{DoN}$=0.93). Existing solutions that adopt object detection or occupancy grids are prone to take a detour or get stuck. 
In contrast, NeuPAN successfully controls the vehicle to move with an S-shaped trajectory to avoid boxes and mannequins as illustrated in Figs.~\ref{byd1}--\ref{byd3}.

Second, we conduct a simple yet specially designed experiment to validate the performance of NeuPAN. In this challenge, the vehicle needs to pass an extremely narrow passage with a width of $1.82\,$m within a certain time budget (corresponding to an average speed of $5\,$km/h).
As the vehicle width is already $1.77\,$m, there is only about $5$ centimeters tolerance ($\mathsf{DoN}$=0.97).
We compare our NeuPAN with: 1) hybrid A star implemented by Autoware~\cite{kurzer2016path, kato2018autoware}, a state-of-the-art open-source autonomous driving solution, and 2) human driving. 
First, it can be seen that even under a $\mathsf{DoN}$ close to 1, our NeuPAN solution successfully navigates the vehicle through the gap at a speed of $7\,$km/h, as shown in Fig.~\ref{byd_m1}.
Second, the human driver is very likely to fail given a speed of $5\,$km/h. Only when we reduce the speed to a lower value (e.g. $2\,$km/h) can the human driver tackle this situation by careful observations, as shown in Fig~\ref{byd_m2}.
Lastly, the Autoware solution, i.e., hybrid A star, converts the lidar scans into occupancy grids, as shown in Fig.~\ref{byd_astar}. This approach treats the narrow gap as impassable, thereby generating alternative directions for detouring.
In contrast, our NeuPAN solution directly utilizes the lidar points as input, achieving higher accuracy and successfully finding the optimal path, as illustrated in Fig.~\ref{byd_neupan}.
The above experiments demonstrate the advantage of the proposed framework in extremely cluttered environments. This advantage makes our approach very promising in handling various difficult cases.

\section{Conclusion}\label{section 7}

This article proposes NeuPAN, an end-to-end model-based learning approach for direct point robot navigation. 
A novel tightly-coupled perception-to-control framework, including DUNE and NRMP, is developed to directly map the raw points to the robot actions without error propagation. Owing to the interpretable deep unfolding neural network, DUNE can be trained in a sim-to-real fashion to convert obstacle points to latent distance features rapidly. The output distance features are embedded as neural regularizers to generate collision-free and time-efficient robot actions by NRMP, represented by differentiable convex optimization layers. Exhaustive experiments on multiple robot platforms in diverse scenarios are conducted to validate NeuPAN, which show that NeuPAN outperforms existing state-of-the-art approaches with over $2$x accuracy improvement, reduced navigation time, higher average speed, better robustness and generalization capabilities. Since NeuPAN generates perception-aware and physically interpretable motions in the real-time, it empowers autonomous systems to work in cluttered environments that are previously considered impassable, triggering new applications such as cluttered-room housekeeping and limited-space operating.

\bibliographystyle{IEEEtran}
\bibliography{reference/Thesis}



\appendices

\section{Ackermann and Differential Models} \label{Models}

Car-like robots adhere to the Ackermann model has the nonlinear kinematic function~\cite{lynch2017modern}:
$$f({\mathbf{s}_t},{\mathbf{u}_t}) = \left[
    {{v_t}\cos ({\theta _t})}, 
    {{v_t}\sin ({\theta _t})}, 
    {\frac{{{v_t}\tan {\psi _t}}}{L}}
  \right]^T,$$
where $v_t$ and $\psi_t$ are the linear speed and steering angle of the robot at time $t$, respectively. $L$ is the wheelbase. $\theta _t$ is the orientation of the robot. This non-linear function can be linearized into the structure of Eq.~\eqref{kinematics} by the first-order Taylor polynomial, where the associated coefficients $(\mathbf{A}_t$, $\mathbf{B}_t$, $\mathbf{c}_t)$ at time $t$ in Eq.~\eqref{kinematics} are given by:
\begin{equation}
\begin{aligned}
    \mathbf{A}_t &= \left[ {\begin{array}{*{20}{c}}
    1&0&{ - \bar{v}_t\sin (\bar{\theta}_t) {\Delta t} }\\
    0&1&{\bar{v}_t\cos (\bar{\theta}_t) {\Delta t}}\\
    0&0&{1}
    \end{array}} \right], 
    \\
    \mathbf{B}_t &= \left[ {\begin{array}{*{20}{c}}
    {\cos (\bar{\theta}_t) {\Delta t} }&0
    \\
    {\sin (\bar{\theta}_t) {\Delta t} }&0
    \\
    {\frac{{\tan \bar{\psi}_t {\Delta t}}}{L}}&{\frac{\bar{v}_t {\Delta t}}{{L{{\cos }^2}\bar{\psi}_t }}}
    \end{array}} \right],
    \\ 
    \mathbf{c}_t &= \left[ {\begin{array}{*{20}{c}}
    {\bar{\theta}_t \bar{v}_t\sin (\bar{\theta}_t) {\Delta t} }\\
    { - \bar{\theta}_t \bar{v}_t\cos (\bar{\theta}_t) {\Delta t} }\\
    { - \frac{{\bar{\psi}_t \bar{v}_t {\Delta t}}}{{L{{\cos }^2}\bar{\psi_t} }}}
    \end{array}} \right],
    \label{car_kinematics}
\end{aligned}
\end{equation}
where $\bar{v}_t$ and $\bar{\psi}_t$ are the nominal linear speed and steering angle of the control vector $\mathbf{\bar{u}}_t$ at time $t$, respectively. Similarly, the differential drive robots with the control vector: linear velocity $v_t$ and angular velocity $\omega_t$ have the kinematic function $f({\mathbf{s}_t},{\mathbf{u}_t}) = \left[
    {{v_t}\cos ({\theta _t})}, 
    {{v_t}\sin ({\theta _t})}, 
    {\omega_t}
  \right]^T$, can be modeled to be a linear form by the following coefficients:
\begin{equation}
  \begin{aligned}
    \mathbf{A}_t &= \left[ {\begin{array}{*{20}{c}}
    1&0&{ - \bar{v}_t\sin (\bar{\theta}_t) {\Delta t} }\\
    0&1&{\bar{v}_t\cos (\bar{\theta}_t) {\Delta t}}\\
    0&0&{1}
    \end{array}} \right], 
    \\
    \mathbf{B}_t &= \left[ {\begin{array}{*{20}{c}}
    {\cos (\bar{\theta}_t) {\Delta t} }&{0}
    \\
    {\sin (\bar{\theta}_t) {\Delta t} }&{0}
    \\
    {0}&{\Delta t}
    \\
    \end{array}} \right],
    \\ 
    \mathbf{c}_t &= \left[ {\begin{array}{*{20}{c}}
    {\bar{\theta}_t \bar{v}_t\sin (\bar{\theta}_t) {\Delta t} }\\
    { - \bar{\theta}_t \bar{v}_t\cos (\bar{\theta}_t) {\Delta t} }\\
    { 0 }
    \end{array}} \right],
    \label{diff_kinematics}
\end{aligned}
\end{equation}

\section{Equivalence between P and Q} \label{Equivalence}

To prove this, we substitute $D_{i,t}$ into ${\bf{dist}}(\mathbb{Z}_t(\mathbf{s}_t), \mathbb{P}_t)$ in Eq.~\eqref{dist}. Then, the constraint Eq.~\eqref{ps_b} in $\mathsf{P}$ is equivalent to
\begin{subequations}
\begin{align}
    {\bf{dist}}(\mathbb{Z}_t(\mathbf{s}_t), \mathbb{P}_t;\{{\bm{\mu}}_t^i,{\bm{\lambda}}_t^i\})
    &\geq d_{\rm{min}}  \iff
    \nonumber\\
    \mathop{\min_{i=1,\cdots,M}}~
    \left\{
    {{\bm{\lambda}}_t^i}^\top\left[\mathbf{t}_t(\mathbf{s}_t)-\mathbf{p}_t^i\right]
    -{{\bm{\mu}}_t^i} ^\top \mathbf{h} 
    \right\}&\geq d_{\rm{min}}, \label{dist2}
    \\
    {\bm{\mu}_t^i}^\top\mathbf{G} + {\bm{\lambda}_t^i}^\top\mathbf{R}(\mathbf{s}_t)&=\bm{0},   \label{lambda2} \\
    \{\bm{\mu}_t^i,{\bm{\lambda}_t^i}\} \in\mathcal{G}.
\end{align}
\end{subequations}
Based on the definitions of $I$ in Eq.~\eqref{penaltyI} and $E$ in Eq.~\eqref{penaltyH}, constraints Eq.~\eqref{dist2} and Eq.~\eqref{lambda2} are equivalently re-written as
\begin{align}
\min\left\{I(\mathbf{s}_{t},{{\bm{{\mu}}}}_{t}^i,{\bm{\lambda}}_{t}^i), 0\right\}=0,~E(\mathbf{s}_{t},{{\bm{{\mu}}}}_{t}^i,{\bm{\lambda}}_{t}^i) =0. \label{equalities}
\end{align}
According to the penalty theory, the equality constraints Eq.~\eqref{equalities} can be removed and automatically satisfied by adding penalty terms $\|\min(I,0)\|_2^2+\|E\|_2^2$ for all $(i,t)$ to the objective of $\mathsf{P}$. 
Then problem $\mathsf{P}$ is equivalently transformed into $\mathsf{Q}$.

\section{Proof of Theorem 1} \label{appendix proof}

To prove the theorem, we first define $\mathcal{X}=\{\mathcal{S},\mathcal{U}\}$ and 
$\mathcal{Y}=\{\mathcal{M},\mathcal{L}\}$.
Let $f(\mathcal{X})= C_0(\mathcal{S},\mathcal{U})+\mathbb{I}_{\mathcal{F}}(\mathcal{S},\mathcal{U})$ and $g(\mathcal{Y})= \mathbb{I}_{\mathcal{G}}(\mathcal{S},\mathcal{U})$, where $\mathbb{I}_{\mathcal{F}},\mathbb{I}_{\mathcal{G}}$ are indicator functions for the feasible sets $\mathcal{F},\mathcal{G}$.
Then problem $\mathcal{Q}$ is equivalently written as
\begin{align}
    \mathsf{M}:\min_{\substack{\mathcal{X},\mathcal{Y}}}~~&
\underbrace{f(\mathcal{X})+g(\mathcal{Y})+C_r(\mathcal{X},\mathcal{Y})}_{:=
    \Psi(\mathcal{X},\mathcal{Y})}
\end{align}
It can be seen that $\Psi$ is an extended proper and lower semicontinuous function of $C_{\mathrm{e2e}}$, by incorporating constraints as indicator functions into $C_{\mathrm{e2e}}$, and $C_r$ is a continuous $C^1$ function.
Based on Sections IV and V, it is clear that NeuPAN is equivalent to the following alternating discrete dynamical system in the form 
of $\mathcal{X}^{[k]},\mathcal{Y}^{[k]}\rightarrow \mathcal{X}^{[k]},\mathcal{Y}^{[k+1]}\rightarrow
\mathcal{X}^{[k+1]},\mathcal{Y}^{[k+1]}$ (given the initial $\mathcal{X}^{[0]},\mathcal{Y}^{[0]}$):
\begin{subequations}
\begin{align}
     \mathcal{Y}^{[k+1]}&=     
     \mathop{\mathrm{argmin}}_{\mathcal{Y}}~\Big\{{\widehat\Psi}(\mathcal{X}^{[k]},\mathcal{Y}) \Big\},~~
     \mathsf{(DUNE)},
     \label{y(k+1)}
     \\
     \mathcal{X}^{[k+1]}&=     
     \mathop{\mathrm{argmin}}_{\mathcal{X}}~\Big\{\Psi(\mathcal{X},\mathcal{Y}^{[k+1]})
     \nonumber\\
     &\quad\quad\quad\quad\quad
+\frac{b_k}{2}\|\mathbf{x}-\mathbf{x}^{[k]}\|_2^2\Big\}
,~~
     \mathsf{(NRMP)},
     \label{x(k+1)}
\end{align}
\end{subequations}
where $\mathbf{x}=\mathrm{vec}(\mathcal{X})$ and $\mathbf{y}=\mathrm{vec}(\mathcal{Y})$ are vectorization of sets $\mathcal{X},\mathcal{Y}$, and $\{b_k\}$ are positive real numbers.
The function ${\widehat\Psi}= f+g+\widehat{C}_r$, where $\widehat{C}_r$ is an universal approximate of $C_r$ with the differences between their function values and gradients bounded as
\begin{subequations}
\begin{align}
&
\|
\partial C_r(\mathcal{X},\mathcal{Y})
-\partial{\widehat C_r}(\mathcal{X},\mathcal{Y})
\|_2
\leq \epsilon, \label{epsilon1}
\\
&\|C_r(\mathcal{X},\mathcal{Y})
-{\widehat C_r}(\mathcal{X},\mathcal{Y})
\|_2
\leq \epsilon,
\end{align}
\end{subequations}
where $\epsilon\rightarrow 0$ as the number of layers of DUNE goes to infinity.

\subsection{Proof of Part (i)}

With the above notations and assumptions, we now prove part (i).
First, substituting $\mathcal{Y}^{[k+1]}$ and $\mathcal{Y}^{[k]}$ into Eq.~\eqref{y(k+1)}, we have 
\begin{align}
{\widehat\Psi}(\{\mathcal{X}^{[k]},\mathcal{Y}^{[k]}\})
&\geq{\widehat\Psi}(\{\mathcal{X}^{[k]},\mathcal{Y}^{[k+1]}\}).
\label{appendix-1}
\end{align}
Putting $\mathcal{X}^{[k+1]}$ and $\mathcal{X}^{[k]}$ into Eq.~\eqref{x(k+1)}, we have 
\begin{align}
\Psi(\{\mathcal{X}^{[k]},\mathcal{Y}^{[k+1]}\})
&\geq\Psi(\{\mathcal{X}^{[k+1]},\mathcal{Y}^{[k+1]}\})
\nonumber\\
&\quad{} +\frac{b_k}{2}\|\mathbf{x}^{[k+1]}-\mathbf{x}^{[k]}\|_2^2 \}.
\label{appendix-2}
\end{align}
Combining the above two equations Eq.~\eqref{appendix-1}Eq.~\eqref{appendix-2} leads to
\begin{align}
&
{\Psi}(\{\mathcal{X}^{[k]},\mathcal{Y}^{[k]}\})-
\Psi(\{\mathcal{X}^{[k+1]},\mathcal{Y}^{[k+1]}\})
\geq
\nonumber\\
&
\frac{b_k}{2}\|\mathbf{x}^{[k+1]}-\mathbf{x}^{[k]}\|_2^2
-
\frac{\widehat{L}_2+L_2}{2}\|\mathbf{y}^{[k+1]}-\mathbf{y}^{[k]}\|_2^2,
\end{align}
where $L_2$ and $\widehat{L}_2$ are Lipschitz constants of $C_r$ and $\widehat{C}_r$, respectively. 
Since $C_0$ and $C_r$ are norm functions, it is clear that $L_2$ and $\widehat{L}_2$ are finite numbers.
Now we consider two cases.
\begin{itemize}
    \item $\|\mathbf{x}^{[k+1]}-\mathbf{x}^{[k]}\|_2\neq 0$. In this case, there always exists some $b_k$ such that $b_k>(\widehat{L}_2+L_2)\frac{\|\mathbf{y}^{[k+1]}-\mathbf{y}^{[k]}\|_2^2}{\|\mathbf{x}^{[k+1]}-\mathbf{x}^{[k]}\|_2^2}$. 
By setting 
\begin{align}
\varrho = &
\left[b_k-
(\widehat{L}_2+L_2)\frac{\|\mathbf{y}^{[k+1]}-\mathbf{y}^{[k]}\|_2^2}{\|\mathbf{x}^{[k+1]}-\mathbf{x}^{[k]}\|_2^2}\right]
\nonumber\\
&\times
\left(
1+\frac{\|\mathbf{y}^{[k+1]}-\mathbf{y}^{[k]}\|_2^2}{\|\mathbf{x}^{[k+1]}-\mathbf{x}^{[k]}\|_2^2}
\right)^{-1}
>0,
\end{align}
we obtain 
\begin{align}
&
{\Psi}(\{\mathcal{X}^{[k]},\mathcal{Y}^{[k]}\})-
\Psi(\{\mathcal{X}^{[k+1]},\mathcal{Y}^{[k+1]}\})
\geq
\nonumber\\
&
\frac{\varrho}{2}\left[
\|\mathbf{x}^{[k+1]}-\mathbf{x}^{[k]}\|_2^2
+
\|\mathbf{y}^{[k+1]}-\mathbf{y}^{[k]}\|_2^2\right].
\label{Psi_induction}
\end{align}
    \item $\|\mathbf{x}^{[k+1]}-\mathbf{x}^{[k]}\|_2=0$. 
    In this case, $\|\mathbf{y}^{[k+1]}-\mathbf{y}^{[k]}\|_2=0$ must hold. This is because $\mathbf{y}=\mathrm{vec}(\mathcal{Y})=\mathrm{vec}(\{\mathcal{M},\mathcal{L}\})$ represents point-to-shape distance, and is uniquely defined by the robot state $\mathbf{x}$. The distances associated with two equivalent robot states are guaranteed to be the same.
    This is exactly why we can abandon proximal terms in DUNE.
    This implies that Eq.~\eqref{Psi_induction} also holds.
\end{itemize}
Combining the above two cases, the proof for part (i) is completed.

\subsection{Proof of Part (ii)}

To prove part (ii), we sum up Eq.~\eqref{Psi_induction} from $k=0$ to $N-1$, where $N$ is a positive integer, and obtain
\begin{align}
&
\sum_{k=0}^{N-1}
\left(
\|\mathbf{x}^{[k+1]}-\mathbf{x}^{[k]}\|_2^2
+
\|\mathbf{y}^{[k+1]}-\mathbf{y}^{[k]}\|_2^2
\right)
\nonumber\\
&\leq
\frac{2}{\varrho}
\left[
{\Psi}(\mathcal{X}^{[0]},\mathcal{Y}^{[0]})-
\Psi(\mathcal{X}^{[N]},\mathcal{Y}^{[N]})
\right].
\end{align}
Taking $N\rightarrow\infty$, we have 
\begin{align}
&
\sum_{k=0}^{\infty}
\left(
\|\mathbf{x}^{[k+1]}-\mathbf{x}^{[k]}\|_2^2
+
\|\mathbf{y}^{[k+1]}-\mathbf{y}^{[k]}\|_2^2
\right)< \infty,
\end{align}
which gives 
$\lim_{k\rightarrow \infty}
\|\mathbf{x}^{[k+1]}-\mathbf{x}^{[k]}\|_2
\rightarrow 0$.

\subsection{Proof of Part (iii)}

To prove part (iii), we first take the sub-differential on both sides of Eq.~\eqref{y(k+1)}Eq.~\eqref{x(k+1)}, which yields
\begin{subequations}
\begin{align}
&
\partial_{\mathbf{y}}{\widehat\Psi}(\mathbf{x}^{[k]},\mathbf{y}^{[k+1]})
+c_k(\mathbf{y}^{[k+1]}-\mathbf{y}^{[k]})
=0,
\\
&
\partial_{\mathbf{x}}{\Psi}(\mathbf{x}^{[k+1]},\mathbf{y}^{[k+1]})
+b_k(\mathbf{x}^{[k+1]}-\mathbf{x}^{[k]})
=0.
\end{align}
\end{subequations}
Combining the above equations and $\partial\Psi=\partial f+ \partial g+ \partial C_r$, and further applying Eq.~\eqref{epsilon1}, the following equation holds
\begin{align}
&
\lim_{k\rightarrow \infty}
\mathbf{dist}(0, 
\partial {\Psi}(\mathbf{x}^{[k]},\mathbf{y}^{[k]})
=\mathbf{dist}(0, 
\partial {\Psi}(\mathbf{x}^{*},\mathbf{y}^{*})
\leq \epsilon. \nonumber
\end{align}
This proves that every limit point $\{\mathcal{X}^{*},\mathcal{Y}^{*}\}$ of $\{\mathcal{X}^{[k]},\mathcal{Y}^{[k]}\}$ is a critical point for problem $\mathsf{Q}$.
Consequently, to conclude the proof, we only need to show that the sequence converges in finite steps. 
By leveraging the Kurdyka-Lojasiewicz (KL) property of $\Psi$ and \cite{palm}[Theorem 1], it can be shown that $\{\mathcal{X}^{[k]},\mathcal{Y}^{[k]}\}$ is a Cauchy sequence and hence is a convergent sequence approaching the limit $\{\mathcal{X}^{*},\mathcal{Y}^{*}\}$.

\section{More Results by Ir-sim } \label{appendix_irsim}

\begin{figure*}[t]
  \centering
  \begin{subfigure}[t]{0.23\textwidth}
      \includegraphics[width=0.9\textwidth]{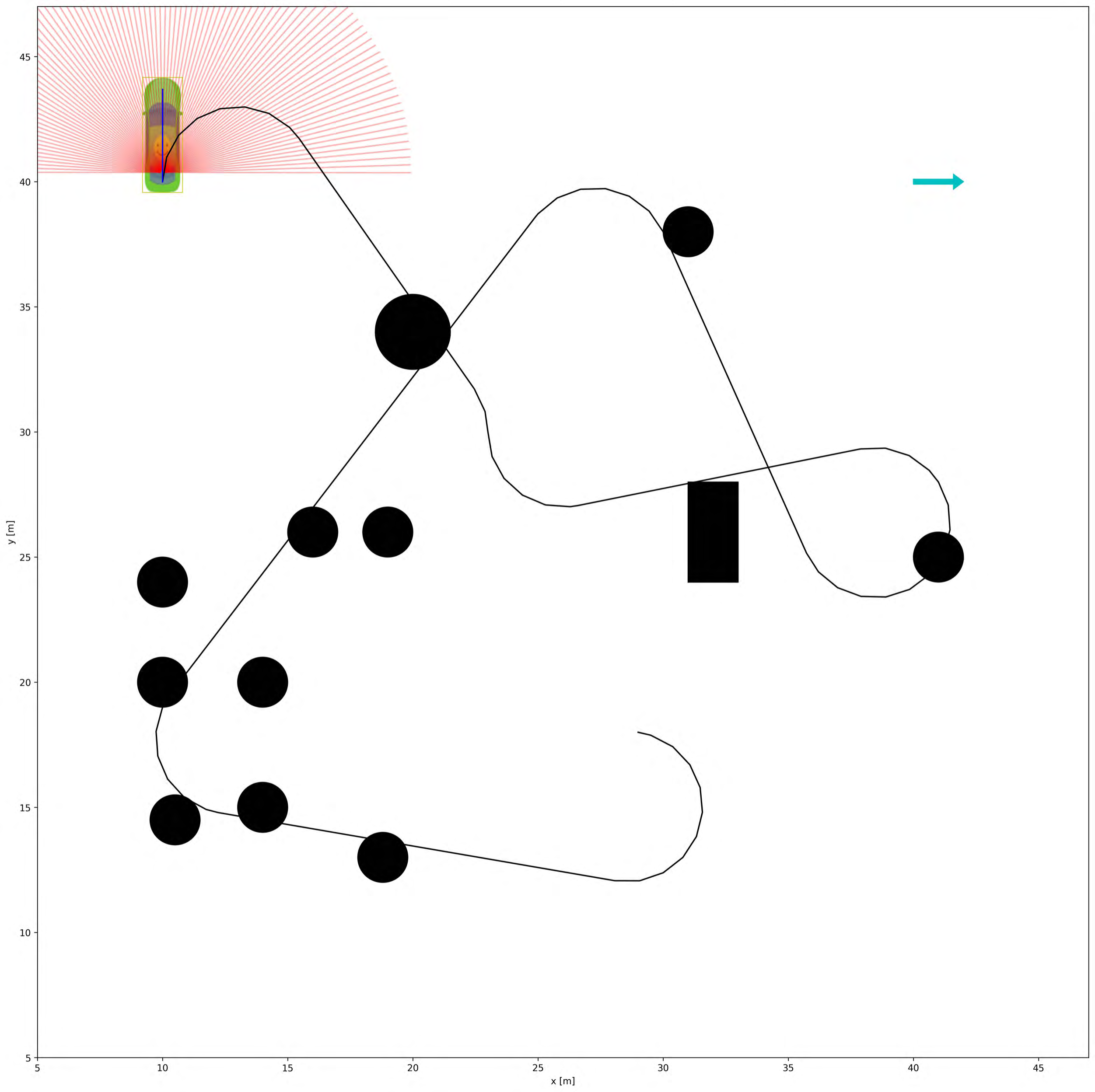}
      \caption{ RDA at time step 0 }
  \end{subfigure}
  \hfill
  \begin{subfigure}[t]{0.23\textwidth}
    \includegraphics[width=0.9\textwidth]{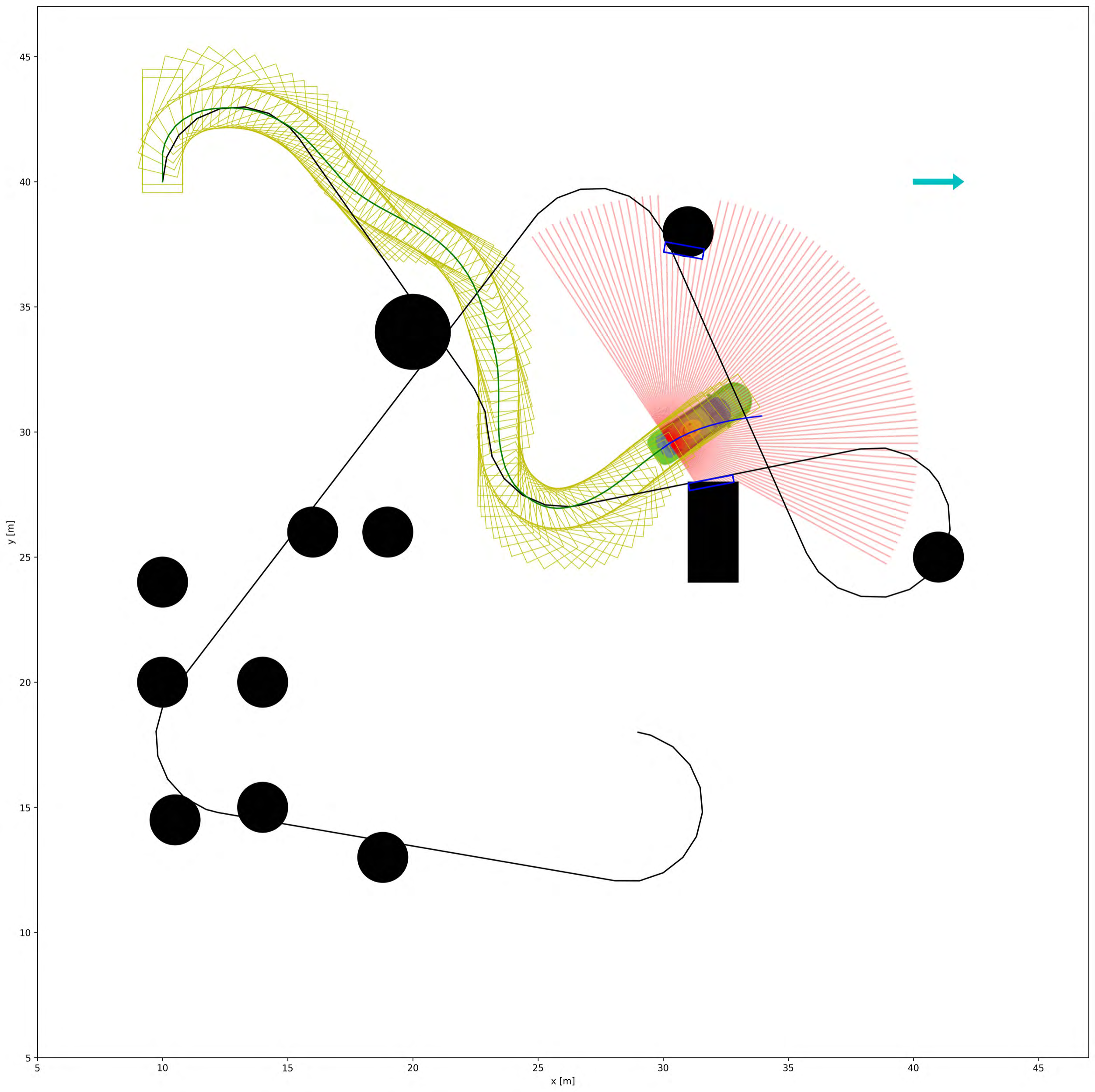}
    \caption{RDA at time step 93}
  \end{subfigure}
  \hfill
  \begin{subfigure}[t]{0.23\textwidth}
    \includegraphics[width=0.9\textwidth]{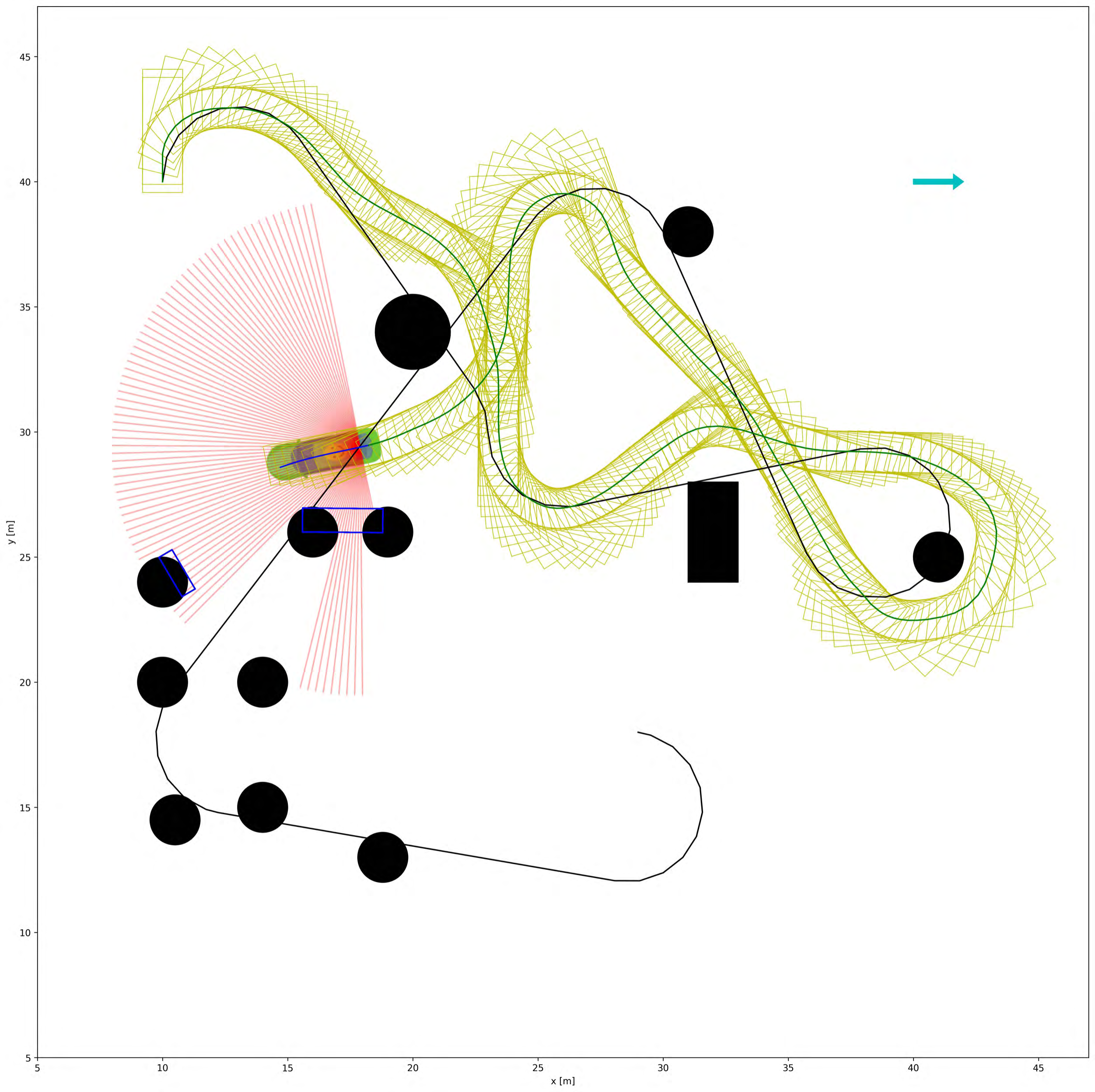}
    \caption{RDA at time step 241}
  \end{subfigure}
  \hfill
  \begin{subfigure}[t]{0.23\textwidth}
    \includegraphics[width=0.9\textwidth]{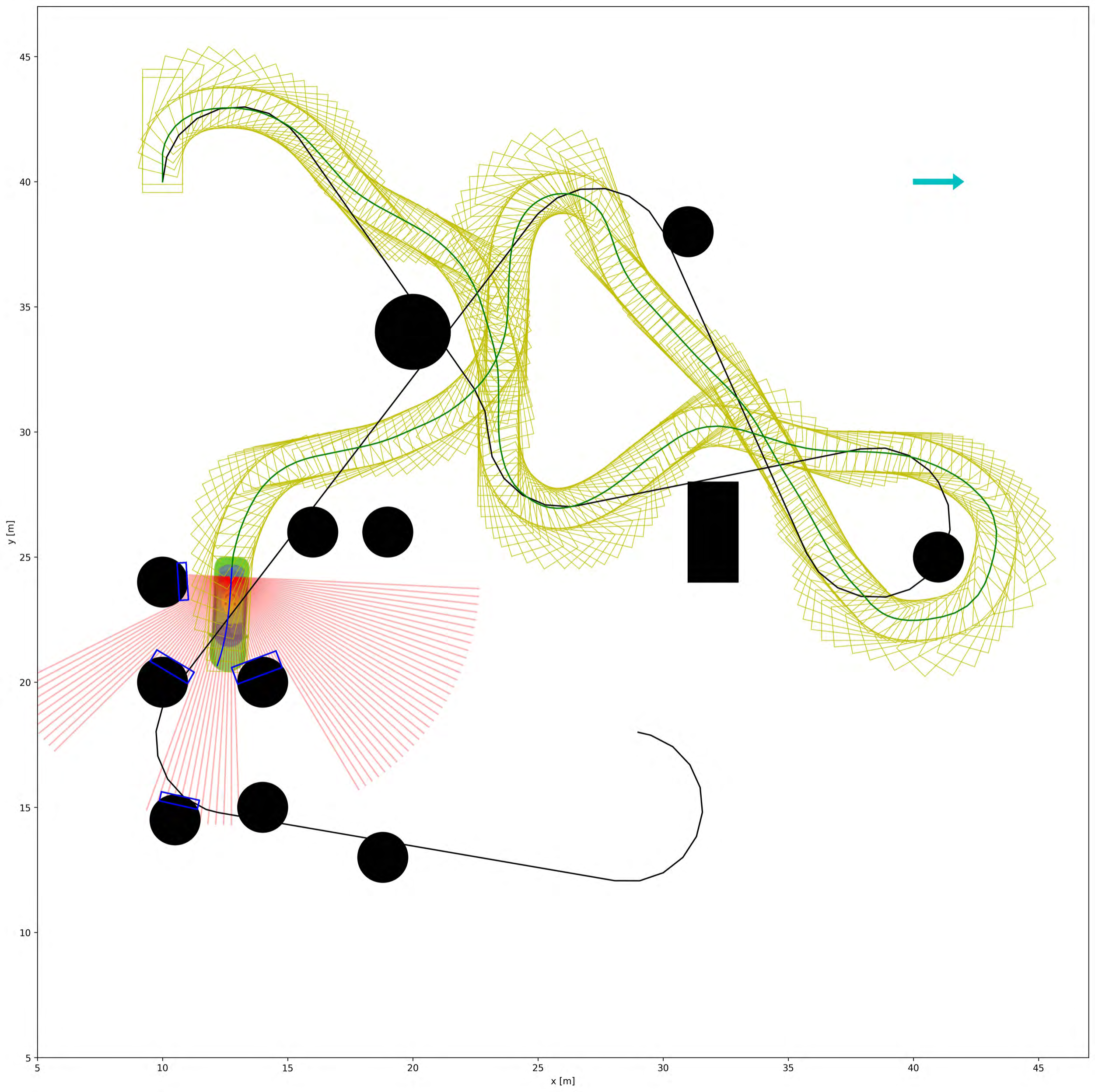}
    \caption{RDA gets stuck}
  \end{subfigure}

  \begin{subfigure}[t]{0.23\textwidth}
    \includegraphics[width=0.9\textwidth]{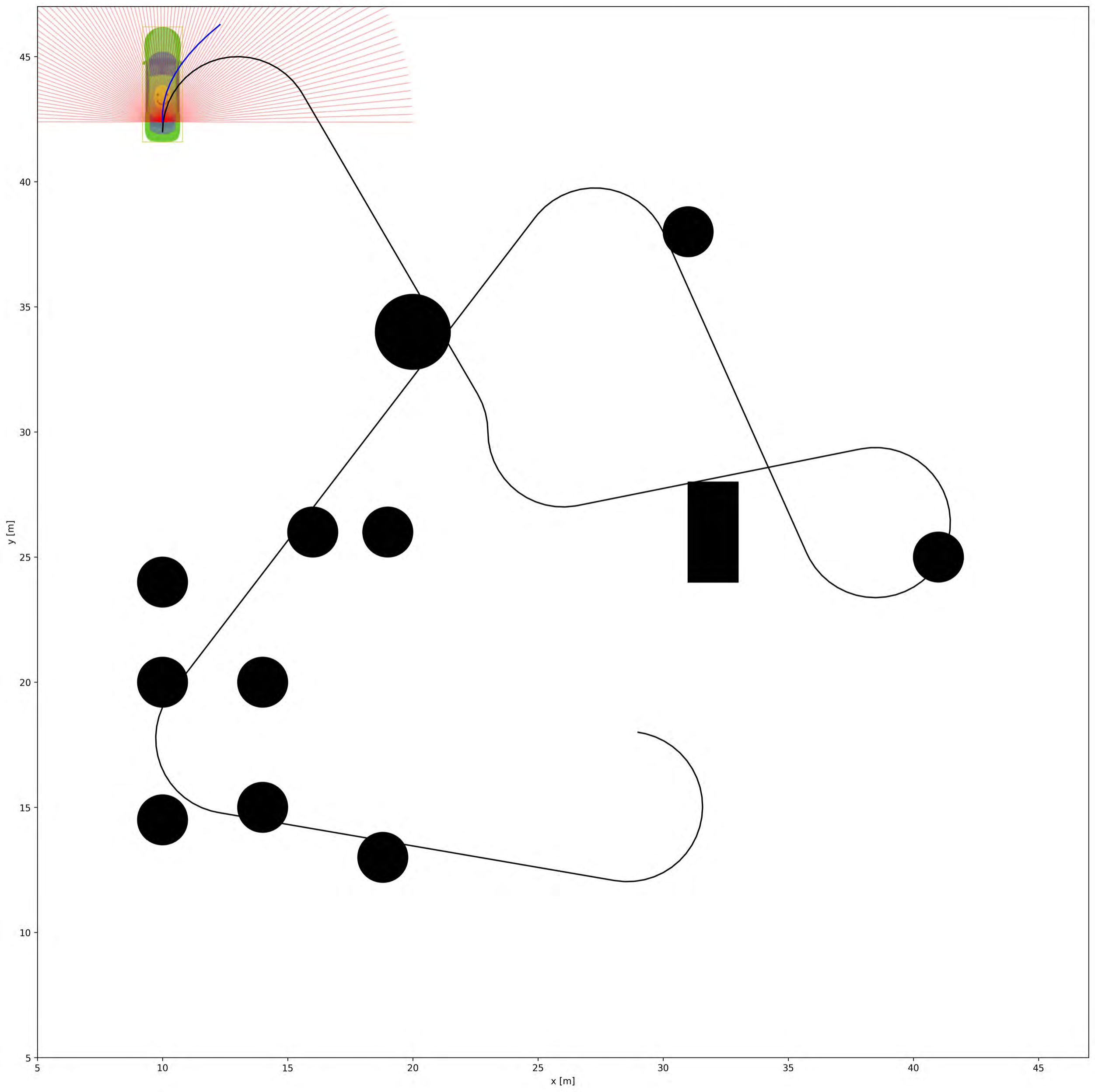}
    \caption{NeuPAN at time step 0 }
  \end{subfigure}
  \hfill
  \begin{subfigure}[t]{0.23\textwidth}
      \includegraphics[width=0.9\textwidth]{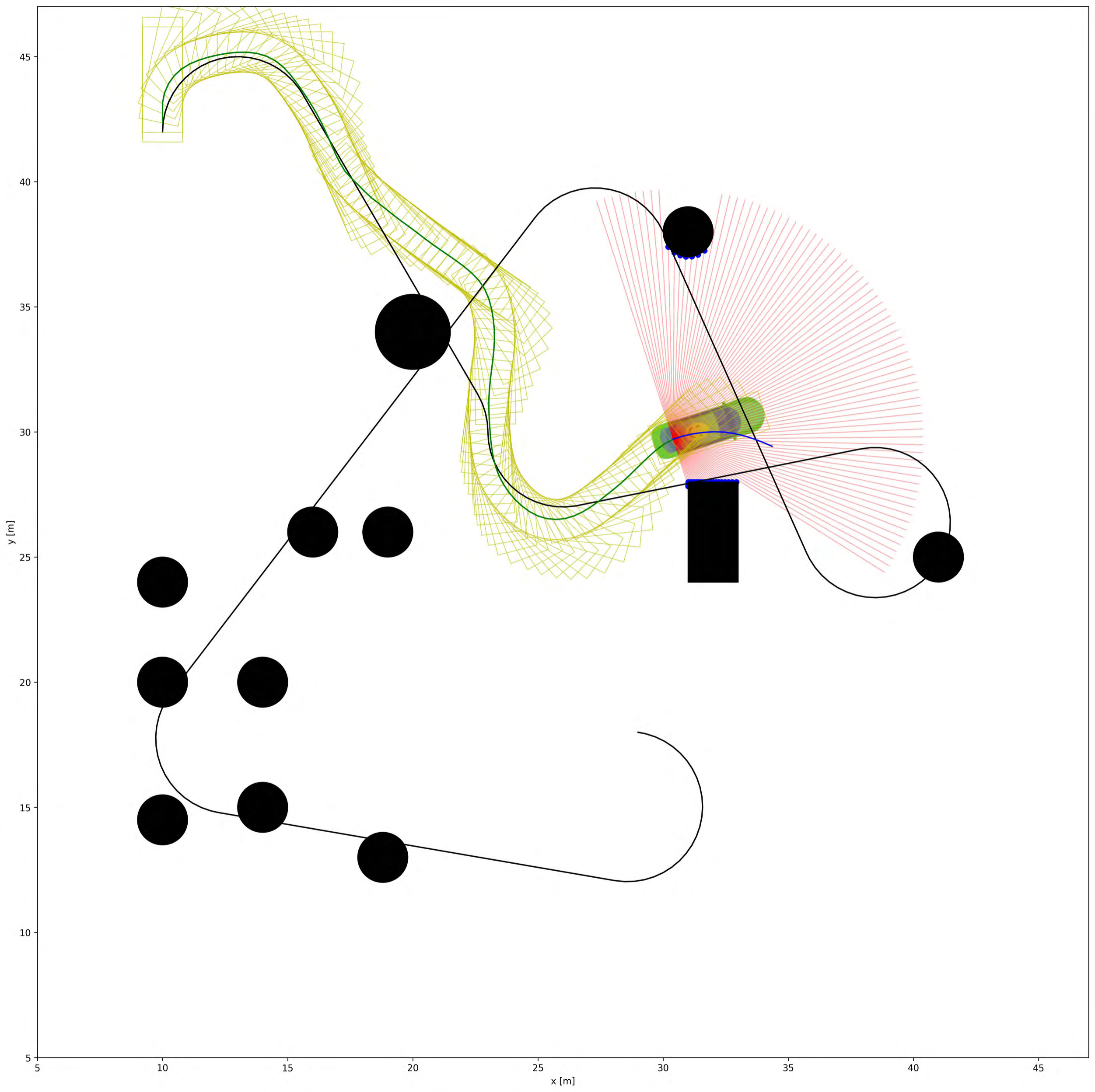}
      \caption{NeuPAN at time step 93 }
  \end{subfigure}
  \hfill
  \begin{subfigure}[t]{0.23\textwidth}
    \includegraphics[width=0.9\textwidth]{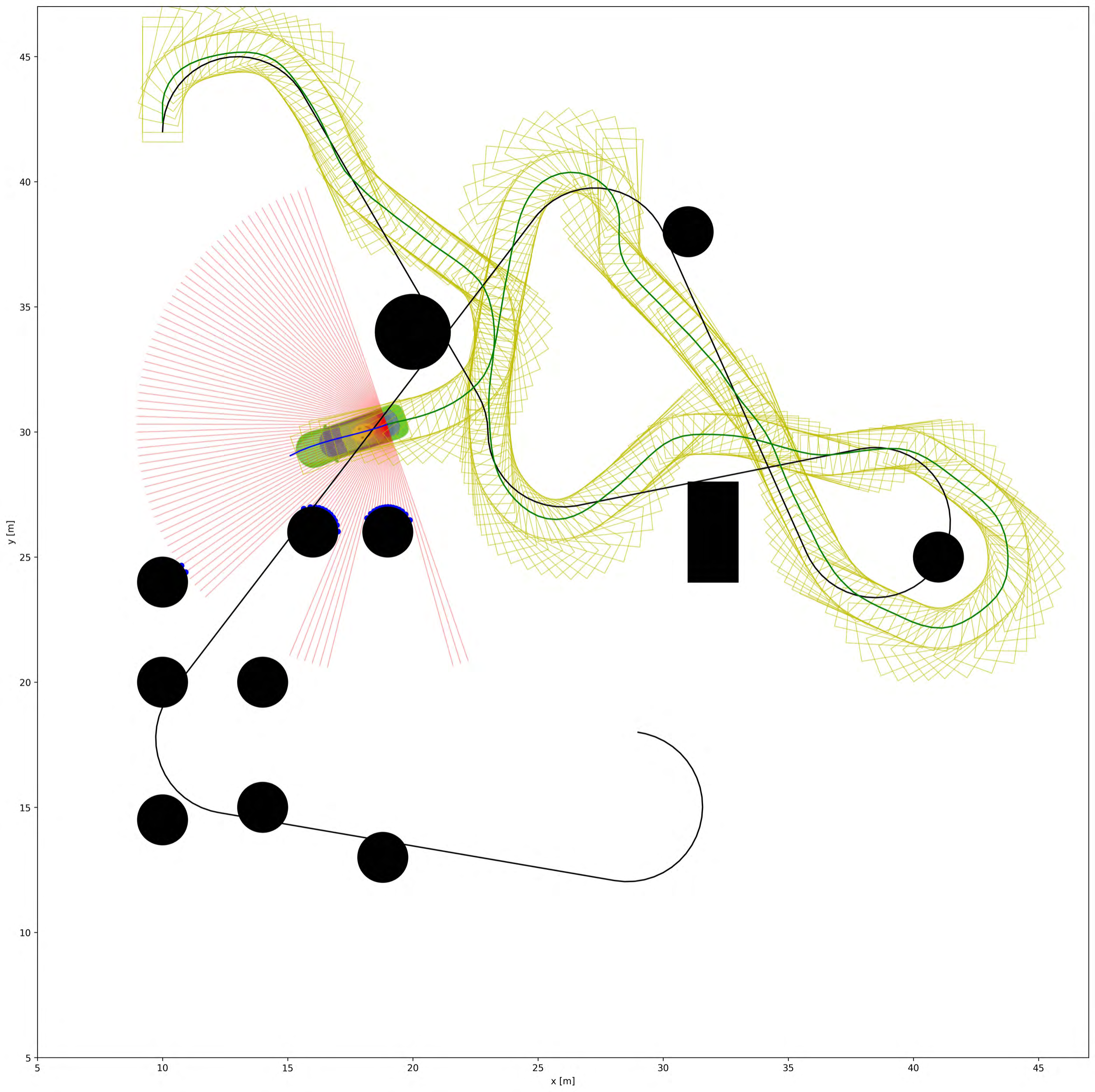}
    \caption{NeuPAN at time step 241}
  \end{subfigure}
  \hfill
  \begin{subfigure}[t]{0.23\textwidth}
    \includegraphics[width=0.9\textwidth]{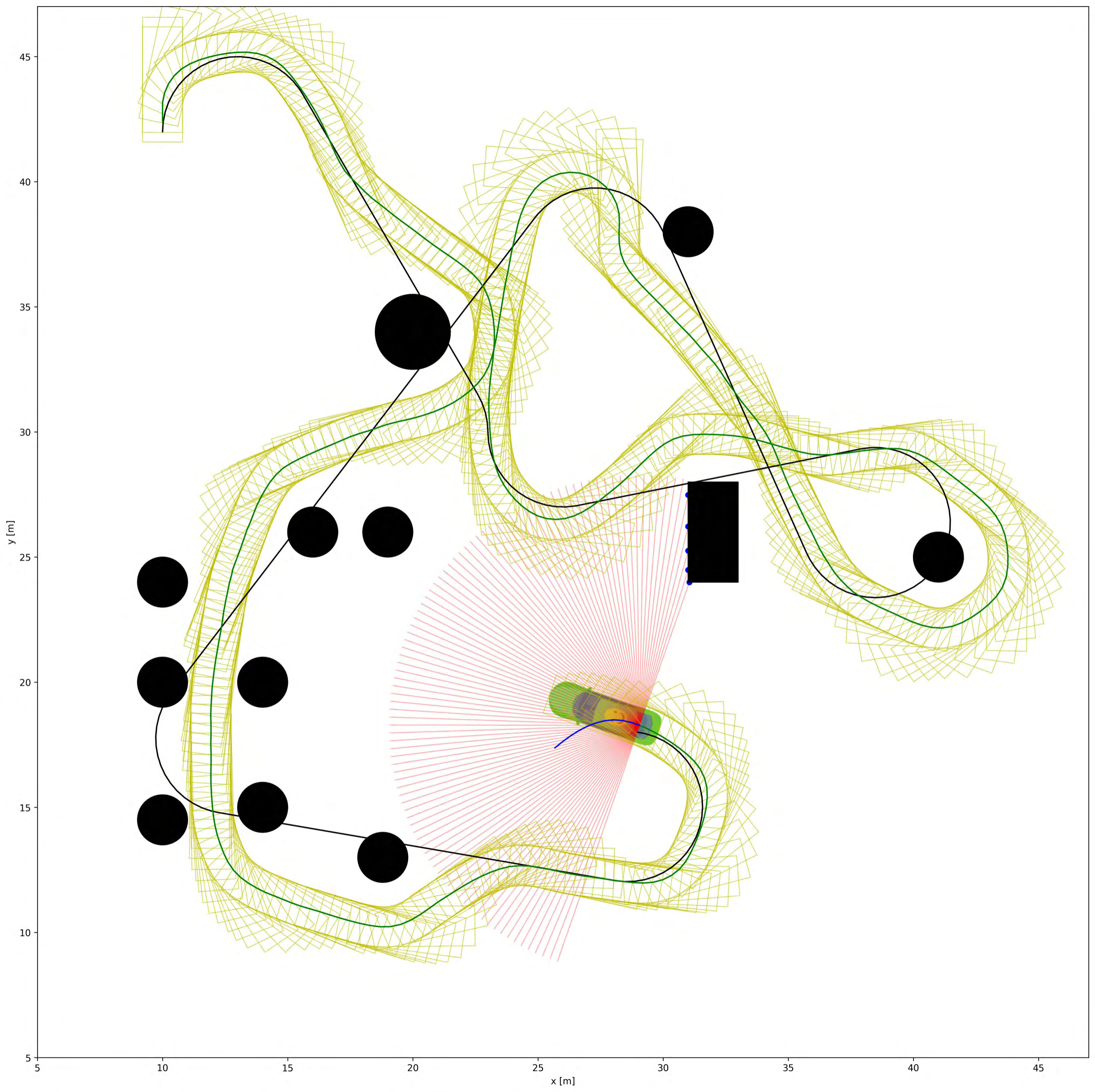}
    \caption{NeuPAN learns from failure}
  \end{subfigure}
  \caption{Navigation process and trajectory comparison of NeuPAN and RDA planner in the path following scenario. The robot is required to plan a trajectory (blue line) to follow the naive path (black line) avoiding the cluttered obstacles (black circle). RDA require the detection algorithm to convert the lidar scan to the convex objects (blue rectangle). }
  \label{sim1}
\end{figure*}

This section presents more simulated results in Ir-sim to showcase the effectiveness, efficiency, and trajectory quality of NeuPAN.

\begin{figure*}[t]
    \centering
    \begin{subfigure}[t]{0.19\textwidth}
        \includegraphics[width=0.95\textwidth]{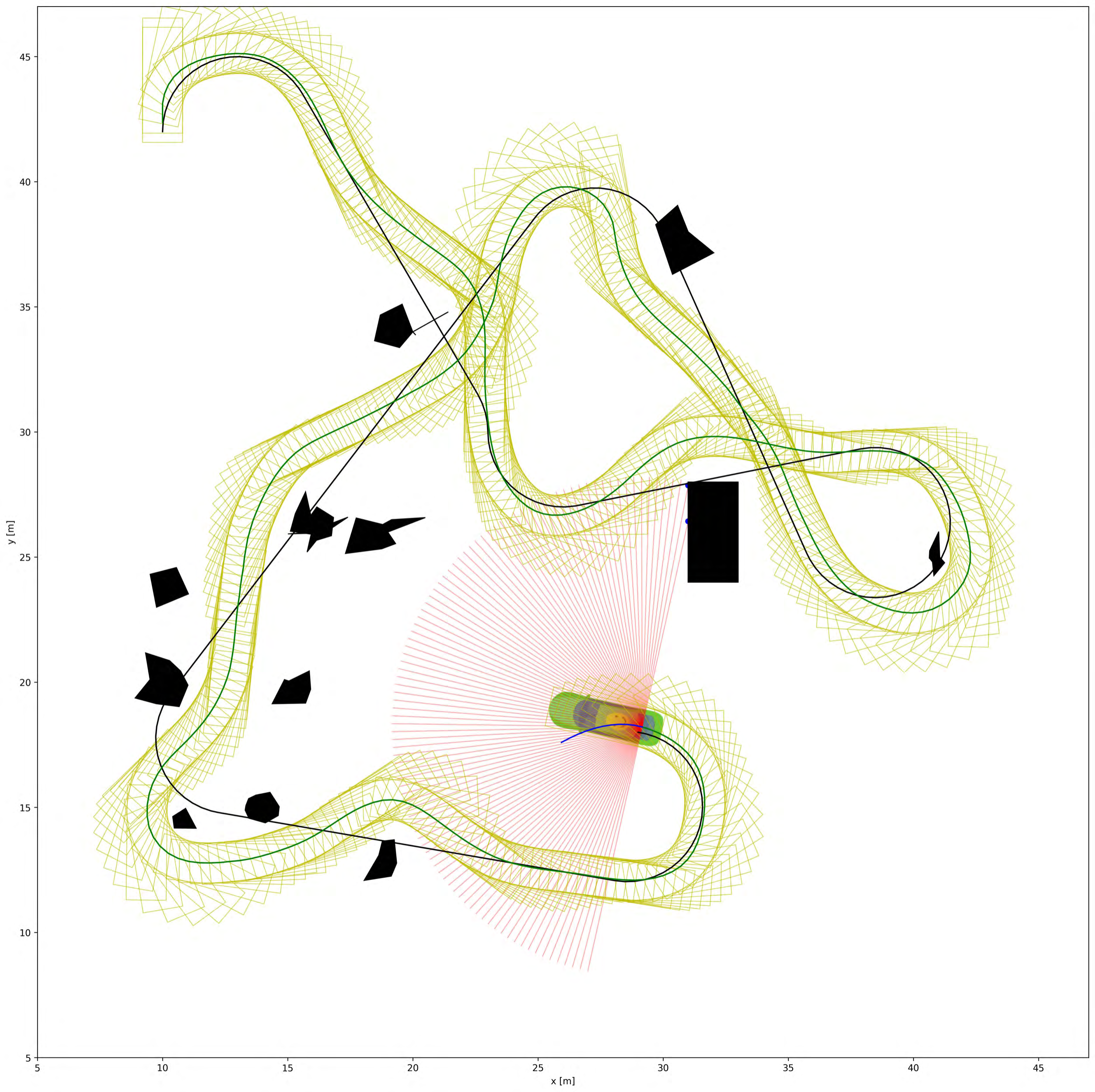}
        \caption{Acker. w/ nonconvex. }
        \label{npo_acker}
    \end{subfigure}
    \begin{subfigure}[t]{0.19\textwidth}
      \includegraphics[width=0.95\textwidth]{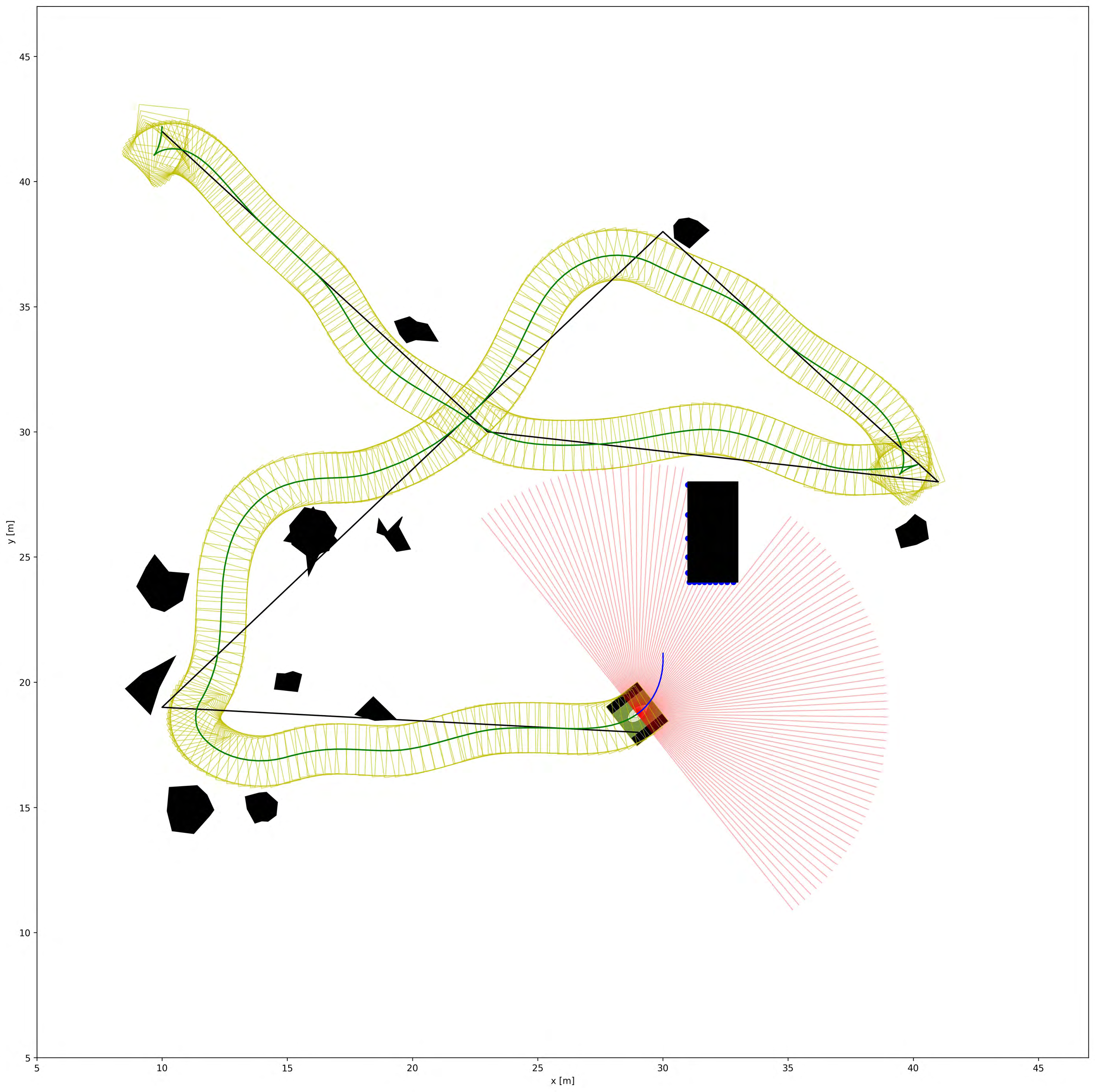}
      \caption{Diff. w/ nonconvex. }
      \label{npo_diff}
    \end{subfigure}
    \begin{subfigure}[t]{0.19\textwidth}
      \includegraphics[width=0.95\textwidth]{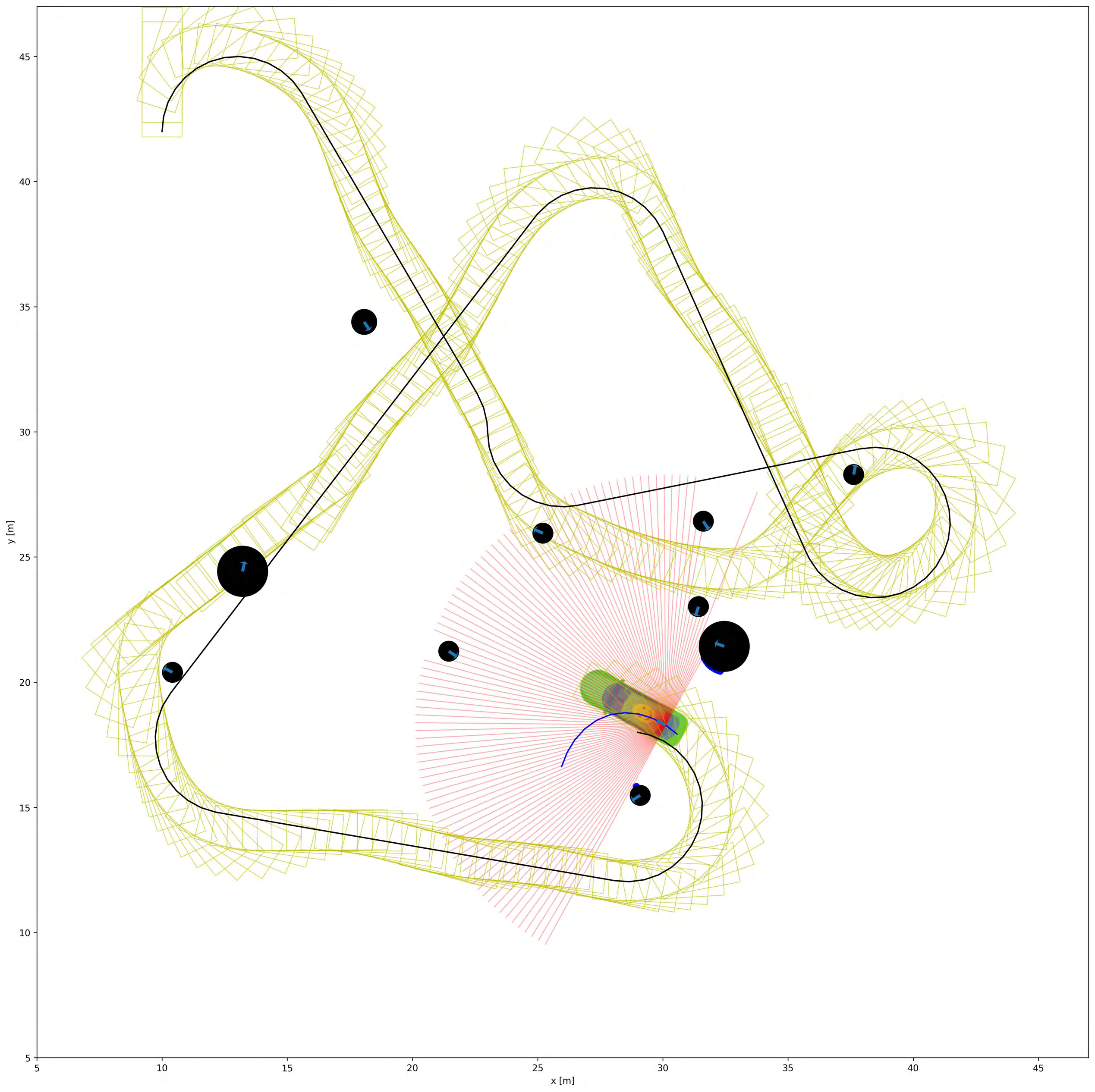}
      \caption{Acker. w/ dynamic. }
      \label{dyna_acker}
    \end{subfigure}
    \begin{subfigure}[t]{0.19\textwidth}
        \includegraphics[width=0.95\textwidth]{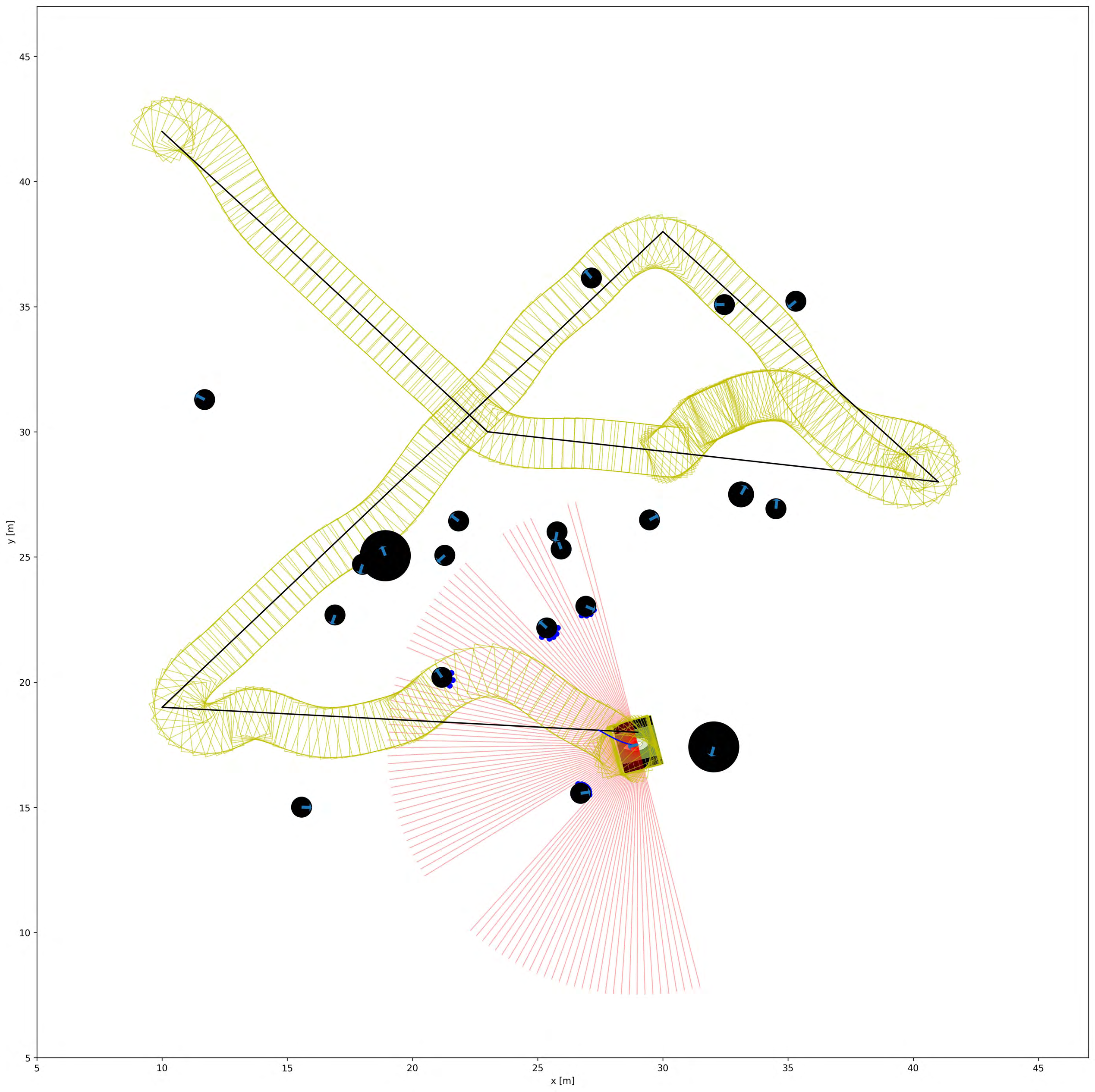}
        \caption{Diff. w/ dynamic. }
        \label{dyna_diff}
    \end{subfigure}
    \begin{subfigure}[t]{0.19\textwidth}
      \includegraphics[width=0.95\textwidth]{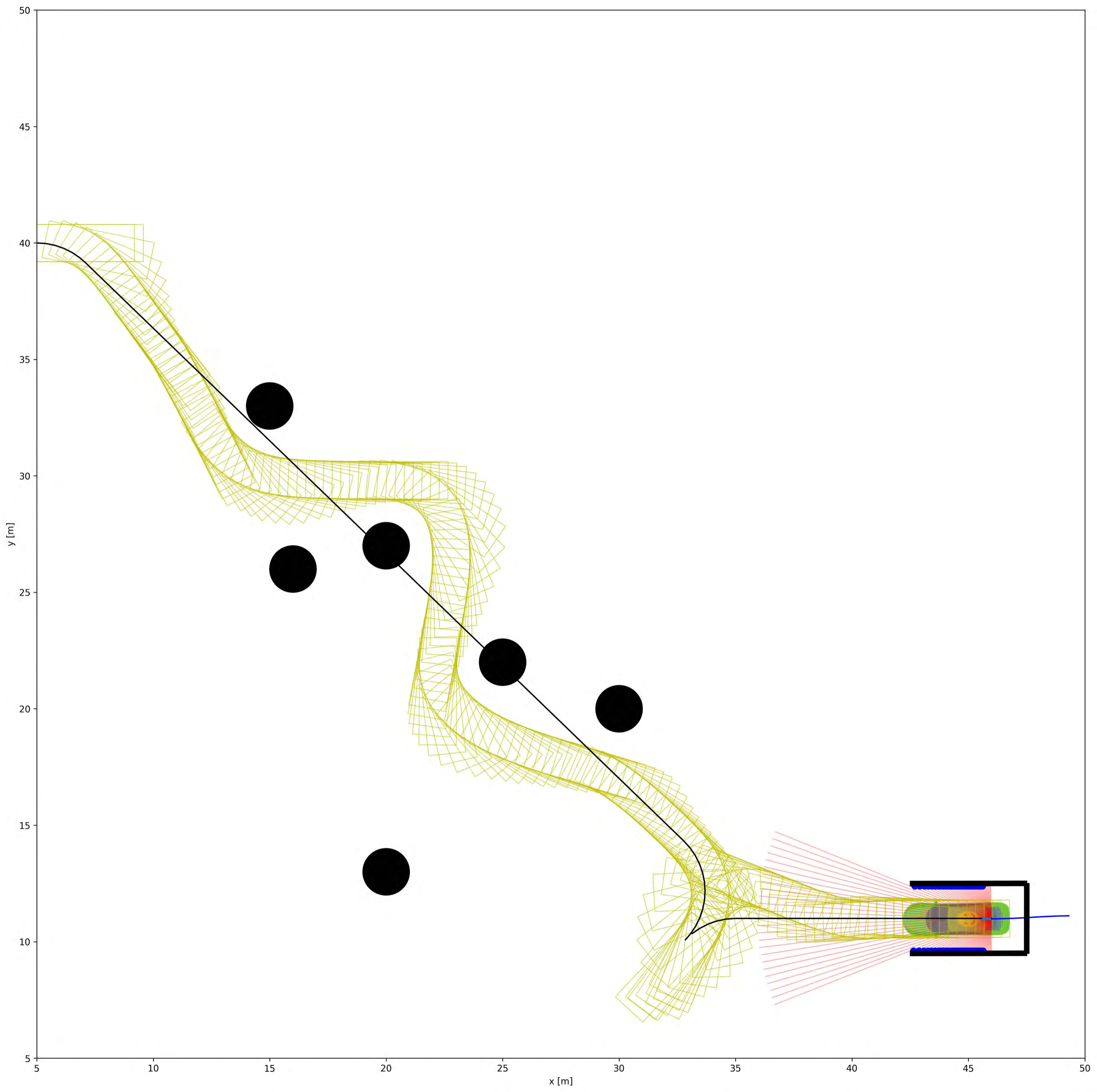}
      \caption{Acker. w/ parking. }
      \label{reverse_acker}
    \end{subfigure}
    \caption{Verification of NeuPAN in various scenarios on two types of robots.}
    \label{sim2}
    \vspace{-0.5cm}
\end{figure*}

\subsection{Path Following} 
In this scenario, the robot needs to follow a predefined route full of cluttered obstacles, as depicted in Fig.~\ref{sim1} (a). A car-like rectangular robot equipped with a 2D 100-line lidar sensor is considered. 
The trajectories obtained from the benchmark and the NeuPAN at different frames are shown in Fig.~\ref{sim1}(a)-(d) and Fig.~\ref{sim1}(e)-(h), respectively. 
It can be seen that both schemes are capable of maneuvering the robot in open spaces with sparsely distributed obstacles.
However, since the benchmark scheme RDA needs to transform laser scans into boxes or convex sets using object detection (as shown in Fig.~\ref{sim1}(c)), errors are inevitably introduced between the detection results and the actual objects. 
As a consequence, even under ideal lidar and detector in the considered experiment, the modular approach may still result in collisions in densely cluttered environments (as shown in Fig.~\ref{sim1}(d)). 
In contrast, the proposed NeuPAN directly processes the points, thus bypassing intermediate steps, resulting in a more accurate navigation performance in cluttered scenarios. 

\subsection{Nonconvex, Dynamic, and Parking Scenarios} 

To illustrate the comprehensiveness of NeuPAN, we conduct experiments in diverse scenarios using two different robot models: the car-like robot (Ackermann) and the differential-drive robot.
The target scenarios and robot trajectories are depicted in Fig.~\ref{sim2}. 
First, Fig.~\ref{npo_acker} and Fig.~\ref{npo_diff} demonstrate the capability of NeuPAN passing through marginal gaps between nonconvex obstacles. 
To date, nonconvex obstacles pose a significant challenge for existing optimization-based approaches~\cite{zhang2020optimization, han2023rda}.
Approximating these nonconvex objects as convex forms would lead to inaccurate representations and dividing them into a group of convex sets would lead to high computation costs. 
In contrast, the proposed direct point approach efficiently address these complex, nonconvex, and unstructured obstacles.
Second, the validation of NeuPAN in dynamic collision avoidance scenarios is illustrated in Figs.~\ref{dyna_acker} and \ref{dyna_diff}. 
Finally, the verification of NeuPAN in the reverse parking scenario is presented in Fig.~\ref{reverse_acker}. 
These results demonstrate the comprehensiveness and generalizability of our NeuPAN system, which corroborate Table~\ref{related_work}.

\begin{figure*}[tb]
  \centering
  \begin{subfigure}[t]{0.22\textwidth}
      \includegraphics[width=0.9\textwidth]{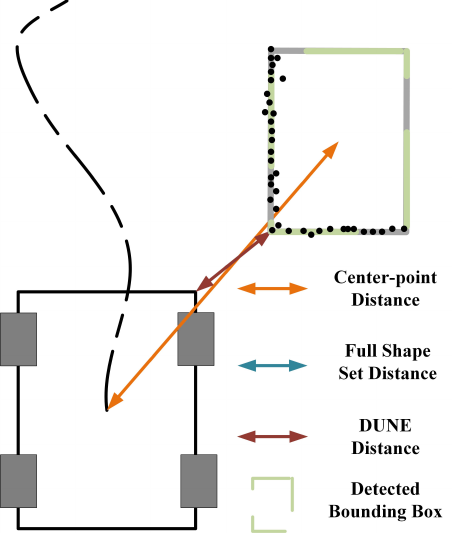}
      \caption{ Box detection without error }
      \label{collision_check:1}
  \end{subfigure}
  \hfill
  \begin{subfigure}[t]{0.22\textwidth}
    \includegraphics[width=0.9\textwidth]{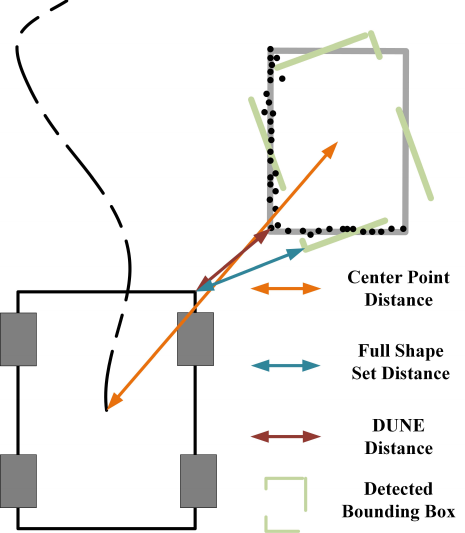}
    \caption{ Box detection with error }
    \label{collision_check:2}
  \end{subfigure}
  \begin{subfigure}[t]{0.22\textwidth}
    \includegraphics[width=0.9\textwidth]{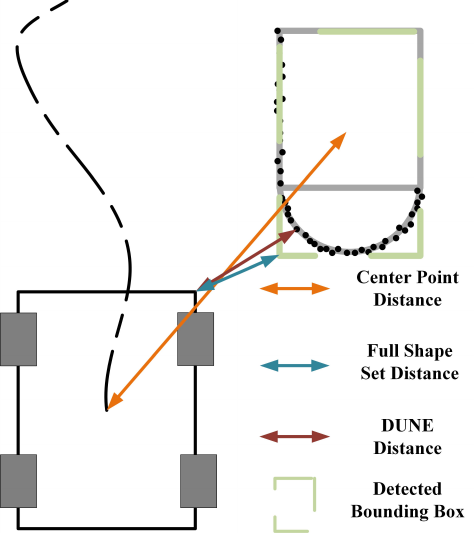}
    \caption{ Non-box convex objects}
    \label{collision_check:3}
  \end{subfigure}
  \hfill
  \begin{subfigure}[t]{0.22\textwidth}
    \includegraphics[width=0.9\textwidth]{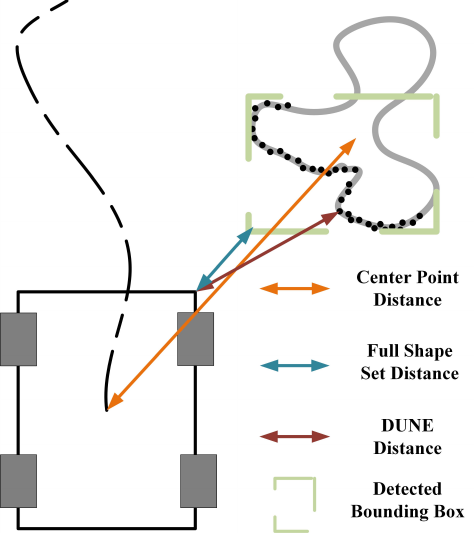}
    \caption{Arbitrary shape objects}
    \label{collision_check:4}
  \end{subfigure}
  \caption{Intuitive comparison of DUNE distances, center point distances, and full shape set distances. DUNE distances are more accuracy than other distances.}
  \label{collision_check}
\end{figure*}

\begin{figure*}[ht]
  \centering
  \begin{subfigure}[t]{0.30\textwidth}
      \centering
      \includegraphics[width=0.95\textwidth]{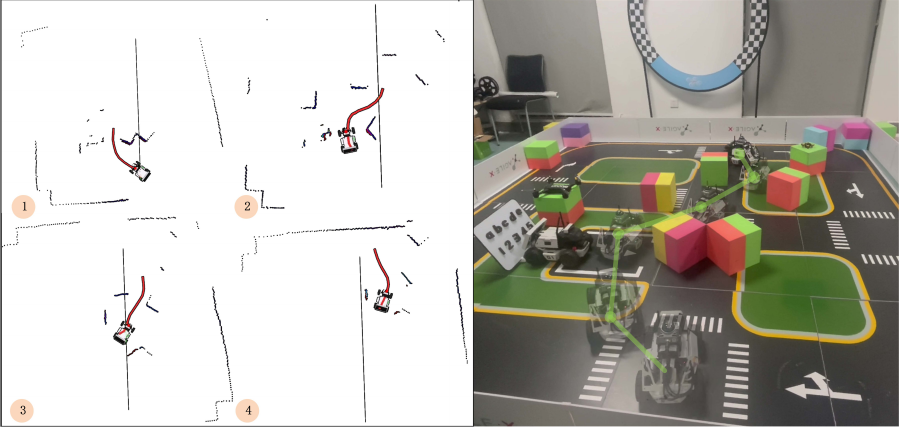}
      \caption{Differential robot navigation through crowded cubes.}
      \label{limo_real1}
  \end{subfigure}
  \hfill
  \begin{subfigure}[t]{0.30\textwidth}
    \centering
    \includegraphics[width=0.95\textwidth]{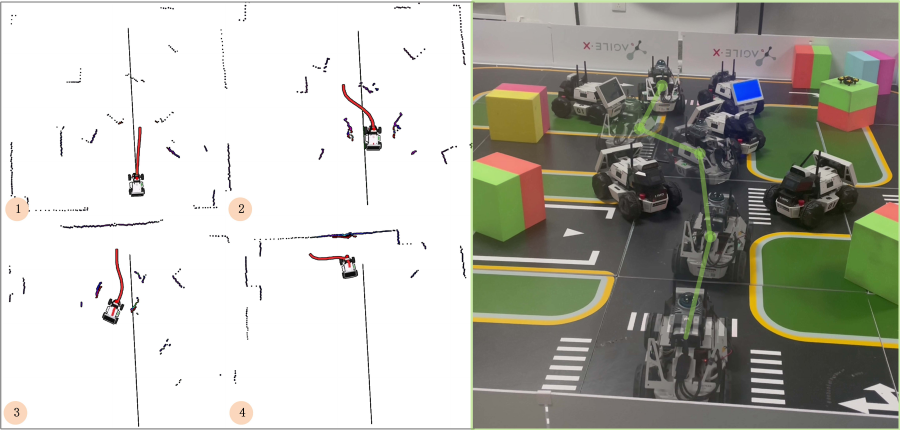}
    \caption{Differential robot navigation through nonconvex robots.}
    \label{limo_real2}
  \end{subfigure}
  \hfill
  \begin{subfigure}[t]{0.3\textwidth}
      \centering
      \includegraphics[width=0.95\textwidth]{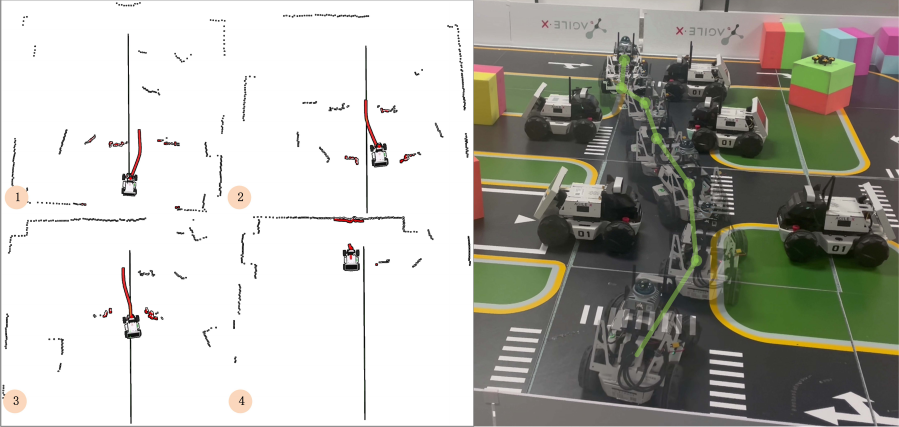}
      \caption{Ackermann robot navigation through nonconvex robots.}
      \label{limo_real3}
    \end{subfigure}
  \caption{Navigating differential and Ackermann robots in real world testbed. Red lines are NeuPAN generated trajectories. Black lines are the straight line from the start to end points. Black points are lidar scans. }
  \label{limo_real}
\end{figure*}

\section{Comparison with Other Distances}\label{distance_compare}
To demonstrate the high-accuracy property of DUNE, we compare the exact distances generated by DUNE and inexact distances generated by other existing approaches. 
An intuitive comparison is illustrated in Fig.~\ref{collision_check}. 
It can be seen that the center point distance has the lowest accuracy, since it does not take the object shape into account. 
The full shape set distance is accurate if and only if the object has a box shape and the detector is $100\,\%$ accurate as shown in Fig.~\ref{collision_check:1}. 
Otherwise, the set distance would involve non-negligible distance errors as shown in Fig.~\ref{collision_check:2} (if the detector is inaccurate) and Fig.~\ref{collision_check:3} (if the shape is not a box).
Lastly, the proposed DUNE can handle arbitrary nonconvex shapes as shown in Fig.~\ref{collision_check:4}, whereas all other approaches break down.

\section{More Demonstrations in the real world} \label{appendix_ground}

To this end, we conduct real-world experiments in a confined testbed to demonstrate the robustness and sim-to-real generalizability of the proposed NeuPAN. The associated results are presented in Fig.~\ref{limo_real}. 
The robot is tasked to navigate through a crowd of obstacles to reach the goal point by onboard lidar. 
Fig.~\ref{limo_real1} shows that the ego-robot under differential steering mode passes the ``impassable'' area filled with crowded cubes. We have tested solutions including TEB, A-Star, Hybrid A-Star, OBCA, RDA on this challenge and all of them fail. That is why we call it ``impassable''. 
Then, this challenge is further upgraded by replacing the box-shape cubes with nonconvex-shape robots.
In such a dense and unstructured geometry, the robot driven by NeuPAN is still able to navigate efficiently, as shown in Fig.~\ref{limo_real2}. 
Finally, we switch the robot mode from differential to Ackermann, which imposes more limitations on robot movements. 
Under the above change of robot dynamics, most existing learning-based approaches require model retraining. 
Nonetheless, due to the embedded physical engine in NRMP and the size of the robot does not change, model retraining is not needed for NeuPAN.
That is, NeuPAN trained on the differential robot succeeds in directly adapting to the Ackermann robot, as demonstrated in Fig.~\ref{limo_real3}. 
All these experimental results confirm the high-accurate and easy-to-deploy features of NeuPAN in real-world dense (but structured) scenarios.

\noindent 


\end{document}